\documentclass{article}

\usepackage{microtype}
\usepackage{graphicx}
\usepackage{subcaption}
\usepackage{booktabs}
\usepackage{hyperref}
\usepackage{xurl}
\usepackage{amsmath}
\usepackage{amssymb}
\usepackage{mathtools}
\usepackage{amsthm}
\usepackage[capitalize,noabbrev]{cleveref}
\usepackage[textsize=tiny]{todonotes}

\theoremstyle{plain}
\newtheorem{theorem}{Theorem}

\newtheorem{lemma}{Lemma}
\newtheorem{corollary}{Corollary}
\theoremstyle{definition}

\theoremstyle{plain}
\newtheorem{assumption}{Assumption}
\newtheorem{remark}{Remark}

\newcommand{\R}{\mathbb{R}}
\newcommand{\V}{\mathcal{V}}
\newcommand{\D}{\mathcal{D}}
\newcommand{\X}{\mathcal{X}}
\newcommand{\I}{\mathcal{I}}

\newcommand{\Loss}{\mathcal{L}}

\newcommand{\softmax}{\operatorname{softmax}}
\newcommand{\Tr}{\mathrm{Tr}}
\newcommand{\op}{\mathrm{op}}
\newcommand{\ip}[2]{\left\langle #1,#2\right\rangle}
\newcommand{\1}{\mathbf{1}}

\makeatletter
\let\revision@addcontentsline\addcontentsline
\newcommand{\appendixcontents}{
  \begingroup
  \color{CrossReferenceRed}
  \section*{Appendix Contents}
  \setcounter{tocdepth}{2}
  \setlength{\parskip}{2pt}
  \@starttoc{apc}
  \endgroup
}
\newcommand{\enableappendixcontents}{
  \setcounter{secnumdepth}{2}
  \numberwithin{equation}{section}
  \numberwithin{theorem}{section}
  \numberwithin{lemma}{section}
  \numberwithin{algorithm}{section}
  \numberwithin{corollary}{section}
  \numberwithin{figure}{section}
  \numberwithin{table}{section}
  \let\revision@origsection\section
  \let\revision@origsubsection\subsection
  \renewcommand{\section}{\@ifstar{\revision@origsection*}{\revision@section}}
  \renewcommand{\subsection}{\@ifstar{\revision@origsubsection*}{\revision@subsection}}
}
\newcommand{\revision@section}[1]{
  \revision@origsection{#1}
  \revision@addcontentsline{apc}{section}{\protect\numberline{\thesection}#1}
}
\newcommand{\revision@subsection}[1]{
  \revision@origsubsection{#1}
  \revision@addcontentsline{apc}{subsection}{\protect\numberline{\thesubsection}#1}
}
\makeatother

\usepackage[preprint]{icml2026}
\definecolor{ReferenceBlue}{rgb}{0,0.19,0.56}
\definecolor{CrossReferenceRed}{rgb}{0.7,0,0}
\hypersetup{
  colorlinks=true,
  linkcolor=CrossReferenceRed,
  citecolor=ReferenceBlue,
  urlcolor=ReferenceBlue,
  filecolor=ReferenceBlue,
  linktoc=all
}

\makeatletter
\renewcommand{\ICML@preprint}{\textit{Preprint.}}
\makeatother
\usepackage{etoolbox}
\patchcmd{\printAffiliationsAndNotice}
  {Correspondence to:}
  {Email:}
  {}
  {\PackageWarning{author-info}{Could not replace correspondence label}}

\icmltitlerunning{Theoretical Analysis of Muon on Transformers for Next-Token Prediction}

\begin{document}

\twocolumn[
  \icmltitle{Theoretical Analysis of Muon on Transformers for Next-Token Prediction}
  \icmlsetsymbol{equal}{*}

  \begin{icmlauthorlist}
    \icmlauthor{Shengwei Wu}{equal,ZJU}
    \icmlauthor{Yixuan Yang}{equal,ZJU}
    \icmlauthor{Gao Huang}{equal,Singapore}
    \icmlauthor{Song Li}{ZJU}
  \end{icmlauthorlist}

  \icmlaffiliation{ZJU}{School of Mathematical Sciences, Zhejiang University, Hangzhou, Zhejiang, China}
  \icmlaffiliation{Singapore}{Department of Mathematics, National University of Singapore, Singapore}

  \icmlcorrespondingauthor{Shengwei Wu}{wsw24@zju.edu.cn}
  \icmlcorrespondingauthor{Yixuan Yang}{yixuanyang@zju.edu.cn}
  \icmlcorrespondingauthor{Gao Huang}{hgmath@nus.edu.sg}
  \icmlcorrespondingauthor{Song Li}{songli@zju.edu.cn}

  \icmlkeywords{implicit bias, Muon, softmax attention, maximum margin}

  \vskip 0.3in
]

\printAffiliationsAndNotice{\icmlEqualContribution}

\begin{abstract}
Muon has recently shown strong empirical performance in training Transformer-based large language models (LLMs), yet its optimization dynamics remain poorly understood from a theoretical perspective.
In this paper, we analyze the training dynamics of Muon for next-token prediction (NTP) in Transformers.
The training procedure consists of two phases: the preprocessing phase learns the output-value component, while the main phase learns the key-query component. 
For the preprocessing phase, we characterize the optimization trajectory exactly and show that the normalized spectral margin is optimal after each update.
For the main phase, we establish nonasymptotic convergence rates for the training loss and the normalized margin gap under softmax attention.
These results suggest that Muon's update rule can help models efficiently learn which tokens matter for prediction, offering a possible explanation for its strong performance in large language model training.
The learned weight structure also yields a generalization guarantee for unseen contexts.
Furthermore, we validate our theoretical insights with experiments on next-token prediction tasks.
\end{abstract}

\section{Introduction}
The Transformer \cite{vaswani2017attention} has become a popular architecture in modern machine learning.
Large language models (LLMs) built on this architecture have demonstrated strong capabilities in language understanding \cite{devlin2019bert}, text generation and in-context learning \cite{brown2020language}, and reasoning tasks \cite{wei2022chain}.
Training these models requires substantial data and computational resources, and the choice of optimizer can affect both training efficiency and model performance.
Understanding the behavior of optimization algorithms can therefore help reduce training costs and guide the development of more effective training methods for Transformers.

Adaptive first order methods such as Adam and AdamW are standard choices for Transformer training \cite{kingma2014adam,loshchilov2017decoupled,vaswani2017attention,brown2020language}.
These methods treat parameters as vectors and adapt the update scale of each coordinate separately, without explicitly using the matrix structure of the weights.
Matrix structure are particularly relevant to Transformers, where key and query projections interact through inner products to determine attention scores, while value and output transformations jointly map the attended information to prediction logits.
More recently, Muon was proposed as an optimizer for hidden layer weight matrices, constructing update directions by approximately orthogonalizing matrix momentum \cite{jordan2024muon}.
It has shown strong empirical performance in image classification and language model training, including large language model pretraining \cite{jordan2024muon,liu2025muon}.

For a loss function $\Loss(W)$ with a matrix parameter $W$, Muon updates the momentum and parameter matrices as follows:
\begin{align*}
& M_t=\beta M_{t-1}+(1-\beta)\nabla\Loss(W_t),\\
& W_{t+1}=W_t-\eta_tU_tV_t^\top.
\end{align*}
Here, $\beta\in[0,1)$ is the momentum parameter, and $\eta_t>0$ is the step size.
The matrices $U_t$ and $V_t$ are the left and right singular vector matrices in the compact singular value decomposition $M_t=U_t\Sigma_tV_t^\top$, where $\Sigma_t$ contains the nonzero singular values on its diagonal.
Muon thus constructs its update direction by preserving the singular vectors of the momentum matrix and setting its nonzero singular values to one, yielding the compact polar factor $U_tV_t^\top$.
In practice, this factor is approximated using Newton--Schulz iterations, avoiding an explicit singular value decomposition \cite{bernstein2024old,jordan2024muon}.

In this paper, we study the training dynamics of Muon for next-token prediction in a one-layer Transformer model.
Training this model requires learning how to map tokens to output labels and select relevant information from the context.
Our study focuses on the following question:
\begin{center}
\textit{How does Muon optimize Transformers for next-token prediction?}
\end{center}

We investigate this question through a two phase training procedure.
During the preprocessing phase (Phase I), Muon trains the linear mapping from token representations to output scores using contexts containing a single token.
During the main phase (Phase II), this mapping remains unchanged while Muon trains the matrix that determines the attention scores of tokens in the context.

\paragraph{Contributions.}
We characterize Muon's training dynamics under this two phase procedure.
\begin{itemize}
  \item We characterize the entire Muon trajectory in the preprocessing phase.
  Every normalized iterate attains the optimal spectral margin, and the cross entropy loss decays at an explicit rate.
  We further prove that the training loss in the main phase converges to the limit determined by the preprocessing phase.
  Together, these results clarify the roles of the two phases: preprocessing determines how confidently each token predicts its target label, while the main phase learns to direct attention toward the correct token in the context.

  \item In the main phase, we prove that the normalized spectral margin converges to its optimal value.
  We also derive explicit nonasymptotic bounds for the training loss and the normalized margin gap, quantifying how Muon's updates help attention distinguish the correct token from other tokens in the same context.
  With step sizes proportional to $(t+1)^{-1/2}$, our bound on the normalized margin gap is $O((1+\log t)/\sqrt{t})$.
  This is the fastest asymptotic order among the bounds obtained for the power law schedules considered, providing theoretical support for square root step size decay.

  \item We establish a finite iteration lower bound on the probability of the correct next token in unseen contexts.
  This offers a theoretical perspective on the generalization ability of the Transformer architecture, showing how attention can reuse learned token relations to support prediction in new contexts.
\end{itemize}

\section{Preliminaries}
\paragraph{Notations.}
For a matrix $A$, let $A^{\top}$ denote its transpose and $A[i,j]$ denote its $(i,j)$-th entry.
For a positive integer $m$, write $[m]:=\{1,\ldots,m\}$.
For a vector $u$, the notation $\Vert u\Vert_2$ denotes the Euclidean norm. 
The Frobenius inner product is defined by $\ip{A}{B}:=\Tr(A^\top B)$.
We use $\Vert A\Vert_{F}$, $\Vert A\Vert_{\op}$, and $\Vert A\Vert_{\ast}$ to denote the Frobenius, spectral, and nuclear norms, respectively.
The spectral and nuclear norms are dual: $\Vert A\Vert_{\ast} = \sup_{\Vert B\Vert_{\op}\le 1}\ip{A}{B}$. 
For a vector $a\in\R^m$, the softmax function is defined coordinatewise by $[\softmax(a)]_i=\exp(a_i)/\bigl(\sum_{j=1}^m\exp(a_j)\bigr)$ for $i\in[m]$.
The vector $e_i\in\R^K$ denotes the $i$-th standard basis vector.
For nonnegative sequences $a_t$ and $b_t$, with $b_t>0$ for all sufficiently large $t$, we write $a_t=O(b_t)$ if $a_t\le Cb_t$ and $a_t=\Omega(b_t)$ if $a_t\ge cb_t$ for all sufficiently large $t$, where $C,c>0$ are constants independent of $t$.
We write $a_t=o(b_t)$ if $a_t/b_t\to0$ as $t\to\infty$.

\paragraph{Next-token prediction.}
Let $\V\subset\R^d$ be a finite vocabulary of $K$ orthonormal token embeddings, where $2\le K\le d$.
In high dimensional spaces, suitably normalized random vectors are often nearly pairwise orthogonal \cite{vershynin2019high}.
We adopt the exact orthonormal model to remove correlations among token representations and focus on the effect of Muon's matrix updates.

Let $I:\V\to[K]$ be a bijection assigning each token an index.
The training dataset consists of $N$ samples,
\begin{equation*}
  \D:=\left((X^{(n)},x_{L_n+1}^{(n)})\right)_{n=1}^{N},
\end{equation*}
where $X^{(n)}=[x_1^{(n)},\ldots,x_{L_n}^{(n)}]$ is a nonempty context of tokens in $\V$ and $x_{L_n+1}^{(n)}\in\V$ is the next token.
For each sample, $q_n:=x_{L_n}^{(n)}$ is the query token and $y_n:=I(x_{L_n+1}^{(n)})$ is the target label.
We assume that for every $n\in[N]$ and $1\le j<L_n$, $\D$ also contains a sample that uses the first $j$ tokens of $X^{(n)}$ to predict $x_{j+1}^{(n)}$.

For the preprocessing phase, let $\D_0$ denote the set of all training pairs whose contexts contain a single token.
We assume that every token in $\V$ appears as a context in $\D_0$.
The corresponding target labels are specified by a bijection $h:\V\to[K]$, so that $h(x)=I(x')$ for every $([x],x')\in\D_0$.

\paragraph{Architecture.}
We consider a one-layer Transformer model with softmax attention and a linear prediction layer \cite{huang2024non,li2024mechanics,ildiz2024self,tarzanagh2023transformers}.
Let $X=[x_1,\ldots,x_L]\in\R^{d\times L}$ denote an input sequence and $q:=x_L$ its query token.
Let $W^{\mathrm{k}},W^{\mathrm{q}},W^{\mathrm{v}}\in\R^{d\times d}$ be the key, query, and value matrices, respectively, and let $W^{\mathrm{o}}\in\R^{K\times d}$ be the output matrix.
The model produces the following probability vector over the vocabulary:
\begin{equation*}
  f(X):=\softmax\left(W^{\mathrm{o}}W^{\mathrm{v}}X\softmax\left((W^{\mathrm{k}}X)^\top W^{\mathrm{q}}q\right)\right).
\end{equation*}
The inner softmax assigns attention weights to the contextual tokens, and the outer softmax converts prediction logits into vocabulary probabilities.

To simplify the theoretical analysis, we directly train the combined prediction and attention matrices, following similar simplifications in related theoretical studies \cite{tarzanagh2023transformers,li2024mechanics,ildiz2024self,huang2024non,huang2025theoretical,huang2025transformers,julistiono2026mirror}:
\begin{equation*}
  W_1:=W^{\mathrm{o}}W^{\mathrm{v}}\in\R^{K\times d},\qquad
  W_2:=(W^{\mathrm{k}})^\top W^{\mathrm{q}}\in\R^{d\times d}.
\end{equation*}
We apply Muon directly to $W_1$ and $W_2$, which determine the prediction logits and attention scores, respectively.
Writing $\theta:=(W_1,W_2)$, the prediction probabilities are
\begin{equation*}
  f_\theta(X):=\softmax(W_1X\alpha_\theta(X)),
\end{equation*}
where $\alpha_\theta(X):=\softmax(X^\top W_2q)$ is the vector of attention weights.
The vector $X\alpha_\theta(X)$ is a weighted average of the contextual token embeddings, and $W_1$ maps this representation to prediction logits.

The vector $f_\theta(X)\in\R^K$ gives the probabilities of the vocabulary tokens through the index map $I$.
In particular, the probability assigned to the next token $x_{L+1}$ is $\mathbb P_\theta\{x_{L+1}\mid X\}:=[f_\theta(X)]_{I(x_{L+1})}$.

\paragraph{Loss.}
We train the model by minimizing the empirical cross entropy loss
\begin{equation*}
  \Loss(\theta):=-\frac{1}{N}\sum_{n=1}^N\log\left(e_{y_n}^\top f_\theta(X^{(n)})\right).
\end{equation*}

\paragraph{Algorithm.}
We train the model with Muon in two phases, as summarized in Algorithm \ref{alg:two-phase-muon}.
Both $W_1$ and $W_2$ are initialized at zero.

During the preprocessing phase (Phase I), we update only $W_1$ on $\D_0$ to learn the correspondence between tokens and their target labels.
Since each context in $\D_0$ contains a single token, its attention weight is one, and the prediction depends only on $W_1$.
The loss in the preprocessing phase is obtained by replacing the full dataset $\D$ with $\D_0$ in the definition of $\Loss$:
\begin{equation*}
  \Loss_0(W_1):=-\frac{1}{K}\sum_{x\in\V}\log\left(e_{h(x)}^\top\softmax(W_1x)\right).
\end{equation*}

After $T$ preprocessing iterations, we train attention on the full dataset $\D$ by updating only $W_2$ in the main phase (Phase II).
The prediction matrix $W_{1,T}$ remains unchanged, and the corresponding training loss is $\Loss_T(W_2):=\Loss(W_{1,T},W_2)$.

\begin{algorithm}[t]
\caption{Two Phase Muon Training}\label{alg:two-phase-muon}
\begin{algorithmic}[1]
\STATE \textbf{\textsc{Phase I}: Preprocessing}
\STATE \textbf{Input} $\D_0$; $T$; $\{\eta_{1,s}\}_{s=0}^{T-1}$; $\beta_1\in[0,1)$.
\STATE \textbf{Initialization} $W_{1,0}=0$ and $M_{1,-1}=0$.
\FOR{$s=0,1,\ldots,T-1$}
\STATE $M_{1,s}=\beta_1 M_{1,s-1}+(1-\beta_1)\nabla\Loss_0(W_{1,s})$.
\STATE $(U_{1,s},\Sigma_{1,s},V_{1,s})=\operatorname{SVD}(M_{1,s})$.
\STATE $W_{1,s+1}=W_{1,s}-\eta_{1,s} U_{1,s} (V_{1,s})^\top$.
\ENDFOR
\STATE \textbf{Output} $W_{1,T}$.
\par\vspace{0.4ex}\hrule\vspace{0.4ex}
\STATE \textbf{\textsc{Phase II}: Main Training}
\STATE \textbf{Input} $\D$; $W_{1,T}$; $t$; $\{\eta_{2,r}\}_{r=0}^{t-1}$; $\beta_2\in[0,1)$.
\STATE \textbf{Initialization} $W_{2,0}=0$ and $M_{2,-1}=0$.
\FOR{$r=0,1,\ldots,t-1$}
\STATE $M_{2,r}=\beta_2 M_{2,r-1}+(1-\beta_2)\nabla\Loss_T(W_{2,r})$.
\STATE $(U_{2,r},\Sigma_{2,r},V_{2,r})=\operatorname{SVD}(M_{2,r})$.
\STATE $W_{2,r+1}=W_{2,r}-\eta_{2,r}U_{2,r}(V_{2,r})^\top$.
\ENDFOR
\STATE \textbf{Output} $(W_{1,T},W_{2,t})$.
\end{algorithmic}
\end{algorithm}

\section{Main Results}
This section is organized as follows.
We first introduce the assumptions for the two phase analysis.
We then present the exact characterization of the preprocessing phase and the nonasymptotic bounds for the training loss and normalized margin in the main phase.
Finally, we study prediction on unseen contexts.

We impose three assumptions on the training contexts.

\begin{assumption}\label{ass:token-label}
For every $n\in[N]$, the context $X^{(n)}$ contains the token $x_n^\star:=h^{-1}(y_n)$.
\end{assumption}

Assumption \ref{ass:token-label} requires each context to contain a token whose label under $h$ matches the target label.
We refer to $x_n^\star$ as the correct token in the context.

To state the remaining two assumptions, let $c_{n,x}$ denote the number of occurrences of token $x\in\V$ in $X^{(n)}$.
We write $\X_n^-:=\{x\in\V:c_{n,x}>0,\ x\ne x_n^\star\}$ for the set of tokens appearing in the context other than the correct token.

\begin{assumption}\label{ass:weak-count}
For every $n\in[N]$, the correct token occurs at least as often as any other token in the same context:
\begin{equation*}
  c_{n,x_n^\star}\ge c_{n,x}\qquad\text{for every }x\in\X_n^-.
\end{equation*}
\end{assumption}

Assumption \ref{ass:weak-count} ensures that the correct token receives at least as much total attention weight as any other token at zero initialization.
Indeed, each position receives weight $1/L_n$, so the total attention weight over all occurrences of token $x$ is $c_{n,x}/L_n$.

\begin{assumption}\label{ass:separability}
There exists a matrix $W\in\R^{d\times d}$ that separates the correct token from every other token in each training context:
\begin{equation*}
  (x_n^\star-x)^\top W q_n\ge 1
  \qquad(n\in[N],\ x\in\X_n^-).
\end{equation*}
\end{assumption}

Assumption \ref{ass:separability} requires a common matrix that assigns the correct token a strictly larger attention score than every other token in each training context.
The separating matrix $W$ need not minimize the spectral norm.
Similar separability assumptions are used in attention maximum-margin analyses \cite{tarzanagh2023transformers,vasudeva2024implicit,julistiono2026mirror}.

\subsection{Preprocessing Phase: Exact Training Dynamics}
We first analyze the preprocessing phase (Phase I), which updates $W_1$ using examples whose contexts contain only one token.
We then turn to Phase II, where $W_1$ remains fixed at $W_{1,T}$ and only $W_2$ is updated.

In this phase, $W_1$ learns to distinguish each token's target label from the other labels.
To characterize the label separation achieved by Muon's iterates, we introduce a spectral hard margin problem associated with the preprocessing dataset $\D_0$.
Let $R_1^\star$ denote its optimal value:
\begin{equation}\label{eq:hard-margin-ov}
  \begin{aligned}
    \min_{W\in\R^{K\times d}}\quad & \Vert W\Vert_{\op} \\
    \text{subject to}\quad & (e_{h(x)}-e_v)^\top Wx \ge 1, \\
    & \forall x\in\V,\quad v\in[K]\setminus\{h(x)\}.
  \end{aligned}
\end{equation}
Let $\gamma_1:=1/R_1^{\star}$ denote the optimal normalized spectral margin. 
For any $W\in\R^{K\times d}$, define the unnormalized margin by
\begin{equation*}
  \rho_1(W):=\min_{x\in\V}\min_{v\in[K]\setminus\{h(x)\}}(e_{h(x)}-e_v)^{\top}Wx.
\end{equation*}
The following theorem characterizes every preprocessing iterate exactly.

\begin{theorem}\label{thm:phase1}
Consider the preprocessing phase (Phase I) iterates generated by Algorithm \ref{alg:two-phase-muon}.
Fix $\beta_1\in[0,1)$ and $\eta_{1,0}>0$, and take $\eta_{1,s}=\eta_{1,0}(s+1)^{-1/2}$ for $s\ge0$.
For each integer $T\ge1$, define $\Delta_T:=\sum_{s=0}^{T-1}\eta_{1,s}$.
Then, for every $x\in\V$ and $v\in[K]\setminus\{h(x)\}$,
\begin{equation}\label{eq:thm1-logit-gap}
  (e_{h(x)}-e_v)^\top W_{1,T}x
  =\Vert W_{1,T}\Vert_{\op}
  =\Delta_T.
\end{equation}
The loss satisfies
\begin{equation}\label{eq:thm1-loss}
\begin{aligned}
  \Loss_0(W_{1,T})
  &=\log\left(1+(K-1)e^{-\Delta_T}\right)\\
  &\le(K-1)e^{-\eta_{1,0}\sqrt{T}}.
\end{aligned}
\end{equation}
Moreover, $R_1^\star=1$, and every normalized iterate attains the optimal spectral margin:
\begin{equation}\label{eq:thm1-margin}
  \rho_1\left(\frac{W_{1,T}}{\Vert W_{1,T}\Vert_{\op}}\right)
  =\gamma_1
  =1.
\end{equation}
\end{theorem}
Theorem \ref{thm:phase1} shows that Muon attains the optimal normalized spectral margin after the first update and maintains it throughout the preprocessing phase.
Subsequent updates enlarge the common logit gap $\Delta_T$, increasing the probability assigned to each token's target label and reducing the cross entropy loss.

\begin{remark}
The uniform weights $1/K$ in $\D_0$, orthonormal token embeddings, and zero initialization keep the gradient and momentum proportional to a fixed matrix, giving Muon a constant polar update direction.
The iterates therefore depend on the step sizes only through $\Delta_T$, and the exact identities hold for any positive step size sequence.
The square root schedule gives the loss decay bound \eqref{eq:thm1-loss}.
\end{remark}

\subsection{Main Phase: Spectral Maximum Margin Convergence}
The preprocessing phase produces a prediction matrix $W_{1,T}$ that associates each token with its target label.
With this matrix unchanged, the main phase (Phase II) trains $W_2$ to assign the correct token a larger attention score than every other token in the same context.
To quantify this separation relative to the spectral norm of $W_2$, we introduce a spectral hard margin problem based on the attention score gaps.

For an attention matrix $W\in\R^{d\times d}$, the score gap between the correct token $x_n^\star$ and another token $x\in\X_n^-$ is $(x_n^\star-x)^\top Wq_n$.
We consider the nontrivial case in which at least one training context contains a token other than its correct token.
Let $R_2^\star$ denote the minimum spectral norm needed to make every such gap at least one:
\begin{equation}\label{eq:hard-margin-kq}
\begin{aligned}
  \min_{W\in\R^{d\times d}}\quad & \Vert W\Vert_{\op}\\
  \text{subject to}\quad & (x_n^\star-x)^\top Wq_n\ge1,\\
  & \forall n\in[N],\quad x\in\X_n^-.
\end{aligned}
\end{equation}
Assumption \ref{ass:separability} ensures that this problem is feasible.

The unnormalized training margin is the smallest attention score gap across the training contexts:
\begin{equation*}
  \rho_2(W):=\min_{\substack{n\in[N],x\in\X_n^-}}
  (x_n^\star-x)^\top Wq_n.
\end{equation*}
Let $\gamma_2:=1/R_2^\star$ denote the spectral maximum margin value, which is the largest value of $\rho_2(W)$ attainable under $\Vert W\Vert_{\op}\le1$.

Our main result gives nonasymptotic bounds for the training loss and the normalized spectral margin gap in the main phase.
We write $\ell_T^\star:=\Loss_0(W_{1,T})$ for the loss after $T$ preprocessing iterations, whose exact value is given in \eqref{eq:thm1-loss}.

\begin{theorem}\label{thm:main}
Suppose that Assumptions \ref{ass:token-label}, \ref{ass:weak-count}, and \ref{ass:separability} hold.
Fix the preprocessing parameters as in Theorem \ref{thm:phase1}, and let $\beta_2\in[0,1)$.
There exists $\bar T\ge0$ such that, for every $T>\bar T$, there exists $\bar\eta_T>0$ with the following property.
For any $0<\eta_{2,0}\le\bar\eta_T$, let $\{W_{2,t}\}_{t\ge0}$ be the main phase iterates generated by Algorithm \ref{alg:two-phase-muon} after $T$ preprocessing iterations, with step sizes $\eta_{2,t}=\eta_{2,0}(t+1)^{-1/2}$.
Then there exist constants $C_1,C_2,C_3,C_4>0$, independent of $t$, such that the following bounds hold.

\textbf{\textup{(i) Training loss.}}
For every $t\ge1$,
\begin{equation}\label{eq:thm2-loss}
  0<\Loss_T(W_{2,t})-\ell_T^\star
  \le C_1t^{C_2}e^{-C_3\sqrt{t}}.
\end{equation}

\textbf{\textup{(ii) Normalized spectral margin.}}
There exists $t_T\ge1$ such that, for every $t\ge t_T$, $W_{2,t}\ne0$ and
\begin{equation}\label{eq:thm2-margin}
  0\le\gamma_2-\rho_2\left(\frac{W_{2,t}}{\Vert W_{2,t}\Vert_{\op}}\right)
  \le C_4\frac{1+\log t}{\sqrt{t}}.
\end{equation}

The constants $C_1,C_2,C_3,C_4$ and $t_T$ depend only on $K$, $\Delta_T$, $\beta_2$, $\eta_{2,0}$, and the finite training data.
\end{theorem}

Together, Theorems \ref{thm:phase1} and \ref{thm:main} characterize how Muon's updates affect token prediction and attention training.
In the preprocessing phase, Muon attains the optimal normalized spectral margin after the first update, and subsequent updates increase the probability assigned to each token's target label.
The resulting prediction matrix encodes the correspondence between tokens and target labels and determines the limiting prediction confidence in the main phase.
With this matrix unchanged, the main phase increases attention score gaps and concentrates attention on the correct token in each training context.
The excess loss $E_T(W):=\Loss_T(W)-\ell_T^\star$ therefore vanishes, and the training loss converges to the value $\ell_T^\star$ determined by the preprocessing phase.

The normalized margin result connects Muon's polar updates to attention score separation relative to input sensitivity.
The spectral norm measures the largest amplification of changes in the query representation by the attention parameter matrix:
\begin{equation*}
  \sup_{\Vert\delta q\Vert_2\le1}\Vert W\delta q\Vert_2=\Vert W\Vert_{\op}  \le\Vert W\Vert_{\mathrm F}.
\end{equation*}
With the contextual token representations fixed, this amplification directly controls how changes in the query affect attention scores.
Theorem \ref{thm:main} shows that the minimum score gap between the correct token and other tokens across the training contexts, normalized by this amplification, approaches its largest possible value.

\begin{remark}\label{rem:power-law-margin-rates}
Our analysis also covers main phase step sizes $\eta_{2,t}=\eta_{2,0}(t+1)^{-\alpha}$ with $0<\alpha\le1$, for each fixed $\alpha$ and sufficiently small $\eta_{2,0}$.
Among these schedules, $\alpha=1/2$ gives the fastest asymptotic order of the upper bound on the normalized margin gap, namely $O((1+\log t)/\sqrt{t})$.
These rate bounds support using square root step size decay in the main phase.
Performance at a fixed iteration also depends on the initial step size and bound constants.
\end{remark}

\begin{figure*}[!t]
\centering
\includegraphics[width=0.98\textwidth]{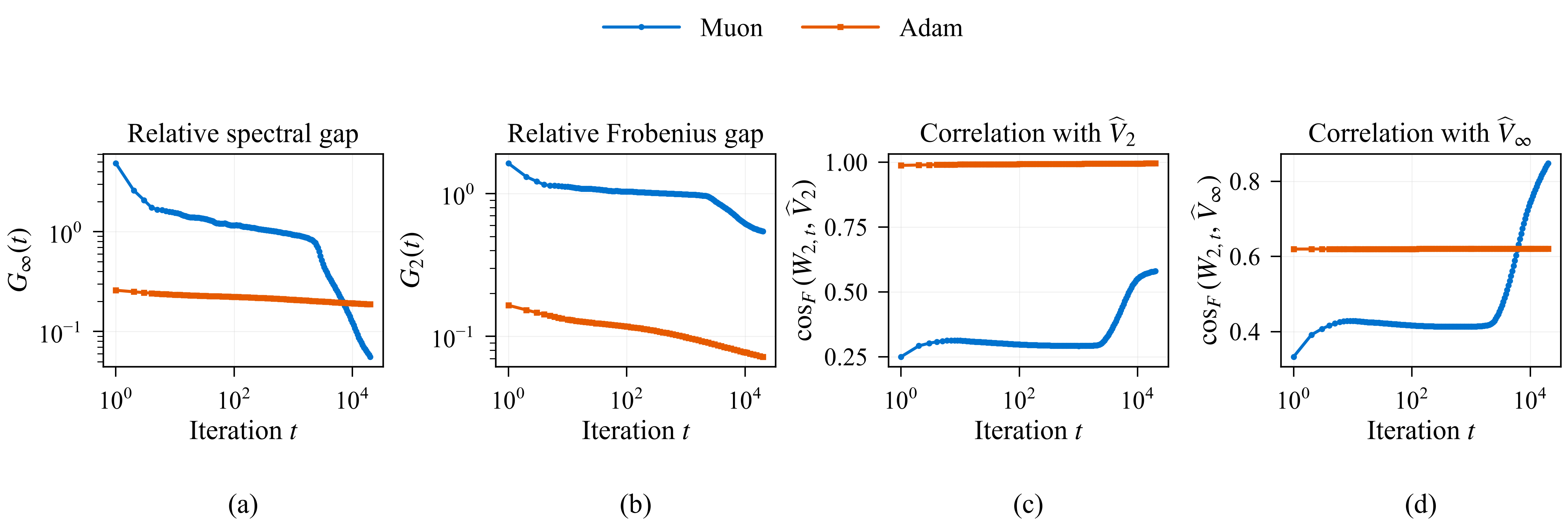}
\caption{Normalized margin behavior of Muon and Adam on the synthetic task.
Subfigures (a) and (b) show the relative spectral and Frobenius margin gaps $G_\infty(t)$ and $G_2(t)$, respectively.
Subfigures (c) and (d) show the Frobenius cosine similarities to the unique Frobenius hard margin solution $\widehat V_2$ and a selected spectral hard margin solution $\widehat V_\infty$, respectively.
Each curve shows the mean over ten independent runs.}
\label{fig:numerical-muon-adam-gaps}
\end{figure*}

\paragraph{Prediction on Unseen Contexts.}
We next show how the attention score orderings learned during training can support prediction on unseen contexts.
For a fixed query $q$, the score difference $(u-v)^\top W_2q$ does not depend on the other tokens in the context.
Thus, once the model assigns $u$ a strictly larger attention score than $v$, the same ordering holds in any new context containing both tokens with query $q$.

We write $u\to_q v$ if some training context with query $q$ has $u$ as its correct token and also contains $v\ne u$.
Once learned, these orderings yield score inequalities that can be combined along finite chains with a common query, even across training contexts.

For an unseen sample $(X,x_{L+1})$, let $X=[x_1,\ldots,x_L]$, $q=x_L$, and $x_X^\star:=h^{-1}(I(x_{L+1}))$.
We consider samples for which $x_X^\star$ appears in $X$ and
\begin{equation}\label{eq:unseen-context-order}
  x_X^\star\to_q\cdots\to_q x_\ell
  \quad(\ell\in[L],\ x_\ell\ne x_X^\star).
\end{equation}
Each chain consists of finitely many steps $u\to_q v$ arising from training contexts with query $q$.
Its intermediate tokens need not appear in $X$.

\begin{theorem}\label{thm:generalization}
Let the iterates be as in Theorem \ref{thm:main}.
Suppose that an unseen sample $(X,x_{L+1})$ satisfies \eqref{eq:unseen-context-order} and that its context contains the correct token $x_X^\star$.
Then, for all sufficiently large $t$,
\begin{equation}\label{eq:generalization-probability-lower-bound}
  \begin{aligned}
    &\mathbb P_{(W_{1,T},W_{2,t})}\{x_{L+1}\mid X\}\\
    &\qquad\ge e^{-\ell_T^\star}-\frac{\Delta_T(L-1)}{2}\,C_1t^{C_2}e^{-C_3\sqrt{t}},
  \end{aligned}
\end{equation}
where $C_1,C_2,C_3$ are the constants in \eqref{eq:thm2-loss}.
\end{theorem}

The first term $e^{-\ell_T^\star}$ is the maximum probability assigned to the correct next token with the prediction matrix $W_{1,T}$ unchanged.
It is obtained when all attention is assigned to $x_X^\star$, whose target label then has a logit gap of $\Delta_T$ over all other labels.
The second term bounds the reduction in this probability caused by attention on other tokens and vanishes as main phase training proceeds.
For fixed $T$, the prediction probability therefore converges to $e^{-\ell_T^\star}$ as $t\to\infty$.
Taking $T\to\infty$ afterward increases the preprocessing logit gap $\Delta_T$ without bound, so this limiting probability approaches one.
This result shows how attention can combine score orderings learned across training contexts to predict on new combinations of contextual tokens with the same query.

\paragraph{Extensions to Schatten $S_p$ Updates and Adam.}
The appendix extends the two phase analysis to Schatten $S_p$ steepest descent and Adam under the same data assumptions.
For each fixed $1\le p<\infty$, the Schatten extension uses normalized momentum steepest descent in the same $S_p$ norm in both phases.
Both phases permit any exact maximizing direction when the momentum is nonzero, and use the zero direction at zero momentum.
The Schatten $S_p$ norm is the $\ell_p$ norm of the singular values.
With sufficiently long preprocessing and a small enough Phase II initial step size, $\rho_2(W_{2,t})/\Vert W_{2,t}\Vert_{S_p}$ converges to the corresponding maximum margin with gap $O((1+\log t)/\sqrt{t})$ under square root step sizes in both phases.
\looseness=1\par

The Adam extension uses bias correction and a fixed $\varepsilon>0$ added outside the square root in the denominator.
Both phases use Adam with step sizes $\eta_{j,s}=\eta_{j,0}(s+1)^{-\alpha_j}$, where $0<\alpha_j\le 1$ for $j\in\{1,2\}$.
The prediction matrix $W_{1,T}$ remains fixed during the main phase.
With sufficiently small initial step sizes in both phases and sufficiently long preprocessing, the normalized margin $\rho_2(W_{2,t})/\Vert W_{2,t}\Vert_F$ converges to the Frobenius maximum margin value.
The margin gap is $O(1/\log t)$ for $0<\alpha_2<1$ and $O(1/\log\log t)$ for $\alpha_2=1$.
The normalized attention matrix $W_{2,t}/\Vert W_{2,t}\Vert_F$ also converges to the unique Frobenius hard margin direction.
\looseness=1\par

These extensions relate the update rule to the norm used for margin normalization: Muon leads to a spectral maximum margin limit, whereas Adam leads to a Frobenius maximum margin limit.

\begin{figure*}[!t]
\centering
\includegraphics[width=0.98\textwidth]{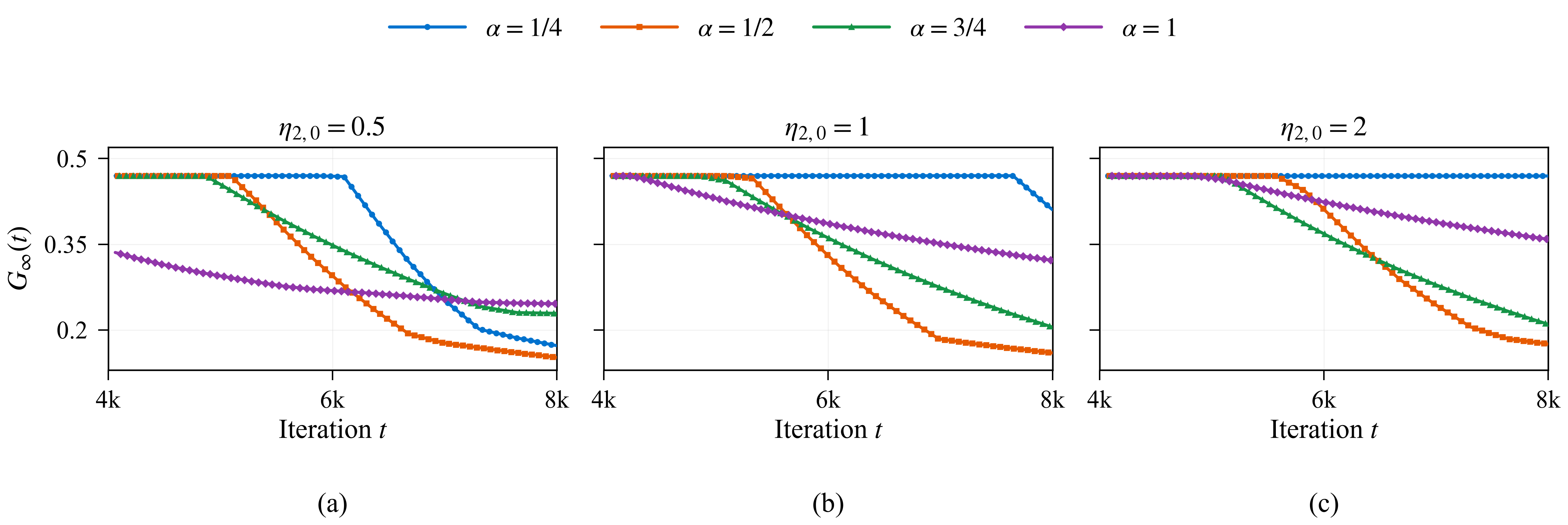}
\caption{Dependence of Muon's relative spectral margin gap $G_\infty(t)$ on the initial step size and decay exponent.
Subfigures (a), (b), and (c) use $\eta_{2,0}=0.5$, $1$, and $2$, respectively, and compare $\alpha\in\{1/4,1/2,3/4,1\}$ under the step size schedule $\eta_{2,t}=\eta_{2,0}(t+1)^{-\alpha}$.
The exponent giving the smallest observed gap varies with both $\eta_{2,0}$ and the iteration count.}
\label{fig:numerical-stepsize-balance}
\end{figure*}

\section{Numerical Experiments}
We compare the normalized margins and parameter directions of Muon and Adam on synthetic tasks and examine how Muon's step sizes affect its margin gap.

Both experiments use $K=64$ orthonormal token embeddings in $\mathbb R^{1024}$ and $N=1024$ training contexts.
We train the prediction matrix $W_{1,T}$ with Muon in Phase I and hold it fixed throughout Phase II.
All step sizes specified below are used to update the attention matrix $W_2$ during Phase II.

For the optimizer comparison in Figure \ref{fig:numerical-muon-adam-gaps}, the maximum context length is $26$.
Within each run, Muon and Adam use the same dataset and prediction matrix and start from $W_{2,0}=0$.
This allows us to compare their margin behavior under the same training objective and initialization.

Let $\gamma_2$ and $\gamma_{2,F}$ denote the spectral and Frobenius maximum margin values, respectively.
For nonzero iterates, we define the relative spectral and Frobenius margin gaps by
\begin{equation*}
  \begin{aligned}
    & G_\infty(t):=1-\frac{\rho_2(W_{2,t})}{\gamma_2\Vert W_{2,t}\Vert_{\op}},\\
    & G_2(t):=1-\frac{\rho_2(W_{2,t})}{\gamma_{2,F}\Vert W_{2,t}\Vert_F}.
  \end{aligned}
\end{equation*}
Smaller gaps indicate that the normalized margin is closer to its maximum value under the corresponding norm.

Each curve in Figure \ref{fig:numerical-muon-adam-gaps} averages ten independent runs of $20{,}000$ main phase iterations.
Muon uses momentum parameter $\beta_2=0.9$ and step size $(t+1)^{-1/2}$.
Adam uses step size $0.01(t+1)^{-1/2}$, first and second moment parameters $0.9$ and $0.999$, respectively, $\varepsilon=1.0\times10^{-6}$.

For nonzero iterates, we also compare parameter directions using the Frobenius cosine similarity
\begin{equation*}
  \cos_F(W_{2,t},\widehat V_p):=\frac{\ip{W_{2,t}}{\widehat V_p}}{\Vert W_{2,t}\Vert_F\Vert\widehat V_p\Vert_F},\qquad p\in\{2,\infty\},
\end{equation*}
where $\widehat V_2$ is the unique solution to the Frobenius hard margin problem and $\widehat V_\infty$ is a selected solution to the spectral hard margin problem.

At $t=20{,}000$, Muon has the smaller spectral margin gap, while Adam has the smaller Frobenius margin gap.
Muon also has the larger cosine similarity to $\widehat V_\infty$, and Adam to $\widehat V_2$.
These observations illustrate how the update rule affects the normalized margin and parameter direction under the same training objective.

For the step size comparison, we run each setting once for $8{,}000$ main phase iterations.
All settings use the same dataset and preprocessing output.
We use $\eta_{2,t}=\eta_{2,0}(t+1)^{-\alpha}$ with $\eta_{2,0}\in\{0.5,1,2\}$ and $\alpha\in\{1/4,1/2,3/4,1\}$.

For nonzero iterates, let
\begin{equation*}
  D(t):=\left|\rho_2\left(\frac{W_{2,t}}{\Vert W_{2,t}\Vert_{\op}}\right)-\gamma_2\right|
\end{equation*}
denote the absolute gap in the normalized spectral margin.
Since the normalized spectral margin is at most $\gamma_2$, we have $D(t)=\gamma_2G_\infty(t)$.
Figure \ref{fig:numerical-stepsize-balance} plots $G_\infty(t)$, while Table \ref{tab:raw-control-terms} lists $D(t)$.

For $t\ge 1$, let $S_t$ and $Q_t$ denote the cumulative step size and the sum of squared step sizes, respectively:
\begin{equation*}
  S_t:=\sum_{r=0}^{t-1}\eta_{2,r},\qquad Q_t:=\sum_{r=0}^{t-1}\eta_{2,r}^2.
\end{equation*}
The normalized margin bound \eqref{eq:common-normalized-margin} contains the two terms $1/S_t$ and $Q_t/S_t$.
For $0<\alpha<1$, we have $S_t=\Theta(\eta_{2,0}t^{1-\alpha})$, while $S_t=\Theta(\eta_{2,0}\log t)$ for $\alpha=1$.
The sum $Q_t$ grows polynomially for $\alpha<1/2$, logarithmically for $\alpha=1/2$, and remains bounded for $\alpha>1/2$.
Consequently, $Q_t/S_t$ dominates $1/S_t$ asymptotically for $\alpha\le 1/2$, and the two terms have the same order for $\alpha>1/2$.

Figure \ref{fig:numerical-stepsize-balance} shows that the exponent giving the smallest observed gap varies with both the initial step size and the iteration count.
At $t=8{,}000$, $\alpha=1/2$ gives the smallest gap for all three initial step sizes.
Table \ref{tab:raw-control-terms} lists $D(t)$, $1/S_t$, and $Q_t/S_t$ for $\eta_{2,0}=0.5$ at this iteration.
Here $Q_t/S_t$ is larger for $\alpha\in\{1/4,1/2\}$, while $1/S_t$ is larger for $\alpha\in\{3/4,1\}$.
Among the four exponents, $\alpha=1/2$ also gives the fastest asymptotic decay of the margin upper bound.

\begin{table}[t]
\caption{Normalized spectral margin gap $D(t)$ and the terms $1/S_t$ and $Q_t/S_t$ for $\eta_{2,0}=0.5$ at iteration $t=8{,}000$.
The last column indicates the larger of the two terms for each decay exponent.}
\centering
\small
\setlength{\tabcolsep}{2.5pt}
\begin{tabular*}{\columnwidth}{
@{\extracolsep{\fill}}ccccc@{}}
\toprule
$\alpha$
& $D(t)$
& $1/S_t$
& $Q_t/S_t$
& Larger term \\
\midrule
$1/4$
& $0.086223$
& $0.001774$
& $0.078711$
& $Q_t/S_t$ \\

$1/2$
& $\mathbf{0.075970}$
& $0.011272$
& $0.026953$
& $Q_t/S_t$ \\

$3/4$
& $0.114558$
& $0.058158$
& $0.037658$
& $1/S_t$ \\

$1$
& $0.122928$
& $0.209107$
& $0.085985$
& $1/S_t$ \\
\bottomrule
\end{tabular*}

\label{tab:raw-control-terms}
\end{table}

\section{Related Work}
\paragraph{Transformers and NTP.}
Prior works interpret Transformer forward computation as implementing gradient descent, preconditioned gradient descent, and mirror descent \cite{akyurek2022learning,von2023transformers,ahn2023transformers,d2026transformers}. 
Other analyses study retrieval and composition in next token prediction, maximum margin token selection, and the directional convergence of self attention \cite{li2024mechanics, tarzanagh2023transformers,ataee2023max,vasudeva2024implicit}.
Related work examines implicit bias in linear NTP and one layer softmax attention, sign gradient descent on two layer Transformers, regular language recognition, Markov structure, representation geometry, and next token prediction capacity \cite{thrampoulidis2024implicit,sheen2024implicit,li2025optimization,huang2025transformers,ildiz2024self,zhao2024implicit,madden2025next}.
Softmax attention has also been studied under mirror descent, preconditioned optimization, and multihead dynamics \cite{julistiono2026mirror,goel2026training,chen2024training}, while prior work has analyzed next-token prediction \cite{huang2024non}.

\paragraph{Muon.}
Muon approximates the compact polar factor of momentum matrices using Newton-Schulz iterations and has been scaled to large language model training through adjusted update scales and weight decay \cite{jordan2024muon,bernstein2024old,liu2025muon}. 
Recent theory studies layerwise loss decrease under spectral updates, simplified Muon in matrix factorization and linear Transformers, associative memory and Transformer feedforward modules, spectral descent in imbalanced models \cite{davis2025spectral,ma2026preconditioning,wang2025muon,vasudeva2025muon}.
Other work provides convergence guarantees, spectral norm constrained interpretations, and broader non Euclidean formulations \cite{shen2025convergence,yang2026convergence,chen2025muonspectralconstraints,crawshaw2025noneuclideanmuon}.
These results do not characterize Muon's implicit bias under nonlinear softmax attention.

\paragraph{Maximum Margin Theory.}
Classical work establishes directional convergence \cite{soudry2018implicit,nacson2019convergence} and develops primal-dual analyses of maximum margin solutions \cite{ji2021characterizing}.
Later studies derive faster margin convergence rates \cite{ji2021fast,wang2023faster} and generalized margin results under broader optimization geometries \cite{sun2022mirror,sun2023unified}.
Extensions to homogeneous networks yield KKT characterizations of normalized limit points \cite{lyu2019gradient,HUI2025131367}, while analyses of separable multiclass linear classification establish directional convergence under suitable assumptions \cite{ravi2024implicit}.
For matrix parameters, recent work studies spectral maximum margin convergence for spectral descent and exact SVD Muon, norm dependent KKT characterizations for Adam and Muon, and optimizer dependent simplicity bias in deep linear networks \cite{fan2026implicit,gronich2026implicit,dragutinovic2026use}.

\textit{In comparison,} we analyze Muon's training dynamics in a Transformer with nonlinear softmax attention for next-token prediction. 
Our analysis accounts for the interaction of nonlinear attention with momentum and polar updates. 
We characterize the Phase I trajectory exactly, establish nonasymptotic bounds for the training loss and normalized margin gap in Phase II, and derive finite iteration lower bounds on the probability of the correct next token.

\section{Conclusion and Future Work}
Our analysis characterizes Muon's two phase training in a Transformer with nonlinear softmax attention.
After preprocessing, the output value matrix is held fixed while Phase II drives the excess cross entropy to zero and the normalized key query margin to the spectral maximum margin value.
For contexts covered by our guarantees, Phase I determines the limiting probability of the correct next token.

Among the power law schedules considered, $\alpha=1/2$ gives the fastest asymptotic order in the proved normalized margin bound.
Experiments also illustrate how optimizer choice and step size schedules affect normalized margins.

Future work includes joint training of the two matrices, extensions to deeper and multihead Transformers, and analysis with finite Newton--Schulz iterations, stochastic gradients, and weaker NTP data assumptions.

\section*{Acknowledgements}
This work was supported by NSFC under grant number 12671672.

\clearpage
\bibliographystyle{icml2026}
\bibliography{muon_ntp_references.bib}
\onecolumn
\appendix
\appendixcontents
\enableappendixcontents
\section{Additional experiments}\label{app:additional-experiments}
We supplement the main experiments by varying the embedding dimension, vocabulary nonorthogonality, and Muon's momentum parameter.
The first two experiments compare the relative margin gaps of Muon and Adam under changes to the token embeddings.
The third examines how momentum affects the decay of Muon's spectral margin gap.
The Phase II update rules and the definitions of $G_\infty(t)$ and $G_2(t)$ are the same as in the main experiments.
The optimizer settings are given below.

\paragraph{Ambient embedding dimension.}
We vary the ambient embedding dimension $d\in\{64,128,256,512,1024,2048\}$ while keeping the vocabulary size, the token indices in each context, and the labels fixed.
Muon and adam use the same optimizer settings at every dimension.

\begin{figure*}[!htbp]
\centering
\includegraphics[width=0.98\textwidth]{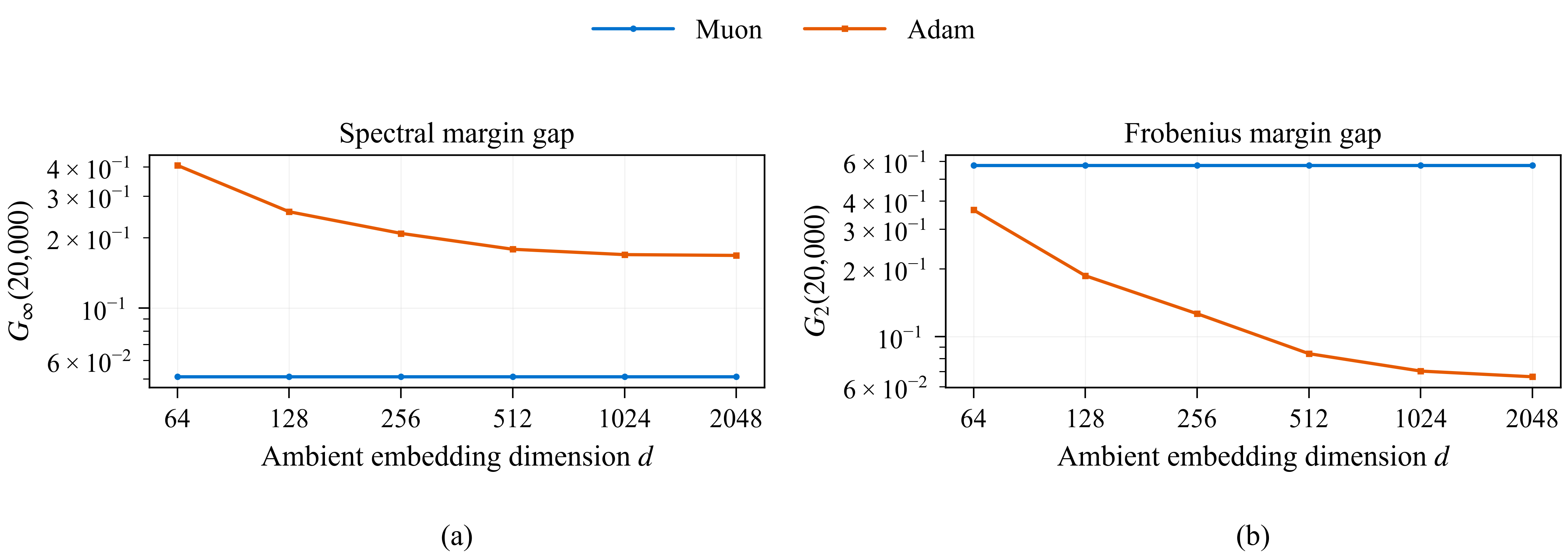}
\caption{Effect of the ambient embedding dimension.
Subfigures (a) and (b) show the relative spectral and Frobenius margin gaps at $t=20{,}000$, respectively.
At every dimension, Muon has the smaller spectral margin gap, while Adam has the smaller Frobenius margin gap.
Both gaps remain unchanged for Muon and decrease for Adam as $d$ increases.}
\label{fig:additional-ambient-dimension}
\end{figure*}

Figure \ref{fig:additional-ambient-dimension} shows the relative spectral and Frobenius margin gaps at $t=20{,}000$.
At every dimension, Muon has a smaller spectral margin gap than Adam, while Adam has a smaller Frobenius margin gap.
Both gaps for Muon remain unchanged across dimensions, consistent with the invariance of its updates under isometric embeddings of the vocabulary span when the vocabulary is orthonormal and the initialization is zero.
Both gaps for Adam decrease as $d$ increases over the tested values.
These results show that Muon's smaller spectral margin gap is not specific to the embedding dimension.

\paragraph{Nonorthogonal vocabularies.}
Our theoretical analysis assumes an orthonormal vocabulary.
We examine sensitivity to this assumption by perturbing the token embeddings while keeping the token indices, context structures, query indices, and labels fixed.
Let $x_1,\ldots,x_K$ be the original orthonormal token embeddings.
We generate a random subspace $\mathcal P\subset\operatorname{span}(\V)^\perp$ of dimension $16$ and sample a unit vector $\xi_j\in\mathcal P$ for each $j\in[K]$.
We keep this subspace and these vectors fixed across all perturbation strengths.
For each $\sigma\in\{0,0.1,0.2,0.5\}$, we define
\begin{equation*}
  \widetilde x_j^{(\sigma)}:=\sqrt{1-\sigma^2}\,x_j+\sigma\xi_j.
\end{equation*}
Since $\xi_j$ is orthogonal to every original token embedding, we have
\begin{equation*}
  \begin{aligned}
    & \left\Vert\widetilde x_j^{(\sigma)}\right\Vert_2^2=(1-\sigma^2)+\sigma^2=1,\\
    & \left\langle\widetilde x_i^{(\sigma)},\widetilde x_j^{(\sigma)}\right\rangle=\sigma^2\langle\xi_i,\xi_j\rangle\qquad(i\ne j).
  \end{aligned}
\end{equation*}
Thus the embeddings remain unit vectors, and the magnitudes of the inner products between distinct tokens scale as $\sigma^2$.

For each perturbed vocabulary, we construct a fixed prediction matrix $W_{1,T}^{(\sigma)}$ satisfying
\begin{equation*}
  W_{1,T}^{(\sigma)}\widetilde x_j^{(\sigma)}=\Delta_T\left(e_{h(x_j)}-K^{-1}\mathbf1\right),\qquad j\in[K].
\end{equation*}
Each token therefore has the same prediction logits as after preprocessing on the original vocabulary.
Muon and Adam use this same prediction matrix throughout Phase II, with their respective optimizer settings unchanged across perturbation strengths.
For each vocabulary, we recompute the spectral and Frobenius maximum margin values used in $G_\infty(t)$ and $G_2(t)$.

\begin{figure*}[t]
\centering
\includegraphics[width=0.98\textwidth]{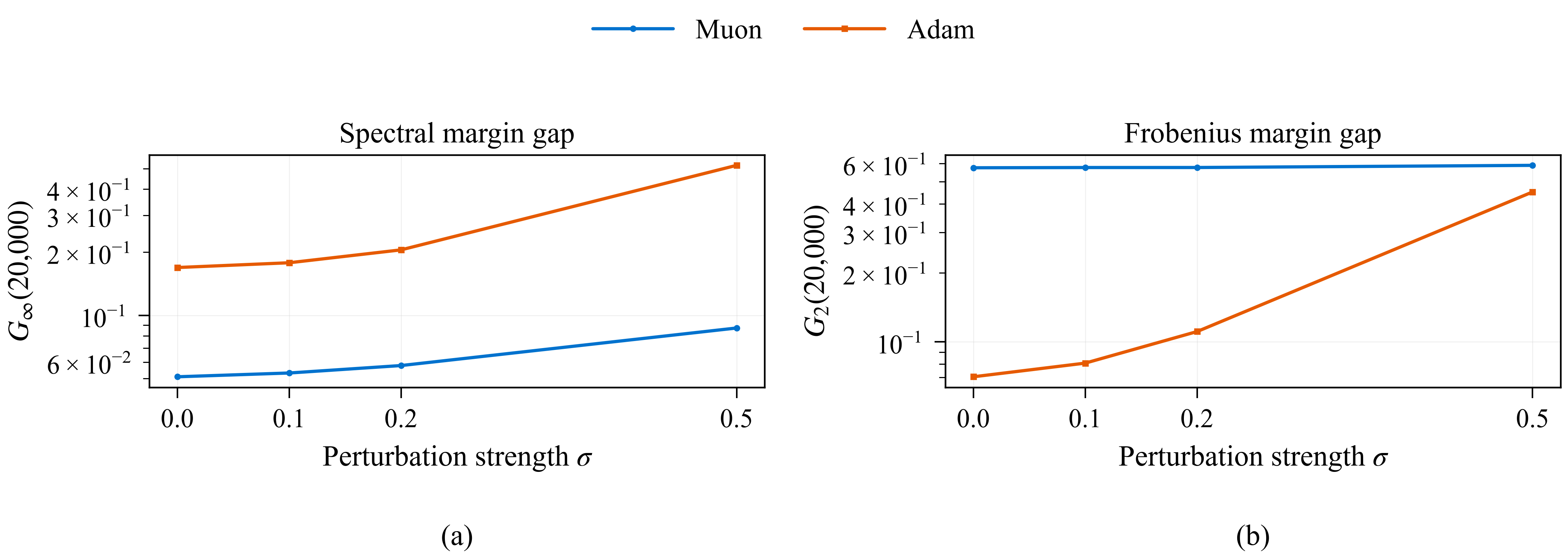}
\caption{Effect of vocabulary nonorthogonality.
Subfigures (a) and (b) show the relative spectral and Frobenius margin gaps at $t=20{,}000$, respectively, for different perturbation strengths $\sigma$.
At every perturbation strength, Muon has the smaller spectral margin gap, while Adam has the smaller Frobenius margin gap.}
\label{fig:additional-nonorthogonal-vocabulary}
\end{figure*}

Figure \ref{fig:additional-nonorthogonal-vocabulary} shows the relative spectral and Frobenius margin gaps at $t=20{,}000$.
At every perturbation strength, Muon has a smaller spectral margin gap than Adam, while Adam has a smaller Frobenius margin gap.
As $\sigma$ increases from $0$ to $0.5$, Muon's spectral margin gap increases from approximately $0.051$ to $0.087$.
Adam's Frobenius margin gap increases from approximately $0.070$ to $0.451$ over the same range.

\begin{figure*}[t]
\centering
\includegraphics[width=0.98\textwidth]{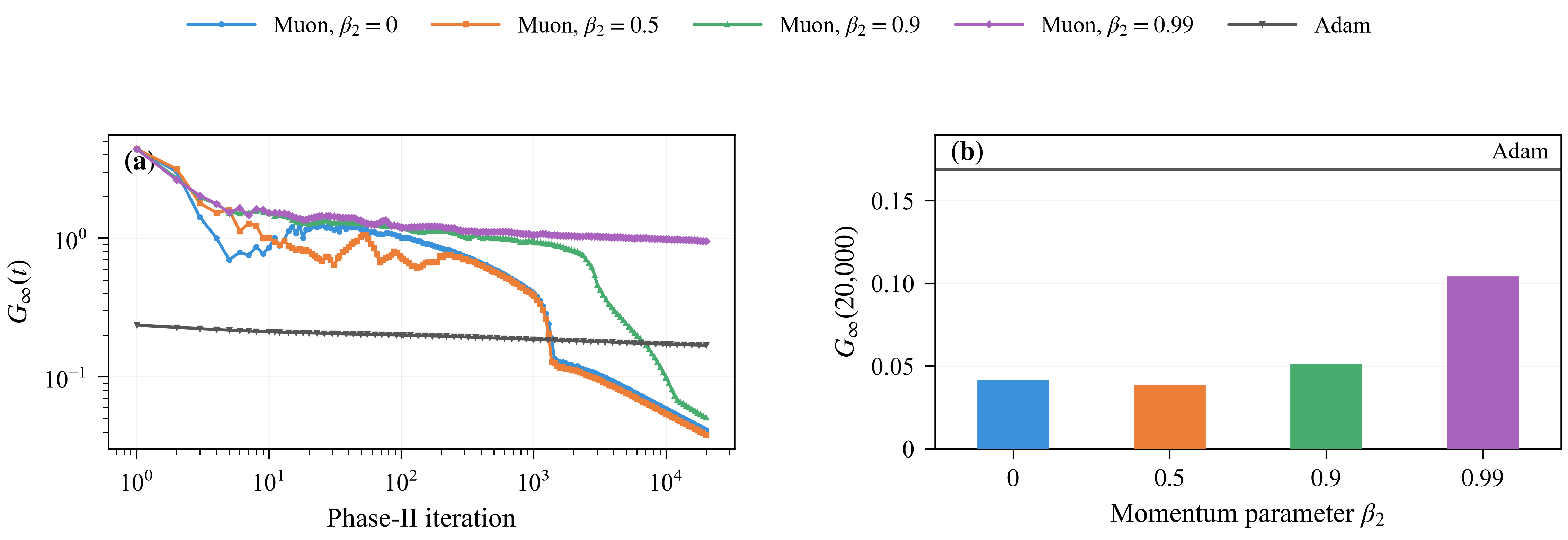}
\caption{Effect of Muon's momentum parameter.
Subfigure (a) shows the relative spectral margin gap over the first $20{,}000$ Phase II iterations for different values of $\beta_2$, together with the Adam trajectory.
Subfigure (b) shows the gap at $t=100{,}000$ for $\beta_2=0.99$ and at $t=20{,}000$ for the other momentum values.
The horizontal line in (b) shows Adam's gap at $t=20{,}000$.
The spectral margin gap decreases more slowly for $\beta_2=0.99$ over the tested iteration range.}
\label{fig:additional-momentum}
\end{figure*}

\paragraph{Momentum.}
We next vary Muon's momentum parameter over $\beta_2\in\{0,0.5,0.9,0.99\}$ while keeping the remaining optimizer settings fixed.
Figure \ref{fig:additional-momentum}(a) compares the relative spectral margin gap over the first $20{,}000$ Phase II iterations.
The values $\beta_2\in\{0,0.5,0.9\}$ give similar final gaps, although their trajectories differ.
For $\beta_2=0.99$, the gap decreases more slowly and remains approximately $0.951$ at $t=20{,}000$.

Figure \ref{fig:additional-momentum}(b) shows the gap at $t=100{,}000$ for $\beta_2=0.99$ and at $t=20{,}000$ for the other tested momentum values.
At $t=100{,}000$, the gap for $\beta_2=0.99$ is approximately $0.104$, still larger than the gaps of approximately $0.041$, $0.038$, and $0.051$ reached by the other three values at $t=20{,}000$.
Thus $\beta_2=0.99$ gives slower convergence over the tested iteration range.

Under the data, preprocessing, and step size conditions of Theorem \ref{thm:main}, the normalized spectral margin converges to $\gamma_2$ for every fixed $\beta_2\in[0,1)$.
The convergence constants and the upper bound on the initial step size may depend on $\beta_2$, so convergence over a fixed number of iterations can differ across momentum values.

\section{Proofs of the main results}\label{app:main-proofs}
We first prove Theorem \ref{thm:phase1} for the preprocessing phase.
For the main phase, we derive loss and margin bounds for a class of matrix norms under power law step sizes.
We then specialize these bounds to the spectral norm to prove Theorem \ref{thm:main}.
Finally, we use the training margin bound to prove Theorem \ref{thm:generalization}.

\subsection{Preprocessing: proof of Theorem \ref{thm:phase1}}\label{app:preprocessing-proof}
We first characterize the preprocessing iterates generated by Algorithm \ref{alg:two-phase-muon} for an arbitrary positive step size sequence.
Theorem \ref{thm:phase1} then follows by choosing the square root step size schedule.

In the preprocessing phase, each context consists of a single token.
Write $\V=\{x_1,\ldots,x_K\}$, and let $y_i:=h(x_i)$ denote the target label for $[x_i]$.
Define
\begin{equation*}
  \begin{aligned}
    & X_{\V}:=[x_1,\ldots,x_K]\in\R^{d\times K},\qquad && P:=[e_{y_1},\ldots,e_{y_K}]\in\R^{K\times K},\\
    & C:=I_K-\frac1K\mathbf1\mathbf1^\top,\qquad && B:=PCX_{\V}^\top.
  \end{aligned}
\end{equation*}
Here $C$ centers the output logits.

Since $h$ is bijective, $P$ is a permutation matrix.
Together with the orthonormality of the vocabulary embeddings and the definition of $C$, this gives
\begin{equation*}
  X_{\V}^\top X_{\V}=I_K,\qquad P\mathbf1=\mathbf1,\qquad PC=CP.
\end{equation*}
The next lemma expresses $W_{1,s}$ as $B$ multiplied by the cumulative step size.

\begin{lemma}\label{lem:exact-U}
Consider the preprocessing iterates generated by Algorithm \ref{alg:two-phase-muon}, using the compact polar factor and an arbitrary positive step size sequence.
Let
\begin{equation*}
  \Delta_s:=\sum_{j=0}^{s-1}\eta_{1,j},\qquad \Delta_0:=0.
\end{equation*}
For every integer $s\ge0$, the following statements hold.
\begin{enumerate}
  \item[\textup{(i)}] The iterates and their logits satisfy
  \begin{align}
    & W_{1,s}=CW_{1,s}=\Delta_sB,\label{eq:phase1-exact-trajectory}\\
    & W_{1,s}x_i=\Delta_s\left(e_{y_i}-\frac1K\mathbf1\right),\qquad i\in[K].\label{eq:phase1-exact-token-logits}
  \end{align}

  \item[\textup{(ii)}] The loss, margin, and operator norm satisfy
  \begin{equation*}
    \begin{aligned}
      & \Loss_0(W_{1,s})=\log\left(1+(K-1)e^{-\Delta_s}\right),\\
      & \rho_1(W_{1,s})=\Vert W_{1,s}\Vert_{\op}=\Delta_s.
    \end{aligned}
  \end{equation*}
  Moreover, the spectral hard margin problem in \eqref{eq:hard-margin-ov} has optimal value $R_1^\star=1$, and hence $\gamma_1=1$.
\end{enumerate}
\end{lemma}

\begin{proof}
Orthonormality, $PC=CP$, and $C^2=C$ give
\begin{equation*}
  BB^\top=PCX_{\V}^\top X_{\V}CP^\top=PCP^\top=C,\qquad CB=B.
\end{equation*}
Since $C$ is the orthogonal projection onto $\mathbf1^\perp$, the identity $BB^\top=C$ implies that $B$ has $K-1$ nonzero singular values, all equal to one.
Hence $\Vert B\Vert_{\op}=1$, $\Vert B\Vert_{\ast}=K-1$, and the compact polar factor of $B$ is $B$.
Moreover,
\begin{equation}\label{eq:phase1-B-token-action}
  Bx_i=PCe_i=e_{y_i}-\frac1K\mathbf1,\qquad i\in[K].
\end{equation}

\textup{(i)} We prove by induction that $W_{1,s}=\Delta_sB$ and $M_{1,s-1}=-\mu_{s-1}B$ for some $\mu_{s-1}\ge0$.
Let $\mu_{-1}=0$.
Both identities hold at $s=0$ by zero initialization.

Suppose they hold at iteration $s$, and let $Z_s:=e^{\Delta_s}+K-1$.
By the induction hypothesis, \eqref{eq:phase1-B-token-action}, and shift invariance of softmax, the probability of $y_i$ is $e^{\Delta_s}/Z_s$ and that of each other label is $1/Z_s$.
Using the uniform preprocessing weights and the softmax cross entropy gradient, we obtain
\begin{equation*}
  \begin{aligned}
    & \softmax(W_{1,s}x_i)-e_{y_i}=-\frac K{Z_s}\left(e_{y_i}-\frac1K\mathbf1\right),\\
    & \nabla\Loss_0(W_{1,s})=\frac1K\sum_{i=1}^K\left(\softmax(W_{1,s}x_i)-e_{y_i}\right)x_i^\top=-Z_s^{-1}B.
  \end{aligned}
\end{equation*}
The momentum update therefore gives
\begin{equation*}
  M_{1,s}=-\mu_sB,\qquad \mu_s:=\beta_1\mu_{s-1}+(1-\beta_1)Z_s^{-1}.
\end{equation*}
Since $\mu_{s-1}\ge0$, $\beta_1\in[0,1)$, and $Z_s>0$, we have $\mu_s>0$.
Thus the compact polar factor of $M_{1,s}$ is $-B$, and the parameter update becomes
\begin{equation*}
  W_{1,s+1}=W_{1,s}+\eta_{1,s}B=(\Delta_s+\eta_{1,s})B=\Delta_{s+1}B.
\end{equation*}
Both induction identities therefore hold at iteration $s+1$.
Combining $W_{1,s}=\Delta_sB$ with $CB=B$ and \eqref{eq:phase1-B-token-action} proves \eqref{eq:phase1-exact-trajectory} and \eqref{eq:phase1-exact-token-logits}.

\textup{(ii)} The probability assigned to the correct label is the same for every preprocessing sample, so
\begin{equation*}
  \Loss_0(W_{1,s})=-\log\frac{e^{\Delta_s}}{e^{\Delta_s}+K-1}=\log\left(1+(K-1)e^{-\Delta_s}\right).
\end{equation*}
Equation \eqref{eq:phase1-exact-token-logits} gives
\begin{equation*}
  (e_{y_i}-e_v)^\top W_{1,s}x_i=\Delta_s,\qquad i\in[K],\quad v\in[K]\setminus\{y_i\}.
\end{equation*}
Hence $\rho_1(W_{1,s})=\Delta_s$.
Since $\Delta_s\ge0$ and $\Vert B\Vert_{\op}=1$, we also have $\Vert W_{1,s}\Vert_{\op}=\Delta_s$.

It remains to determine the optimal value of \eqref{eq:hard-margin-ov}.
For $i\in[K]$ and $v\in[K]\setminus\{y_i\}$, let $C_{x_i,v}:=(e_{y_i}-e_v)x_i^\top$.
Equation \eqref{eq:phase1-B-token-action} gives $\ip{B}{C_{x_i,v}}=1$, so $B$ is feasible.
Since $\Vert B\Vert_{\op}=1$, this yields $R_1^\star\le 1$.
To obtain the reverse inequality, note that
\begin{equation*}
  \sum_{i=1}^K\sum_{v\in[K]\setminus\{y_i\}}C_{x_i,v}=\sum_{i=1}^K(Ke_{y_i}-\mathbf1)x_i^\top=KB.
\end{equation*}
For any feasible matrix $U$, summing its constraints and using duality of the operator and nuclear norms gives
\begin{equation*}
    K(K-1) \le \ip{U}{\sum_{i=1}^K\sum_{v\in[K]\setminus\{y_i\}}C_{x_i,v}}
    = K\ip{U}{B}
    \le K\Vert U\Vert_{\op}\Vert B\Vert_{\ast}
    = K(K-1)\Vert U\Vert_{\op}.
\end{equation*}
Since $K\ge2$, every feasible matrix has operator norm at least one.
Therefore $R_1^\star=1$ and $\gamma_1=1$.
\end{proof}

\begin{proof}[Proof of Theorem \ref{thm:phase1}]
Lemma \ref{lem:exact-U} gives \eqref{eq:thm1-logit-gap}, the exact loss identity in \eqref{eq:thm1-loss}, and $R_1^\star=\gamma_1=1$.
For $\eta_{1,s}=\eta_{1,0}(s+1)^{-1/2}$ and every integer $T\ge1$, we have
\begin{equation*}
  \Delta_T=\eta_{1,0}\sum_{k=1}^T k^{-1/2}\ge\eta_{1,0}\sqrt{T},
\end{equation*}
where the inequality follows from $k^{-1/2}\ge T^{-1/2}$ for $1\le k\le T$.
Combining the exact loss identity with $\log(1+u)\le u$ for $u\ge0$ yields
\begin{equation*}
  \Loss_0(W_{1,T})\le(K-1)e^{-\Delta_T}\le(K-1)e^{-\eta_{1,0}\sqrt{T}}.
\end{equation*}
Finally, $\rho_1(W_{1,T})=\Vert W_{1,T}\Vert_{\op}=\Delta_T>0$, so positive homogeneity of $\rho_1$ gives
\begin{equation*}
  \rho_1\left(\frac{W_{1,T}}{\Vert W_{1,T}\Vert_{\op}}\right)=\frac{\rho_1(W_{1,T})}{\Vert W_{1,T}\Vert_{\op}}=1=\gamma_1.
\end{equation*}
This proves \eqref{eq:thm1-margin}.
\end{proof}

\subsection{Main phase: proof of Theorem \ref{thm:main}}\label{sec:phase2}

\paragraph{Proof overview.}
With $W_{1,T}$ fixed, we express the loss and gradient in terms of the total attention weight of each token.
Sufficient preprocessing ensures that the correct token's total attention weight is uniformly bounded away from zero on the initial loss sublevel set.
For a sufficiently small initial step size, gradient and momentum estimates yield a descent recurrence that keeps all iterates in this set and makes the excess losses summable when weighted by the step sizes.
This summability controls the accumulated error from the sharper gradient lower bound, whose leading coefficient is $\gamma_2$, and yields the excess loss decay estimate.
Finally, we use the relation between excess loss and attention score gaps to bound the training margin from below; combining this with an upper bound on $\Vert W_{2,t}\Vert_{\op}$ gives the normalized spectral margin rate.

\subsubsection*{\textbf{Attention loss and separability}}

\paragraph{Output identities and token orderings.}
Fix an integer $T\ge1$ and keep $W_{1,T}$ fixed throughout Phase II.
Lemma \ref{lem:exact-U} gives
\begin{equation*}
  W_{1,T}=CW_{1,T}=\Delta_TB,\qquad \Delta_T>0.
\end{equation*}
Recall that the preprocessing loss satisfies
\begin{equation*}
  \ell_T^\star=\Loss_0(W_{1,T})=\log\left(1+(K-1)e^{-\Delta_T}\right).
\end{equation*}

To record the required attention score inequalities, let
\begin{equation*}
  \begin{aligned}
    & \I_2:=\{(n,x):n\in[N],\ x\in\X_n^-\},\\
    & \mathcal C_q:=\{(x_n^\star,x):(n,x)\in\I_2,\ q_n=q\},\qquad q\in\V.
  \end{aligned}
\end{equation*}
Thus $(u,v)\in\mathcal C_q$ exactly when $u\to_q v$ in the notation of the main text.
The next lemma characterizes separability in terms of token orderings, which will also be used to establish a gradient lower bound.

\begin{lemma}\label{lem:phase2-separability}
Suppose that the vocabulary is orthonormal and $\I_2\ne\varnothing$.
Assumption \ref{ass:separability} holds if and only if, for every $q\in\V$, there is a permutation $u_{q,1},\ldots,u_{q,K}$ of $\V$ such that
\begin{equation*}
  (u_{q,i},u_{q,j})\in\mathcal C_q\quad\Longrightarrow\quad i<j.
\end{equation*}
Under these equivalent conditions, the minimum in \eqref{eq:hard-margin-kq} is attained, and
\begin{equation*}
  0<R_2^\star<\infty,\qquad \gamma_2>0.
\end{equation*}
\end{lemma}

\begin{proof}
Suppose first that Assumption \ref{ass:separability} holds, and let $W$ be a corresponding separating matrix.
For each $q\in\V$, list the tokens in nonincreasing order of $u^\top Wq$, breaking ties arbitrarily.
For every $(u,v)\in\mathcal C_q$, separability gives $(u-v)^\top Wq\ge1$, so $u$ precedes $v$.

Conversely, suppose that such a permutation exists for every $q\in\V$, and let
\begin{equation*}
  v_q:=\sum_{i=1}^K(K-i)u_{q,i},\qquad W^\star:=\sum_{q\in\V}v_qq^\top.
\end{equation*}
Orthonormality gives, for every $q\in\V$ and $i\in[K]$,
\begin{equation*}
  W^\star q=v_q,\qquad u_{q,i}^\top v_q=K-i.
\end{equation*}
For every $(u_{q,i},u_{q,j})\in\mathcal C_q$, the ordering condition gives $i<j$, so
\begin{equation*}
  (u_{q,i}-u_{q,j})^\top W^\star q=j-i\ge1.
\end{equation*}
Thus $W^\star$ is feasible for \eqref{eq:hard-margin-kq}.

For each $(n,x)\in\I_2$, let $A_{n,x}:=(x_n^\star-x)q_n^\top$.
The rank one norm identity and orthonormality give
\begin{equation*}
  \Vert A_{n,x}\Vert_{\ast}=\Vert x_n^\star-x\Vert_2\Vert q_n\Vert_2=\sqrt{2}.
\end{equation*}
Since $\I_2\ne\varnothing$, fix $(n,x)\in\I_2$.
For every feasible $W$, duality of the operator and nuclear norms yields
\begin{equation*}
  1\le \ip{W}{A_{n,x}}\le \Vert W\Vert_{\op}\Vert A_{n,x}\Vert_{\ast}=\sqrt{2}\Vert W\Vert_{\op}.
\end{equation*}

Intersecting the closed feasible set with $\{W:\Vert W\Vert_{\op}\le \Vert W^\star\Vert_{\op}\}$ gives a nonempty compact set.
Since every feasible matrix outside this set has larger norm than $W^\star$, minimizing the continuous operator norm on this set also solves \eqref{eq:hard-margin-kq}.
Consequently,
\begin{equation*}
  \frac1{\sqrt{2}}\le R_2^\star\le \Vert W^\star\Vert_{\op}<\infty,\qquad \gamma_2=\frac1{R_2^\star}>0.
\end{equation*}
\end{proof}

\paragraph{Attention and loss identities.}
Fix the prediction matrix $W_{1,T}$ satisfying $CW_{1,T}=\Delta_TB$ with $\Delta_T>0$.
For $W\in\R^{d\times d}$, the total attention weight over all occurrences of token $x\in\V$ in sample $n\in[N]$ is
\begin{equation}\label{eq:phase2-grouped-attention-mass}
  a_{n,x}(W):=\frac{c_{n,x}e^{x^\top Wq_n}}{\sum_{z\in\V}c_{n,z}e^{z^\top Wq_n}}.
\end{equation}
The denominator is positive because the context is nonempty.
These weights are nonnegative, sum to one, and vanish whenever $c_{n,x}=0$.
The next lemma expresses the prediction probabilities and losses in terms of these weights and identifies the infimum of $\Loss_T$.

\begin{lemma}\label{lem:phase2-attention-loss-identities}
Suppose that Assumption \ref{ass:token-label} holds.
Then the following statements hold.

\begin{enumerate}
  \item[\textup{(i)}] For every $W\in\R^{d\times d}$, $n\in[N]$, and $x\in\V$, the probability assigned to label $h(x)$ is
  \begin{equation*}
    p_{n,x}(W):=e_{h(x)}^\top f_{(W_{1,T},W)}(X^{(n)})=\frac{e^{\Delta_Ta_{n,x}(W)}}{\sum_{z\in\V}e^{\Delta_Ta_{n,z}(W)}}.
  \end{equation*}
  The corresponding sample loss satisfies
  \begin{equation*}
    \ell_{T,n}(W):=-\log p_{n,x_n^\star}(W)=\log\left(1+\sum_{x\in\V\setminus\{x_n^\star\}}e^{-\Delta_T(a_{n,x_n^\star}(W)-a_{n,x}(W))}\right).
  \end{equation*}
  Moreover, for every $x\in\X_n^-$,
  \begin{equation}\label{eq:attn-odds-margin-bridge}
    \frac{a_{n,x}(W)}{a_{n,x_n^\star}(W)}=\frac{c_{n,x}}{c_{n,x_n^\star}}\exp\left(-\ip{W}{A_{n,x}}\right).
  \end{equation}

  \item[\textup{(ii)}] For every $W\in\R^{d\times d}$ and $n\in[N]$,
  \begin{equation*}
    \ell_{T,n}(W)\ge \ell_T^\star,\qquad \Loss_T(W)\ge \ell_T^\star.
  \end{equation*}
  If Assumption \ref{ass:separability} also holds, then
  \begin{equation*}
    \inf_{W\in\R^{d\times d}}\Loss_T(W)=\ell_T^\star.
  \end{equation*}
\end{enumerate}
\end{lemma}

\begin{proof}
\textup{(i)} The attention output for sample $n$ is $\sum_{x\in\V}a_{n,x}(W)x$.
Using $CW_{1,T}=\Delta_TB$, $Bx=e_{h(x)}-K^{-1}\mathbf1$, and $\sum_{x\in\V}a_{n,x}(W)=1$, its centered prediction logits are
\begin{equation*}
  CW_{1,T}\sum_{x\in\V}a_{n,x}(W)x=\Delta_T\sum_{x\in\V}a_{n,x}(W)e_{h(x)}-\frac{\Delta_T}{K}\mathbf1.
\end{equation*}
Centering and the common term do not affect softmax, so bijectivity of $h$ gives the stated expression for $p_{n,x}(W)$.
The sample loss formula follows from $h(x_n^\star)=y_n$.
Assumption \ref{ass:token-label} ensures $c_{n,x_n^\star}>0$ and hence $a_{n,x_n^\star}(W)>0$.
Taking the ratio in \eqref{eq:phase2-grouped-attention-mass} and using $A_{n,x}=(x_n^\star-x)q_n^\top$ proves \eqref{eq:attn-odds-margin-bridge}.

\textup{(ii)} Since the attention weights lie in $[0,1]$, every difference $a_{n,x_n^\star}(W)-a_{n,x}(W)$ is at most one.
Because $\Delta_T>0$, each of the $K-1$ exponential terms in the sample loss formula is at least $e^{-\Delta_T}$.
Therefore,
\begin{equation*}
  \ell_{T,n}(W)\ge \log\left(1+(K-1)e^{-\Delta_T}\right)=\ell_T^\star.
\end{equation*}
Averaging over the samples gives $\Loss_T(W)\ge \ell_T^\star$.

Under Assumption \ref{ass:separability}, choose a matrix $W^\star$ satisfying $\ip{W^\star}{A_{n,x}}\ge1$ for every $(n,x)\in\I_2$.
For $r\ge0$, these inequalities, \eqref{eq:attn-odds-margin-bridge}, and normalization give
\begin{equation*}
    a_{n,x_n^\star}(rW^\star) = \left(1+\sum_{x\in\X_n^-}\frac{a_{n,x}(rW^\star)}{a_{n,x_n^\star}(rW^\star)}\right)^{-1}
    \ge \left(1+e^{-r}\sum_{x\in\X_n^-}\frac{c_{n,x}}{c_{n,x_n^\star}}\right)^{-1}.
\end{equation*}
The counts are fixed and $c_{n,x_n^\star}>0$, so the lower bound above gives $a_{n,x_n^\star}(rW^\star)\to1$ as $r\to\infty$.
Nonnegativity and normalization imply that every other attention weight tends to zero.
Continuity of the sample loss formula and the finite training set therefore give
\begin{equation*}
  \Loss_T(rW^\star)=\frac{1}{N}\sum_{n=1}^N\ell_{T,n}(rW^\star)\longrightarrow\ell_T^\star.
\end{equation*}
Combining this limit with the lower bound gives $\inf_{W\in\R^{d\times d}}\Loss_T(W)=\ell_T^\star$.
\end{proof}

Define the sample excess loss by
\begin{equation*}
  E_{T,n}(W):=\ell_{T,n}(W)-\ell_T^\star.
\end{equation*}
The empirical excess loss introduced in the main text satisfies
\begin{equation*}
  E_T(W)=\frac{1}{N}\sum_{n=1}^NE_{T,n}(W)=\Loss_T(W)-\ell_T^\star.
\end{equation*}

The next lemma compares the attention weights and bounds the loss at the start of Phase II.
\begin{lemma}\label{lem:initial-barrier}
Suppose that Assumptions \ref{ass:token-label} and \ref{ass:weak-count} hold.
At $W=0$,
\begin{equation*}
    a_{n,x_n^\star}(0)\ge a_{n,x}(0),\qquad n\in[N],\quad x\in\V\setminus\{x_n^\star\},\quad \text{and}\quad 
    \Loss_T(0)\le \log K.
\end{equation*}
\end{lemma}

\begin{proof}
At $W=0$, each position receives attention weight $1/L_n$, so $a_{n,x}(0)=c_{n,x}/L_n$ for every $x\in\V$.
Assumption \ref{ass:weak-count} gives $c_{n,x_n^\star}\ge c_{n,x}$ for every $x\in\X_n^-$, and the same inequality holds when $c_{n,x}=0$.
Thus, for every $n\in[N]$ and $x\in\V\setminus\{x_n^\star\}$,
\begin{equation*}
  a_{n,x_n^\star}(0)-a_{n,x}(0)=\frac{c_{n,x_n^\star}-c_{n,x}}{L_n}\ge0.
\end{equation*}
These nonnegative differences and $\Delta_T>0$ imply that each of the $K-1$ exponential terms in the sample loss formula of Lemma \ref{lem:phase2-attention-loss-identities} is at most one.
Therefore,
\begin{equation*}
  \ell_{T,n}(0)\le \log K.
\end{equation*}
Averaging over the samples yields $\Loss_T(0)\le \log K$.
\end{proof}

With sufficient preprocessing, the correct token's total attention weight is uniformly bounded away from zero on the loss sublevel set $\Loss_T(W)\le \log K$.

\begin{lemma}\label{lem:noncollapse-entry}
Suppose that Assumption \ref{ass:token-label} holds.
For every $W\in\R^{d\times d}$ and $n\in[N]$,
\begin{equation}\label{eq:attention-sample-loss-lower-bound}
  \ell_{T,n}(W)\ge \frac{\Delta_T\left(1-Ka_{n,x_n^\star}(W)\right)}{K-1}.
\end{equation}
Let $\delta_0:=1/(2K)$.
If $\Delta_T\ge2N(K-1)\log K$, then, for every $W\in\R^{d\times d}$,
\begin{equation}\label{eq:preprocessing-initial-barrier}
  \Loss_T(W)\le \log K\quad\Longrightarrow\quad a_{n,x_n^\star}(W)\ge\delta_0\qquad(n\in[N]).
\end{equation}
\end{lemma}

\begin{proof}
Fix $W\in\R^{d\times d}$ and $n\in[N]$, and let $u:=a_{n,x_n^\star}(W)$.
The other $K-1$ attention weights sum to $1-u$, so Jensen's inequality gives
\begin{equation*}
  \sum_{x\in\V\setminus\{x_n^\star\}}e^{\Delta_Ta_{n,x}(W)}\ge (K-1)\exp\left(\frac{\Delta_T(1-u)}{K-1}\right).
\end{equation*}
Using the sample loss formula and discarding the positive term $e^{\Delta_Tu}$ inside the logarithm, we obtain
\begin{equation*}
  \begin{aligned}
    \ell_{T,n}(W) & = \log\left(e^{\Delta_Tu}+\sum_{x\in\V\setminus\{x_n^\star\}}e^{\Delta_Ta_{n,x}(W)}\right)-\Delta_Tu\\
    & \ge \log\left(\sum_{x\in\V\setminus\{x_n^\star\}}e^{\Delta_Ta_{n,x}(W)}\right)-\Delta_Tu\\
    & \ge \log(K-1)+\frac{\Delta_T(1-u)}{K-1}-\Delta_Tu\\
    & \ge \frac{\Delta_T(1-Ku)}{K-1},
  \end{aligned}
\end{equation*}
where the last inequality uses $K\ge2$ and hence $\log(K-1)\ge0$.
This proves \eqref{eq:attention-sample-loss-lower-bound}.

Now suppose that $\Loss_T(W)\le \log K$ and $\Delta_T\ge2N(K-1)\log K$.
Since the empirical loss is the uniform average of nonnegative sample losses,
\begin{equation*}
  \ell_{T,n}(W)\le N\Loss_T(W)\le N\log K.
\end{equation*}
Combining this with \eqref{eq:attention-sample-loss-lower-bound} yields
\begin{equation*}
  u\ge\frac{1}{K}\left(1-\frac{N(K-1)\log K}{\Delta_T}\right)\ge\frac{1}{2K}=\delta_0.
\end{equation*}
\end{proof}

The following corollary translates the logit gap condition in Lemma \ref{lem:noncollapse-entry} into an explicit sufficient condition on $T$.
\begin{corollary}\label{cor:preprocessing-barrier}
Suppose that Assumptions \ref{ass:token-label} and \ref{ass:weak-count} hold.
Let the preprocessing iterates be as in Theorem \ref{thm:phase1}, and let
\begin{equation*}
  \bar T:=\left(\frac{2N(K-1)\log K}{\eta_{1,0}}\right)^2.
\end{equation*}
For every integer $T>\bar T$,
\begin{equation*}
  \Delta_T\ge2N(K-1)\log K,\qquad \Loss_T(0)\le \log K,
\end{equation*}
and \eqref{eq:preprocessing-initial-barrier} holds.
\end{corollary}

\begin{proof}
For $T>\bar T$, the square root step size and the definition of $\bar T$ give
\begin{equation*}
  \Delta_T=\eta_{1,0}\sum_{k=1}^T k^{-1/2}\ge\eta_{1,0}\sqrt{T}\ge2N(K-1)\log K.
\end{equation*}
Lemma \ref{lem:initial-barrier} gives $\Loss_T(0)\le \log K$.
Applying Lemma \ref{lem:noncollapse-entry} then proves \eqref{eq:preprocessing-initial-barrier}.
\end{proof}

\paragraph{Excess loss and gradients.}
We express the sample excess loss as a function of the attention weights to derive loss bounds and gradient identities.

For each sample $n\in[N]$, define the set of admissible attention vectors by
\begin{equation*}
  \mathcal A_n:=\left\{a\in\R^{\V}:a_x\ge0\ (x\in\V),\ \sum_{x\in\V}a_x=1,\ a_x=0\ \text{whenever }c_{n,x}=0\right\}.
\end{equation*}
The attention vector $a_n(W):=\bigl(a_{n,x}(W)\bigr)_{x\in\V}$ belongs to $\mathcal A_n$.
For each $x\in\V$, let $e_x\in\R^{\V}$ denote the standard basis vector indexed by $x$.
Define the total attention weight assigned to the other tokens in the context and its empirical average by
\begin{equation*}
  s_n(W):=1-a_{n,x_n^\star}(W)=\sum_{x\in\X_n^-}a_{n,x}(W),\qquad \bar s(W):=\frac{1}{N}\sum_{n=1}^Ns_n(W).
\end{equation*}

For a fixed sample $n\in[N]$, let $c:=x_n^\star$.
For $a\in\R^{\V}$ and $x\in\V$, define
\begin{equation*}
    p_{n,x}(a):=\frac{e^{\Delta_Ta_x}}{\sum_{z\in\V}e^{\Delta_Ta_z}},\qquad
    F_{T,n}(a):=\log\left(\sum_{z\in\V}e^{\Delta_Ta_z}\right)-\Delta_Ta_c-\ell_T^\star.
\end{equation*}
These formulas define smooth functions on $\R^{\V}$, which we will evaluate on $\mathcal A_n$.
The preceding attention and loss identities give
\begin{equation*}
  p_{n,x}(W)=p_{n,x}(a_n(W)),\qquad E_{T,n}(W)=F_{T,n}(a_n(W)).
\end{equation*}
Assumption \ref{ass:token-label} ensures $e_c\in\mathcal A_n$, and the definition of $\ell_T^\star$ gives $F_{T,n}(e_c)=0$.
For the following estimates, let
\begin{equation*}
  Z_T:=e^{\Delta_T}+K-1,\qquad \kappa_T:=\frac{K\Delta_T}{Z_T}>0.
\end{equation*}

For $x\ne c$, let $D_{n,x}(a)$ denote the difference between the partial derivatives with respect to $a_x$ and $a_c$.
For $a\in\mathcal A_n$, let $H_n(a)$ be the sum of these differences weighted by the attention weights:
\begin{equation}\label{eq:phase2-grouped-derivative-differences}
  \begin{aligned}
    & D_{n,x}(a):=\langle\nabla_aF_{T,n}(a),e_x-e_c\rangle=\Delta_T\left(1+p_{n,x}(a)-p_{n,c}(a)\right),\qquad x\ne c,\\
    & H_n(a):=\sum_{x\in\V\setminus\{c\}}a_xD_{n,x}(a)=\langle\nabla_aF_{T,n}(a),a-e_c\rangle.
  \end{aligned}
\end{equation}
The inner product expression for $H_n(a)$ follows from $\sum_{x\in\V}a_x=1$ for $a\in\mathcal A_n$.

The next lemma bounds the sample excess loss and $H_n(a)$ in terms of the total attention weight assigned to the other tokens.
These bounds will be used in the gradient estimates.
\begin{lemma}\label{lem:vertex}
For every $n\in[N]$ and $a\in\mathcal A_n$, let $s_n(a):=1-a_{x_n^\star}$.
Then
\begin{equation}\label{eq:phase2-ce-attention-comparison}
  \kappa_Ts_n(a)\le F_{T,n}(a)\le H_n(a)\le 2\Delta_Ts_n(a).
\end{equation}
Consequently, for every $W\in\R^{d\times d}$,
\begin{equation*}
  \kappa_T\bar s(W)\le E_T(W)\le 2\Delta_T\bar s(W).
\end{equation*}
\end{lemma}

\begin{proof}
Fix $n\in[N]$ and $a\in\mathcal A_n$, and let $v:=a-e_c$.
The function $F_{T,n}$ is convex, and its partial derivatives satisfy
\begin{equation*}
  \partial_xF_{T,n}(a)=\Delta_T\left(p_{n,x}(a)-\1_{\{x=c\}}\right).
\end{equation*}
Since $a\in\mathcal A_n$,
\begin{equation*}
  v_c=-s_n(a),\qquad \sum_{x\in\V\setminus\{c\}}v_x=s_n(a).
\end{equation*}
At $e_c$, the partial derivatives are
\begin{equation*}
  \partial_cF_{T,n}(e_c)=-\frac{(K-1)\Delta_T}{Z_T},\qquad \partial_xF_{T,n}(e_c)=\frac{\Delta_T}{Z_T}\quad(x\ne c).
\end{equation*}
Applying convexity at $e_c$ and using $F_{T,n}(e_c)=0$, we obtain
\begin{equation*}
    F_{T,n}(a)  \ge \langle\nabla_aF_{T,n}(e_c),v\rangle
    = -\frac{(K-1)\Delta_T}{Z_T}v_c+\frac{\Delta_T}{Z_T}\sum_{x\in\V\setminus\{c\}}v_x
    = \frac{K\Delta_T}{Z_T}s_n(a)
    = \kappa_Ts_n(a).
\end{equation*}

Since all softmax probabilities lie in $(0,1)$, the derivative differences in \eqref{eq:phase2-grouped-derivative-differences} satisfy
\begin{equation*}
  0<D_{n,x}(a)\le 2\Delta_T,\qquad x\ne c.
\end{equation*}
Applying convexity at $a$ gives
\begin{equation*}
  0=F_{T,n}(e_c)\ge F_{T,n}(a)+\langle\nabla_aF_{T,n}(a),e_c-a\rangle=F_{T,n}(a)-H_n(a).
\end{equation*}
Summing the upper bounds on $D_{n,x}(a)$ with weights $a_x$ gives $H_n(a)\le 2\Delta_Ts_n(a)$.
This proves \eqref{eq:phase2-ce-attention-comparison}.

Finally, $F_{T,n}(a_n(W))=E_{T,n}(W)$ and $s_n(a_n(W))=s_n(W)$.
Evaluating the sample bounds at $a_n(W)$ and averaging over the $N$ samples proves the empirical bounds.
\end{proof}

The next lemma converts derivatives with respect to the attention weights into a gradient formula for $W$.

\begin{lemma}\label{lem:grouped-loss-chain-rule}
Fix $n\in[N]$, and let $f_n$ be differentiable on an open neighborhood of $\mathcal A_n$ in $\R^{\V}$.
For $a\in\mathcal A_n$ and $x\in\V$, let
\begin{equation*}
  d_{n,x}(a):=\partial_xf_n(a)-\sum_{z\in\V}a_z\partial_zf_n(a).
\end{equation*}
Let $\varphi_n(W):=f_n(a_n(W))$.
Then, for every $W\in\R^{d\times d}$,
\begin{equation*}
  -\nabla_W\varphi_n(W)=\sum_{x\in\X_n^-}a_{n,x}(W)d_{n,x}(a_n(W))A_{n,x}.
\end{equation*}
Moreover, for every $a\in\mathcal A_n$,
\begin{equation*}
  \sum_{x\in\X_n^-}a_xd_{n,x}(a)=-a_cd_{n,c}(a)=a_c\nabla_af_n(a)^\top(a-e_c).
\end{equation*}
The coefficients $d_{n,x}(a)$ need not be nonnegative.
\end{lemma}

\begin{proof}
Let $s_{n,x}:=x^\top Wq_n$ denote the attention score of token $x$, and let $a:=a_n(W)$.
Differentiating the attention formula \eqref{eq:phase2-grouped-attention-mass} gives
\begin{equation*}
  \frac{\partial a_{n,z}}{\partial s_{n,x}}=a_z\left(\1_{\{z=x\}}-a_x\right),\qquad x,z\in\V.
\end{equation*}
The chain rule therefore yields
\begin{equation*}
    \frac{\partial\varphi_n}{\partial s_{n,x}}  = \sum_{z\in\V}\partial_zf_n(a)\frac{\partial a_{n,z}}{\partial s_{n,x}}
    = a_x\left(\partial_xf_n(a)-\sum_{z\in\V}a_z\partial_zf_n(a)\right)
    = a_xd_{n,x}(a).
\end{equation*}
For every $a\in\mathcal A_n$, the definition of $d_{n,x}$ and $\sum_{x\in\V}a_x=1$ imply
\begin{equation*}
  \sum_{x\in\V}a_xd_{n,x}(a)=0.
\end{equation*}
Applying this identity at $a=a_n(W)$ and using $\nabla_Ws_{n,x}=xq_n^\top$, we obtain
\begin{equation*}
    \nabla_W\varphi_n(W)  = \sum_{x\in\V}a_xd_{n,x}(a)xq_n^\top
    = \sum_{x\in\V}a_xd_{n,x}(a)(x-c)q_n^\top
    = -\sum_{x\in\X_n^-}a_xd_{n,x}(a)A_{n,x}.
\end{equation*}
In the last equality, the correct token contributes zero, and tokens absent from the context have zero attention weight.

For any $a\in\mathcal A_n$, the identity $\sum_{x\in\V}a_xd_{n,x}(a)=0$ and the zero coordinates of absent tokens give
\begin{equation*}
  \sum_{x\in\X_n^-}a_xd_{n,x}(a)=-a_cd_{n,c}(a).
\end{equation*}
By the definition of $d_{n,c}$,
\begin{equation*}
  \nabla_af_n(a)^\top(a-e_c)=\sum_{z\in\V}a_z\partial_zf_n(a)-\partial_cf_n(a)=-d_{n,c}(a).
\end{equation*}
Multiplying this identity by $a_c$ proves the remaining equality.
\end{proof}

For the cross entropy loss, we use $D_{n,x}$ and $H_n$ for gradient lower bounds and introduce centered residuals to control changes in the gradient.
For $a\in\mathcal A_n$ and $x\in\V$, let
\begin{equation*}
  r_{n,x}(a):=p_{n,x}(a)-\1_{\{x=c\}},\qquad \bar r_n(a):=\sum_{z\in\V}a_zr_{n,z}(a),\qquad b_{n,x}(a):=r_{n,x}(a)-\bar r_n(a).
\end{equation*}
For $f_n=F_{T,n}$, the definition of $d_{n,x}$ and $\partial_xF_{T,n}(a)=\Delta_Tr_{n,x}(a)$ give
\begin{equation}\label{eq:phase2-centered-derivative}
  d_{n,x}(a)=\Delta_Tb_{n,x}(a).
\end{equation}
We also write $r_{n,x}(W):=r_{n,x}(a_n(W))$, and use the same convention for $\bar r_n$ and $b_{n,x}$.

The following lemma gives the score derivatives and the empirical loss gradient.

\begin{lemma}\label{lem:grouped-gradient}
Fix $W\in\R^{d\times d}$ and $n\in[N]$, and let $a:=a_n(W)$.
Let $g_{n,x}:=\partial\ell_{T,n}/\partial s_{n,x}$ denote the loss derivative with respect to the attention score $s_{n,x}$.
Then
\begin{equation}\label{eq:phase2-score-gradient}
  g_{n,c}=-a_cH_n(a),\qquad g_{n,x}=\Delta_Ta_xb_{n,x}(a),\qquad x\ne c.
\end{equation}
Consequently,
\begin{equation}\label{eq:kq-gradient-comparison}
  -\nabla_W\Loss_T(W)=\frac{1}{N}\sum_{n=1}^N\sum_{x\in\X_n^-}a_{n,x}(W)\left(D_{n,x}(a_n(W))-H_n(a_n(W))\right)A_{n,x}.
\end{equation}
\end{lemma}

\begin{proof}
The sample loss satisfies $\ell_{T,n}(W)=F_{T,n}(a_n(W))+\ell_T^\star$, where $\ell_T^\star$ is constant.
Taking $f_n=F_{T,n}$ in Lemma \ref{lem:grouped-loss-chain-rule} and using \eqref{eq:phase2-centered-derivative}, we obtain
\begin{equation*}
  g_{n,x}=\Delta_Ta_xb_{n,x}(a),\qquad x\in\V.
\end{equation*}
The definition of $H_n$ and $\partial_xF_{T,n}(a)=\Delta_Tr_{n,x}(a)$ give
\begin{equation*}
  H_n(a)=\Delta_T\left(\sum_{z\in\V}a_zr_{n,z}(a)-r_{n,c}(a)\right)=\Delta_T\left(\bar r_n(a)-r_{n,c}(a)\right).
\end{equation*}
The definitions of $b_{n,x}$ and $D_{n,x}$ then imply
\begin{equation*}
  \Delta_Tb_{n,c}(a)=-H_n(a),\qquad \Delta_Tb_{n,x}(a)=D_{n,x}(a)-H_n(a)\quad(x\ne c).
\end{equation*}
Substituting the first identity into the expression for $g_{n,c}$ proves \eqref{eq:phase2-score-gradient}.

Finally, substituting $d_{n,x}(a)=D_{n,x}(a)-H_n(a)$ for $x\ne c$ into the matrix gradient identity in Lemma \ref{lem:grouped-loss-chain-rule} gives
\begin{equation*}
  -\nabla_W\ell_{T,n}(W)=\sum_{x\in\X_n^-}a_{n,x}(W)\left(D_{n,x}(a_n(W))-H_n(a_n(W))\right)A_{n,x}.
\end{equation*}
Averaging over the $N$ samples proves \eqref{eq:kq-gradient-comparison}.
\end{proof}

\subsubsection*{\textbf{Gradient and momentum estimates}}
The following estimates bound the loss gradient and control the error arising from momentum.
The attention bounds, convex loss estimates, and scalar momentum bound below will also be used in the Schatten and Adam analyses.

\paragraph{Stability of attention weights.}
To compare gradients at different iterates, we first control changes in the attention weights under perturbations of $W$.

\begin{lemma}\label{lem:attn-stab}
For any $W,W'\in\R^{d\times d}$, let $r:=\Vert W'-W\Vert_{\op}$.
Then, for every $n\in[N]$ and every $x\in\V$ with $c_{n,x}>0$,
\begin{equation}\label{eq:phase2-attention-ratio-stability}
  e^{-2r}\le \frac{a_{n,x}(W')}{a_{n,x}(W)}\le e^{2r}.
\end{equation}
If $\X_n^-\ne\varnothing$, then
\begin{equation}\label{eq:phase2-competing-mass-ratio-stability}
  e^{-2r}\le \frac{s_n(W')}{s_n(W)}\le e^{2r}.
\end{equation}
Moreover, for every $n\in[N]$,
\begin{equation}\label{eq:phase2-attention-relative-l1}
  \Vert a_n(W')-a_n(W)\Vert_1\le 2\left(e^{2r}-1\right)s_n(W).
\end{equation}
\end{lemma}

\begin{proof}
Fix $n\in[N]$.
The vocabulary embeddings and the query have unit Euclidean norm, so
\begin{equation*}
  |x^\top(W'-W)q_n|\le \Vert x\Vert_2\Vert W'-W\Vert_{\op}\Vert q_n\Vert_2=r,\qquad x\in\V.
\end{equation*}
Let $Z_n(W):=\sum_{z\in\V}c_{n,z}e^{z^\top Wq_n}$.
Since the counts are fixed and nonnegative, the score bound gives
\begin{equation*}
  e^{-r}Z_n(W)\le Z_n(W')\le e^rZ_n(W).
\end{equation*}
For every $x\in\V$ with $c_{n,x}>0$,
\begin{equation*}
  \frac{a_{n,x}(W')}{a_{n,x}(W)}=e^{x^\top(W'-W)q_n}\frac{Z_n(W)}{Z_n(W')}.
\end{equation*}
Both factors on the right lie in $[e^{-r},e^r]$, proving \eqref{eq:phase2-attention-ratio-stability}.

Summing these bounds over $x\in\X_n^-$ gives
\begin{equation*}
  e^{-2r}s_n(W)\le s_n(W')\le e^{2r}s_n(W).
\end{equation*}
If $\X_n^-\ne\varnothing$, then $s_n(W)>0$, so dividing by $s_n(W)$ proves \eqref{eq:phase2-competing-mass-ratio-stability}.

By \eqref{eq:phase2-attention-ratio-stability} and $1-e^{-2r}\le e^{2r}-1$, we have
\begin{equation*}
  |a_{n,x}(W')-a_{n,x}(W)|\le \left(e^{2r}-1\right)a_{n,x}(W),\qquad x\in\X_n^-.
\end{equation*}
The correct token has weight $1-s_n(W)$, and tokens absent from the context have zero weight.
The triangle inequality therefore gives
\begin{equation*}
  \begin{aligned}
    \Vert a_n(W')-a_n(W)\Vert_1 & = |s_n(W')-s_n(W)|+\sum_{x\in\X_n^-}|a_{n,x}(W')-a_{n,x}(W)|\\
    & \le 2\sum_{x\in\X_n^-}|a_{n,x}(W')-a_{n,x}(W)|\\
    & \le 2\left(e^{2r}-1\right)s_n(W).
  \end{aligned}
\end{equation*}
This proves \eqref{eq:phase2-attention-relative-l1}, including the case $\X_n^-=\varnothing$, when both sides are zero.
\end{proof}

\paragraph{Matrix norms and maximum margins.}
Muon and Schatten updates subtract the step size times a maximizer of the inner product with the momentum matrix over the corresponding norm unit ball.
We use a common matrix norm to state the attention, gradient, and momentum estimates for both methods.

Let $\mathsf N$ be a norm on $\R^{d\times d}$ with dual norm $\mathsf N_*$, satisfying
\begin{equation}\label{eq:common-norm-interface}
  \Vert H\Vert_{\op}\le \mathsf N(H)\quad(H\in\R^{d\times d}),\qquad \mathsf N_*(A_{n,x})=\sqrt{2}\quad((n,x)\in\I_2).
\end{equation}
The first condition allows us to control attention changes using $\mathsf N(W'-W)$; the second condition requires each $A_{n,x}$ in the gradient expansion to have dual norm $\sqrt{2}$.

Under Assumption \ref{ass:separability} and $\I_2\ne\varnothing$, let $R_{2,\mathsf N}^{\star}$ denote the optimal value of
\begin{equation}\label{eq:common-hard-margin}
  \begin{aligned}
    \min_{W\in\R^{d\times d}}\quad & \mathsf N(W)\\
    \text{subject to}\quad & \ip{W}{A_{n,x}}\ge1,\qquad (n,x)\in\I_2.
  \end{aligned}
\end{equation}
Lemma \ref{lem:phase2-separability} supplies a feasible matrix $W^\star$.
Norm duality and \eqref{eq:common-norm-interface} imply $\mathsf N(W)\ge1/\sqrt{2}$ for every feasible $W$, so
\begin{equation*}
  \frac{1}{\sqrt{2}}\le R_{2,\mathsf N}^{\star}\le \mathsf N(W^\star)<\infty.
\end{equation*}
Norm sublevel sets are compact in finite dimensions, so the same argument as in the proof of Lemma \ref{lem:phase2-separability} shows that the minimum is attained.

Let $\gamma_{2,\mathsf N}:=1/R_{2,\mathsf N}^{\star}$.
Let $W_{\mathsf N}^{\star}$ be a minimizer, and let $V_{\mathsf N}^{\star}:=W_{\mathsf N}^{\star}/R_{2,\mathsf N}^{\star}$.
Then $\mathsf N(V_{\mathsf N}^{\star})=1$, and feasibility gives $\rho_2(V_{\mathsf N}^{\star})\ge\gamma_{2,\mathsf N}$.
Conversely, if $\mathsf N(V)\le 1$ and $\rho_2(V)>0$, then $V/\rho_2(V)$ is feasible, giving
\begin{equation*}
  R_{2,\mathsf N}^{\star}\le \frac{\mathsf N(V)}{\rho_2(V)}\le \frac{1}{\rho_2(V)}.
\end{equation*}
Hence $\rho_2(V)\le \gamma_{2,\mathsf N}$.
This upper bound also holds when $\rho_2(V)\le 0$, since $\gamma_{2,\mathsf N}>0$.
Consequently,
\begin{equation}\label{eq:common-margin-witness}
  \mathsf N(V_{\mathsf N}^{\star})=1,\qquad \rho_2(V_{\mathsf N}^{\star})=\gamma_{2,\mathsf N}=\max_{\mathsf N(V)\le 1}\rho_2(V).
\end{equation}

For Muon, let $\mathsf N=\Vert\cdot\Vert_{\op}$ and $\mathsf N_*=\Vert\cdot\Vert_{\ast}$, so $R_{2,\mathsf N}^{\star}=R_2^\star$ and $\gamma_{2,\mathsf N}=\gamma_2$.
For the Schatten $S_p$ norm with $p\in[1,\infty]$, let
\begin{equation*}
  \mathsf N=\Vert\cdot\Vert_{S_p},\qquad \mathsf N_*=\Vert\cdot\Vert_{S_{p'}},\qquad \frac{1}{p}+\frac{1}{p'}=1,
\end{equation*}
with the convention $1/\infty=0$.
The first condition in \eqref{eq:common-norm-interface} follows because the largest singular value is at most the Schatten $S_p$ norm.
Each $A_{n,x}$ has rank one and its only nonzero singular value is $\sqrt{2}$, so $\Vert A_{n,x}\Vert_{S_{p'}}=\sqrt{2}$.
Thus both conditions hold for every $p\in[1,\infty]$, including the endpoints.

\paragraph{Gradient lower bounds for convex losses.}
We state the following gradient lower bounds for convex functions of the attention weights so that they can also be used in the Schatten and Adam extensions.
In the Muon analysis, the first bound is used on the initial loss sublevel set to establish loss decrease and summability of the excess losses weighted by the step sizes.
Combining this summability with the second bound, whose leading coefficient is $\gamma_{2,\mathsf N}$, gives the sharper excess loss estimate needed for convergence of the normalized margin.

In the next lemma and its proof, $D_{n,x}$ and $H_n$ are defined using $F_n$.

\begin{lemma}\label{lem:hierarchical-gradient}
Suppose that Assumption \ref{ass:separability} holds, $\I_2\ne\varnothing$, and $\mathsf N$ satisfies \eqref{eq:common-norm-interface}.
For each $n\in[N]$, let $c:=x_n^\star$ and let $F_n$ be differentiable on an open neighborhood of $\mathcal A_n$ in $\R^{\V}$ and convex on $\mathcal A_n$, with $F_n(e_c)=0$.
Let
\begin{equation*}
  E_F(W):=\frac{1}{N}\sum_{n=1}^NF_n(a_n(W)),\qquad D_{n,x}(a):=\langle\nabla_aF_n(a),e_x-e_c\rangle,
\end{equation*}
and suppose that $D_{n,x}(a)>0$ for every $n\in[N]$, $a\in\mathcal A_n$, and $x\in\X_n^-$.
There is a constant $c_{\mathsf N}\in(0,\delta_0/2]$, depending only on the training data and $\mathsf N$, such that the following estimates hold.
\begin{enumerate}
  \item[(i)] For every $W\in\R^{d\times d}$ satisfying $a_{n,c}(W)\ge\delta_0$ for all $n\in[N]$,
  \begin{equation*}
    \mathsf N_*(\nabla_WE_F(W))\ge c_{\mathsf N}E_F(W).
  \end{equation*}

  \item[(ii)] If there are $L,\kappa>0$ such that
  \begin{equation*}
    D_{n,x}(a)\le L\quad(x\in\X_n^-),\qquad F_n(a)\ge\kappa s_n(a),\qquad n\in[N],\quad a\in\mathcal A_n,
  \end{equation*}
  then, for every $W\in\R^{d\times d}$,
  \begin{equation*}
    \mathsf N_*(\nabla_WE_F(W))\ge\gamma_{2,\mathsf N}E_F(W)-\frac{\sqrt{2}NL}{\kappa^2}E_F(W)^2.
  \end{equation*}
\end{enumerate}
\end{lemma}

\begin{proof}
To prove (i), we construct an auxiliary matrix using the token ordering for each query.
Geometrically increasing weights make the correct token's positive contribution dominate the possible negative contributions under the assumed attention lower bound.

For each $q\in\V$ with $\mathcal C_q\ne\varnothing$, the ordering in Lemma \ref{lem:phase2-separability} gives indices $r_q:\V\to\{0,\ldots,K-1\}$ satisfying
\begin{equation*}
  r_q(x_n^\star)\ge r_q(x)+1,\qquad (n,x)\in\I_2,\quad q_n=q.
\end{equation*}
Let
\begin{equation*}
  \Lambda:=4K=\frac{2}{\delta_0},\qquad \widetilde W:=\sum_{\substack{q\in\V\\\mathcal C_q\ne\varnothing}}\left(\sum_{x\in\V}\Lambda^{r_q(x)}x\right)q^\top.
\end{equation*}
Orthonormality gives
\begin{equation*}
  \Vert \widetilde W\Vert_F^2=\sum_{\substack{q\in\V\\\mathcal C_q\ne\varnothing}}\sum_{x\in\V}\Lambda^{2r_q(x)}\le K^2\Lambda^{2(K-1)}.
\end{equation*}
Since $\I_2\ne\varnothing$, there is a query $q$ with $\mathcal C_q\ne\varnothing$.
For this query and any $x\in\V$,
\begin{equation*}
  1\le \Lambda^{r_q(x)}=|x^\top \widetilde Wq|\le \Vert \widetilde W\Vert_{\op}\le \mathsf N(\widetilde W).
\end{equation*}
We may therefore take
\begin{equation*}
  c_{\mathsf N}:=\frac{\delta_0}{2\mathsf N(\widetilde W)}\in(0,\delta_0/2].
\end{equation*}

Fix $W$ satisfying the attention lower bound in (i).
For a sample $n$ with $\X_n^-\ne\varnothing$, let $a:=a_n(W)$, $w_x:=\Lambda^{r_{q_n}(x)}$, and
\begin{equation*}
  H_n:=\sum_{x\in\X_n^-}a_xD_{n,x}(a).
\end{equation*}
Positivity of $D_{n,x}(a)$ gives $H_n\ge0$, while convexity and $F_n(e_c)=0$ give
\begin{equation*}
  F_n(a)\le \langle\nabla_aF_n(a),a-e_c\rangle=H_n.
\end{equation*}
The derivatives in Lemma \ref{lem:grouped-loss-chain-rule} satisfy $d_{n,c}(a)=-H_n$ and $d_{n,x}(a)=D_{n,x}(a)-H_n$ for $x\in\X_n^-$.
Using that lemma and orthonormality, we obtain
\begin{equation*}
  \begin{aligned}
    \langle-\nabla_WF_n(a_n(W)),\widetilde W\rangle & = a_cw_cH_n+\sum_{x\in\X_n^-}a_x\left(H_n-D_{n,x}(a)\right)w_x\\
    & \ge a_cw_cH_n-\frac{w_c}{\Lambda}\sum_{x\in\X_n^-}a_xD_{n,x}(a)\\
    & = w_c\left(a_c-\Lambda^{-1}\right)H_n\\
    & \ge \frac{\delta_0}{2}H_n\\
    & \ge \frac{\delta_0}{2}F_n(a).
  \end{aligned}
\end{equation*}
Here we used $0<w_x\le w_c/\Lambda$ for $x\in\X_n^-$, $w_c\ge1$, $a_c\ge\delta_0$, and $\Lambda^{-1}=\delta_0/2$.
If $\X_n^-=\varnothing$, then $a_n(W)=e_c$ for every $W$, so both $F_n(a_n(W))$ and $\nabla_WF_n(a_n(W))$ vanish.
Averaging over the samples and applying norm duality therefore give
\begin{equation*}
  \frac{\delta_0}{2}E_F(W)\le \langle-\nabla_WE_F(W),\widetilde W\rangle\le \mathsf N_*(\nabla_WE_F(W))\mathsf N(\widetilde W).
\end{equation*}
The definition of $c_{\mathsf N}$ proves (i).

Next, assume the additional conditions in (ii), and fix an arbitrary $W\in\R^{d\times d}$.
Let $V_{\mathsf N}^{\star}$ be the matrix in \eqref{eq:common-margin-witness}, and let $m_{n,x}:=\langle V_{\mathsf N}^{\star},A_{n,x}\rangle$.
Equation \eqref{eq:common-margin-witness} and norm duality give
\begin{equation*}
  \gamma_{2,\mathsf N}\le m_{n,x}\le \sqrt{2},\qquad (n,x)\in\I_2.
\end{equation*}
For each sample, let $a:=a_n(W)$ and $H_n:=\sum_{x\in\X_n^-}a_xD_{n,x}(a)$.
The assumptions and convexity imply
\begin{equation*}
  0\le F_n(a)\le H_n\le Ls_n(W).
\end{equation*}
Lemma \ref{lem:grouped-loss-chain-rule} yields
\begin{equation*}
  \begin{aligned}
    \langle-\nabla_WF_n(a_n(W)),V_{\mathsf N}^{\star}\rangle & = \sum_{x\in\X_n^-}a_xD_{n,x}(a)m_{n,x}-H_n\sum_{x\in\X_n^-}a_xm_{n,x}\\
    & \ge \gamma_{2,\mathsf N}H_n-\sqrt{2}s_n(W)H_n\\
    & \ge \gamma_{2,\mathsf N}F_n(a)-\sqrt{2}Ls_n(W)^2.
  \end{aligned}
\end{equation*}
The last inequality uses $H_n\ge F_n(a)$ in the positive term and $H_n\le Ls_n(W)$ in the negative term.
Moreover, $F_n(a_n(W))\ge\kappa s_n(W)\ge0$, so
\begin{equation*}
    \frac{1}{N}\sum_{n=1}^Ns_n(W)^2  \le \frac{1}{N\kappa^2}\sum_{n=1}^NF_n(a_n(W))^2
    \le \frac{1}{N\kappa^2}\left(\sum_{n=1}^NF_n(a_n(W))\right)^2
    = \frac{N}{\kappa^2}E_F(W)^2.
\end{equation*}
Averaging the sample estimates and using $\mathsf N(V_{\mathsf N}^{\star})=1$ give
\begin{equation*}
  \begin{aligned}
    \mathsf N_*(\nabla_WE_F(W)) & \ge \langle-\nabla_WE_F(W),V_{\mathsf N}^{\star}\rangle\\
    & \ge \gamma_{2,\mathsf N}E_F(W)-\frac{\sqrt{2}L}{N}\sum_{n=1}^Ns_n(W)^2\\
    & \ge \gamma_{2,\mathsf N}E_F(W)-\frac{\sqrt{2}NL}{\kappa^2}E_F(W)^2.
  \end{aligned}
\end{equation*}
This proves (ii).

Under the additional conditions in (ii), $E_F(W)\ge0$.
Combining (i) and (ii), for every $W$ satisfying $a_{n,c}(W)\ge\delta_0$ for all $n\in[N]$, we obtain
\begin{equation}\label{eq:grouped-combined-gradient}
  \mathsf N_*(\nabla_WE_F(W))\ge E_F(W)\max\left\{c_{\mathsf N},\gamma_{2,\mathsf N}-\frac{\sqrt{2}NL}{\kappa^2}E_F(W)\right\}.
\end{equation}
\end{proof}

For $F_n=F_{T,n}$, Lemma \ref{lem:vertex} and \eqref{eq:phase2-grouped-derivative-differences} verify the hypotheses of Lemma \ref{lem:hierarchical-gradient} with $L=2\Delta_T$ and $\kappa=\kappa_T$.
Let $C_T:=2\sqrt{2}N\Delta_T/\kappa_T^2$.
Since $E_F=E_T$ and $\nabla_WE_T=\nabla_W\Loss_T$, the combined bound \eqref{eq:grouped-combined-gradient} gives
\begin{equation}\label{eq:hierarchical-gradient-excess}
  \mathsf N_*(\nabla_W\Loss_T(W))\ge E_T(W)\max\{c_{\mathsf N},\gamma_{2,\mathsf N}-C_TE_T(W)\}
\end{equation}
for every $W\in\R^{d\times d}$ satisfying $a_{n,x_n^\star}(W)\ge\delta_0$ for all $n\in[N]$.

We next bound the change in the gradient under perturbations of $W$.
Keeping the factor $\bar s(W)$ allows us to control the momentum error and the Taylor remainder by the excess loss using Lemma \ref{lem:vertex}.

\begin{lemma}\label{lem:log-stab}
Let $\mathsf N$ satisfy \eqref{eq:common-norm-interface}, and let
\begin{equation*}
  L_{1,T}:=2\sqrt{2}\Delta_T(2+\Delta_T).
\end{equation*}
For every $W,W'\in\R^{d\times d}$, let $d_W:=\mathsf N(W'-W)$.
Then
\begin{equation}\label{eq:phase2-gradient-relative-global}
  \mathsf N_*(\nabla_W\Loss_T(W')-\nabla_W\Loss_T(W))\le L_{1,T}(e^{2d_W}-1)\bar s(W).
\end{equation}
The constant $L_{1,T}$ depends only on $\Delta_T$ and is independent of the initial step size in Phase II.
\end{lemma}

\begin{proof}
Fix $n\in[N]$, and let $a:=a_n(W)$ and $a':=a_n(W')$.
For this sample, we omit the index $n$ in $p,r,\bar r$, and $b$, and use primes to indicate evaluation at $a'$.
The residuals lie in $[-1,1]$, and $\bar r$ is their average with attention weights.
Hence $|b_x|\le 2$ and $|b_x'|\le 2$ for every $x\in\V$.

Since $\Vert W'-W\Vert_{\op}\le d_W$, Lemma \ref{lem:attn-stab} gives
\begin{equation*}
  \Vert a'-a\Vert_1\le 2(e^{2d_W}-1)s_n(W).
\end{equation*}
The Jacobian of $p(a)$ is $\Delta_T(\operatorname{diag}(p)-pp^\top)$.
For each $z\in\V$, the sum of the absolute values in column $z$ satisfies
\begin{equation*}
  \sum_{x\in\V}\left|\frac{\partial p_x}{\partial a_z}\right|=2\Delta_Tp_z(1-p_z)\le \frac{\Delta_T}{2},
\end{equation*}
where we used $p_z(1-p_z)\le 1/4$.
Integrating along the segment from $a$ to $a'$ gives
\begin{equation*}
  \Vert p'-p\Vert_1\le \frac{\Delta_T}{2}\Vert a'-a\Vert_1.
\end{equation*}
Using $r'-r=p'-p$, $\Vert r'\Vert_\infty\le 1$, and $\Vert a\Vert_1=1$, we obtain
\begin{equation*}
  |\bar r'-\bar r|=|(a'-a)^\top r'+a^\top(r'-r)|\le \Vert a'-a\Vert_1+\Vert p'-p\Vert_\infty.
\end{equation*}
Consequently, for every $x\in\V$,
\begin{equation*}
  \begin{aligned}
    |b_x'-b_x| & \le |p_x'-p_x|+|\bar r'-\bar r|\\
    & \le \Vert a'-a\Vert_1+2\Vert p'-p\Vert_1\\
    & \le (1+\Delta_T)\Vert a'-a\Vert_1.
  \end{aligned}
\end{equation*}

For $x\in\X_n^-$, the score derivative in \eqref{eq:phase2-score-gradient} is $g_{n,x}(W)=\Delta_Ta_{n,x}(W)b_{n,x}(W)$.
Its change can be written as
\begin{equation*}
  g_{n,x}(W')-g_{n,x}(W)=\Delta_T\left((a_x'-a_x)b_x'+a_x(b_x'-b_x)\right).
\end{equation*}
The attention ratio bound in Lemma \ref{lem:attn-stab} gives
\begin{equation*}
  |a_x'-a_x|\le (e^{2d_W}-1)a_x,\qquad x\in\X_n^-.
\end{equation*}
Combining these estimates with $|b_x'|\le 2$ and $\sum_{x\in\X_n^-}a_x=s_n(W)$, we obtain
\begin{equation*}
  \begin{aligned}
    \sum_{x\in\X_n^-}|g_{n,x}(W')-g_{n,x}(W)| & \le 2\Delta_T(e^{2d_W}-1)s_n(W)+\Delta_T(1+\Delta_T)s_n(W)\Vert a'-a\Vert_1\\
    & \le 2\Delta_T(e^{2d_W}-1)s_n(W)+2\Delta_T(1+\Delta_T)(e^{2d_W}-1)s_n(W)^2\\
    & \le 2\Delta_T(2+\Delta_T)(e^{2d_W}-1)s_n(W).
  \end{aligned}
\end{equation*}
The last inequality uses $0\le s_n(W)\le 1$.

Finally, the gradient representation in Lemma \ref{lem:grouped-gradient}, the dual norm triangle inequality, and $\mathsf N_*(A_{n,x})=\sqrt{2}$ give
\begin{equation*}
  \begin{aligned}
    \mathsf N_*(\nabla_W\Loss_T(W')-\nabla_W\Loss_T(W)) & \le \frac{\sqrt{2}}{N}\sum_{n=1}^N\sum_{x\in\X_n^-}|g_{n,x}(W')-g_{n,x}(W)|\\
    & \le \frac{2\sqrt{2}\Delta_T(2+\Delta_T)}{N}(e^{2d_W}-1)\sum_{n=1}^Ns_n(W)\\
    & = L_{1,T}(e^{2d_W}-1)\bar s(W).
  \end{aligned}
\end{equation*}
\end{proof}

\paragraph{Momentum and gradient alignment.}
To analyze Muon and the Schatten updates together, we write $W_t$, $M_t$, $\eta_t$, and $\beta$ for the attention matrix, momentum matrix, step size, and momentum parameter.
For Phase II of Algorithm \ref{alg:two-phase-muon}, these are $W_{2,t}$, $M_{2,t}$, $\eta_{2,t}$, and $\beta_2$, respectively.
Let $\mathsf N$ satisfy \eqref{eq:common-norm-interface}, and consider the updates
\begin{equation}\label{eq:common-momentum-update}
  \begin{aligned}
    & M_t=\beta M_{t-1}+(1-\beta)\nabla_W\Loss_T(W_t),\\
    & D_t\in\operatorname*{argmax}_{\mathsf N(D)\le 1}\langle M_t,D\rangle,\\
    & W_{t+1}=W_t-\eta_tD_t,
  \end{aligned}
\end{equation}
where $0\le \beta<1$, $M_{-1}=0$, and $D_t=0$ when $M_t=0$.
For $\mathsf N=\Vert\cdot\Vert_{\op}$ and $M_t\ne 0$, Muon chooses $D_t=U_tV_t^\top$, where $M_t=U_t\Sigma_tV_t^\top$ is the compact singular value decomposition.

Let
\begin{equation*}
  \widehat M_t:=\frac{M_t}{1-\beta^{t+1}},\qquad \Omega_t:=\widehat M_t-\nabla_W\Loss_T(W_t),\qquad E_t:=E_T(W_t).
\end{equation*}
Expanding the momentum recurrence from $M_{-1}=0$ gives
\begin{equation*}
  \widehat M_t=\frac{1-\beta}{1-\beta^{t+1}}\sum_{r=0}^t\beta^r\nabla_W\Loss_T(W_{t-r}).
\end{equation*}
The coefficients are nonnegative and sum to one.
When $\beta=0$, this average reduces to the current gradient, so $\Omega_t=0$.
Since $1-\beta^{t+1}>0$, the matrices $M_t$ and $\widehat M_t$ have the same maximizing directions in \eqref{eq:common-momentum-update}.

In this notation, the cumulative sums defined in the main text are
\begin{equation*}
  S_t:=\sum_{r=0}^{t-1}\eta_r,\qquad Q_t:=\sum_{r=0}^{t-1}\eta_r^2,\qquad S_0=Q_0=0.
\end{equation*}
The next lemma bounds the exponential terms that arise when Lemma \ref{lem:log-stab} is applied to past iterates.
Lemma C.1 of \citet{zhang2024implicit} gives a related estimate for power law step sizes with exponent in $(0,1]$.
We give a proof for all $\alpha>0$ and $t\ge 0$, with a constant independent of the initial step size.

\begin{lemma}\label{lem:ema-phase2}
Let $\eta_t:=\eta_0(t+1)^{-\alpha}$, where $\alpha>0$ and $0<\eta_0\le 1$, and let $c>0$ and $\beta\in[0,1)$.
There is a finite constant $C(\alpha,c,\beta)$, depending only on $\alpha,c,\beta$, such that, for every integer $t\ge 0$,
\begin{equation*}
  \frac{1-\beta}{1-\beta^{t+1}}\sum_{r=0}^t\beta^r\left[\exp\left(c\sum_{j=1}^r\eta_{t-j}\right)-1\right]\le C(\alpha,c,\beta)\eta_t.
\end{equation*}
An empty sum is zero.
When $\beta=0$, the sum contains only the term with $r=0$, and we take $C(\alpha,c,0):=0$.
\end{lemma}

\begin{proof}
If $t=0$ or $\beta=0$, the left hand side is zero.
Suppose that $0<\beta<1$ and $t\ge 1$.
Let
\begin{equation*}
  P_t:=\sum_{k=1}^tk^{-\alpha},\qquad U_{t,r}:=\sum_{k=t-r+1}^tk^{-\alpha},\qquad U_{t,0}:=0.
\end{equation*}
Then $\sum_{j=1}^r\eta_{t-j}=\eta_0U_{t,r}$.
For $1\le r\le t$, we have
\begin{equation*}
  U_{t,r}\le \frac{rP_t}{t},\qquad U_{t,r}\le \frac{r(r+1)^\alpha}{(t+1)^\alpha}.
\end{equation*}
The first bound follows because the average of the last $r$ terms of the decreasing sequence $(k^{-\alpha})_{k=1}^t$ is at most the average of all $t$ terms.
The second follows from $U_{t,r}\le r(t-r+1)^{-\alpha}$ and $(t+1)/(t-r+1)\le r+1$.

To control the exponential factor, note that $P_t/t\to 0$ because $k^{-\alpha}\to 0$.
Choose an integer $\tau\ge 1$, depending only on $\alpha,c,\beta$, such that $cP_t/t\le -(\log\beta)/2$ for all $t\ge\tau$.
For $t\ge\tau$, the first bound on $U_{t,r}$ gives $\beta^re^{cU_{t,r}}\le \beta^{r/2}$.
For $1\le t<\tau$, we use $U_{t,r}\le P_\tau$ and $\beta^r\le \beta^{r/2}$.
Thus
\begin{equation*}
  \beta^re^{cU_{t,r}}\le e^{cP_\tau}\beta^{r/2},\qquad t\ge 1,\quad 0\le r\le t.
\end{equation*}
Let
\begin{equation*}
  C(\alpha,c,\beta):=ce^{cP_\tau}\sum_{r=1}^{\infty}r(r+1)^\alpha\beta^{r/2}.
\end{equation*}
The series converges because $0<\beta<1$, so this constant is finite and depends only on $\alpha,c,\beta$.

Convexity of the exponential and $e^v-1\le ve^v$ for $v\ge 0$ give
\begin{equation*}
  e^{c\eta_0U}-1\le \eta_0(e^{cU}-1)\le c\eta_0Ue^{cU},\qquad U\ge 0,\quad 0<\eta_0\le 1.
\end{equation*}
Using $(1-\beta)/(1-\beta^{t+1})\le 1$ and the preceding bounds, we obtain
\begin{equation*}
  \begin{aligned}
    \frac{1-\beta}{1-\beta^{t+1}}\sum_{r=0}^t\beta^r\left[\exp\left(c\sum_{j=1}^r\eta_{t-j}\right)-1\right] & \le c\eta_0\sum_{r=1}^t\beta^rU_{t,r}e^{cU_{t,r}}\\
    & \le \frac{c\eta_0e^{cP_\tau}}{(t+1)^\alpha}\sum_{r=1}^{\infty}r(r+1)^\alpha\beta^{r/2}\\
    & = C(\alpha,c,\beta)\eta_t.
  \end{aligned}
\end{equation*}
\end{proof}

We now combine Lemmas \ref{lem:log-stab} and \ref{lem:ema-phase2} to bound the momentum error by a multiple of $\eta_t\bar s(W_t)$.

\begin{lemma}\label{lem:ema-err}
Let $\mathsf N$ satisfy \eqref{eq:common-norm-interface}, and consider the updates in \eqref{eq:common-momentum-update} with $\eta_t=\eta_0(t+1)^{-\alpha}$, where $\alpha>0$ and $0<\eta_0\le 1$.
Let
\begin{equation*}
  C_{M,T,\alpha}:=L_{1,T}C(\alpha,2,\beta).
\end{equation*}
Then, for every integer $t\ge 0$,
\begin{equation*}
  \mathsf N_*(\Omega_t)\le C_{M,T,\alpha}\eta_t\bar s(W_t).
\end{equation*}
The constant $C_{M,T,\alpha}$ depends only on $\Delta_T,\alpha,\beta$ and is independent of $\eta_0$ and $t$.
\end{lemma}

\begin{proof}
If $t=0$ or $\beta=0$, then $\Omega_t=0$, so the bound holds.
Suppose that $t\ge 1$ and $0<\beta<1$.
Let
\begin{equation*}
  w_{t,r}:=\frac{(1-\beta)\beta^r}{1-\beta^{t+1}},\qquad 0\le r\le t.
\end{equation*}
These weights are nonnegative and satisfy $\sum_{r=0}^tw_{t,r}=1$.
Using the expansion of $\widehat M_t$ and $\sum_{r=0}^tw_{t,r}=1$, we obtain
\begin{equation*}
  \Omega_t=\sum_{r=0}^tw_{t,r}\left(\nabla_W\Loss_T(W_{t-r})-\nabla_W\Loss_T(W_t)\right).
\end{equation*}

For $0\le r\le t$, summing the updates over $s=t-r,\ldots,t-1$ and using $\mathsf N(D_s)\le 1$ gives
\begin{equation*}
  \mathsf N(W_t-W_{t-r})\le \sum_{s=t-r}^{t-1}\eta_s\mathsf N(D_s)\le \sum_{j=1}^r\eta_{t-j}.
\end{equation*}
We apply Lemma \ref{lem:log-stab} with $W=W_t$ and $W'=W_{t-r}$ so that $\bar s(W_t)$ appears as a factor in the bound:
\begin{equation*}
  \mathsf N_*(\nabla_W\Loss_T(W_{t-r})-\nabla_W\Loss_T(W_t))\le L_{1,T}\left[\exp\left(2\sum_{j=1}^r\eta_{t-j}\right)-1\right]\bar s(W_t).
\end{equation*}
The dual norm triangle inequality and Lemma \ref{lem:ema-phase2} with $c=2$ now give
\begin{equation*}
  \begin{aligned}
    \mathsf N_*(\Omega_t) & \le \sum_{r=0}^tw_{t,r}\mathsf N_*(\nabla_W\Loss_T(W_{t-r})-\nabla_W\Loss_T(W_t))\\
    & \le L_{1,T}\bar s(W_t)\sum_{r=0}^tw_{t,r}\left[\exp\left(2\sum_{j=1}^r\eta_{t-j}\right)-1\right]\\
    & \le L_{1,T}C(\alpha,2,\beta)\eta_t\bar s(W_t)\\
    & = C_{M,T,\alpha}\eta_t\bar s(W_t).
  \end{aligned}
\end{equation*}
\end{proof}

\subsubsection*{\textbf{Descent and margin convergence}}

\paragraph{Loss decrease on the initial loss sublevel set.}
We first use the momentum error bound to bound the inner product of the current gradient with the update direction from below.
We then combine this bound with the gradient lower bounds and the Taylor remainder estimate to prove loss decrease for a sufficiently small initial step size.

\begin{lemma}\label{lem:first-order}
Under the assumptions of Lemma \ref{lem:ema-err}, every maximizing direction $D_t$ in \eqref{eq:common-momentum-update} satisfies
\begin{equation*}
  \langle\nabla_W\Loss_T(W_t),D_t\rangle\ge \mathsf N_*(\nabla_W\Loss_T(W_t))-2C_{M,T,\alpha}\eta_t\bar s(W_t)
\end{equation*}
for every integer $t\ge 0$.
\end{lemma}

\begin{proof}
Since $D_t$ maximizes the inner product with $\widehat M_t$ over the unit ball of $\mathsf N$, we have $\langle\widehat M_t,D_t\rangle=\mathsf N_*(\widehat M_t)$ and $\mathsf N(D_t)\le 1$.
Using $\nabla_W\Loss_T(W_t)=\widehat M_t-\Omega_t$, norm duality, and the reverse triangle inequality, we obtain
\begin{equation*}
  \begin{aligned}
    \langle\nabla_W\Loss_T(W_t),D_t\rangle & = \mathsf N_*(\widehat M_t)-\langle\Omega_t,D_t\rangle\\
    & \ge \mathsf N_*(\widehat M_t)-\mathsf N_*(\Omega_t)\\
    & \ge \mathsf N_*(\nabla_W\Loss_T(W_t))-2\mathsf N_*(\Omega_t).
  \end{aligned}
\end{equation*}
Applying Lemma \ref{lem:ema-err} to the last term proves the result.
\end{proof}

The following estimate bounds the remainder in the first order expansion of the loss by a multiple of $\eta^2\bar s(W)$.

\begin{lemma}\label{lem:taylor}
Let $\mathsf N$ satisfy \eqref{eq:common-norm-interface}, and let
\begin{equation*}
  C_{Q,T}:=e^2L_{1,T}.
\end{equation*}
For every $W,D\in\R^{d\times d}$ with $\mathsf N(D)\le 1$ and every $\eta\in[0,1]$,
\begin{equation*}
  \Loss_T(W-\eta D)\le \Loss_T(W)-\eta\langle\nabla_W\Loss_T(W),D\rangle+C_{Q,T}\eta^2\bar s(W).
\end{equation*}
\end{lemma}

\begin{proof}
The fundamental theorem of calculus gives
\begin{equation*}
  \Loss_T(W-\eta D)-\Loss_T(W)+\eta\langle\nabla_W\Loss_T(W),D\rangle=-\int_0^\eta\langle\nabla_W\Loss_T(W-\tau D)-\nabla_W\Loss_T(W),D\rangle\,d\tau.
\end{equation*}
Since $\mathsf N(\tau D)\le\tau$ and $\mathsf N(D)\le 1$, norm duality and Lemma \ref{lem:log-stab} bound the absolute value of the remainder by
\begin{equation*}
  L_{1,T}\bar s(W)\int_0^\eta(e^{2\tau}-1)\,d\tau\le e^2L_{1,T}\eta^2\bar s(W)=C_{Q,T}\eta^2\bar s(W).
\end{equation*}
Here we used $e^{2\tau}-1\le 2e^2\tau$ for $0\le\tau\le\eta\le 1$.
\end{proof}

By Lemma \ref{lem:vertex}, $\bar s(W_t)\le E_t/\kappa_T$.
In the loss increment, this bounds both the momentum contribution and the Taylor remainder by multiples of $\eta_t^2E_t$.
For a fixed exponent $\alpha>0$, let
\begin{equation*}
  C_{D,T,\alpha}:=\frac{2C_{M,T,\alpha}+C_{Q,T}}{\kappa_T}.
\end{equation*}
To ensure that these errors are dominated by the gradient lower bound, let
\begin{equation*}
  \bar\eta_{T,\alpha,\mathsf N}:=\min\left\{1,\frac{c_{\mathsf N}}{2C_{D,T,\alpha}}\right\}>0.
\end{equation*}
For fixed $\Delta_T$, $\alpha$, $\beta$, $\mathsf N$, and training data, these constants are independent of $\eta_0$ and $t$.
For spectral Muon with $\alpha=1/2$ and $\beta=\beta_2$, we take $\bar\eta_T:=\bar\eta_{T,1/2,\Vert\cdot\Vert_{\op}}$ in Theorem \ref{thm:main}.

\begin{lemma}\label{lem:descent}
Suppose that Assumptions \ref{ass:token-label} and \ref{ass:separability} hold, $\I_2\ne\varnothing$, and $\mathsf N$ satisfies \eqref{eq:common-norm-interface}.
Fix a prediction matrix $W_{1,T}$ satisfying $CW_{1,T}=\Delta_TB$ with $\Delta_T\ge 2N(K-1)\log K$, and suppose that $\Loss_T(0)\le \log K$.
Use \eqref{eq:common-momentum-update} with $W_0=0$, $M_{-1}=0$, and
\begin{equation*}
  \eta_t=\eta_0(t+1)^{-\alpha},\qquad \alpha>0,\qquad 0<\eta_0\le \bar\eta_{T,\alpha,\mathsf N}.
\end{equation*}
Then, for every integer $t\ge 0$,
\begin{equation*}
  \Loss_T(W_t)\le \Loss_T(0),\qquad a_{n,x_n^\star}(W_t)\ge\delta_0\quad(n\in[N]),
\end{equation*}
and the excess loss satisfies
\begin{equation}\label{eq:common-unified-descent}
  E_{t+1}\le \left[1-\eta_t\max\{c_{\mathsf N},\gamma_{2,\mathsf N}-C_TE_t\}+C_{D,T,\alpha}\eta_t^2\right]E_t.
\end{equation}
\end{lemma}

\begin{proof}
Let $\mathcal S_0:=\{W:\Loss_T(W)\le \Loss_T(0)\}$.
Since $\Loss_T(0)\le \log K$, Lemma \ref{lem:noncollapse-entry} gives $a_{n,x_n^\star}(W)\ge\delta_0$ for every $W\in\mathcal S_0$ and $n\in[N]$.
We prove by induction that $W_t\in\mathcal S_0$ for every $t\ge 0$, starting from $W_0=0$.

Suppose that $W_t\in\mathcal S_0$.
Since $\eta_t\le 1$, Lemmas \ref{lem:first-order} and \ref{lem:taylor}, together with $E_T=\Loss_T-\ell_T^\star$, give
\begin{equation*}
  \begin{aligned}
    E_{t+1}-E_t & \le -\eta_t\langle\nabla_W\Loss_T(W_t),D_t\rangle+C_{Q,T}\eta_t^2\bar s(W_t)\\
    & \le -\eta_t\mathsf N_*(\nabla_W\Loss_T(W_t))+(2C_{M,T,\alpha}+C_{Q,T})\eta_t^2\bar s(W_t).
  \end{aligned}
\end{equation*}
Using $\kappa_T\bar s(W_t)\le E_t$, the gradient lower bound \eqref{eq:hierarchical-gradient-excess}, and the definition of $C_{D,T,\alpha}$ gives \eqref{eq:common-unified-descent}.
The step size bound also gives
\begin{equation*}
  \max\{c_{\mathsf N},\gamma_{2,\mathsf N}-C_TE_t\}-C_{D,T,\alpha}\eta_t\ge c_{\mathsf N}-C_{D,T,\alpha}\eta_t\ge \frac{c_{\mathsf N}}{2}.
\end{equation*}
Since $E_t\ge 0$ by Lemma \ref{lem:vertex}, the recurrence yields
\begin{equation*}
  E_t-E_{t+1}\ge \frac{c_{\mathsf N}}{2}\eta_tE_t\ge 0.
\end{equation*}
Thus $\Loss_T(W_{t+1})\le \Loss_T(W_t)\le \Loss_T(0)$, which closes the induction.
The attention lower bound and \eqref{eq:common-unified-descent} therefore hold at every iteration.
\end{proof}

\paragraph{Loss decay and margin growth.}
The next lemma converts upper bounds on the excess loss into lower bounds on the training margin.
We will combine it with \eqref{eq:common-unified-descent} in the subsequent rate estimates.
Let $L_{\max}:=\max_{n\in[N]}L_n$ denote the largest training context length.

\begin{lemma}\label{lem:margin-loss}
Suppose that Assumption \ref{ass:token-label} holds and $\I_2\ne\varnothing$.
Let
\begin{equation*}
  \nu_T:=\frac{\kappa_T\delta_0}{NL_{\max}}>0.
\end{equation*}
For every $W\in\R^{d\times d}$ satisfying $a_{n,x_n^\star}(W)\ge\delta_0$ for all $n\in[N]$,
\begin{equation*}
  E_T(W)\ge\nu_Te^{-\rho_2(W)}.
\end{equation*}
\end{lemma}

\begin{proof}
Since $\I_2$ is finite and nonempty, choose $(n,x)\in\I_2$ such that $\langle W,A_{n,x}\rangle=\rho_2(W)$, and let $c:=x_n^\star$.
We have $c_{n,x}\ge 1$ and $0<c_{n,c}\le L_n\le L_{\max}$.
The attention ratio identity \eqref{eq:attn-odds-margin-bridge} and the assumed bound $a_{n,c}(W)\ge\delta_0$ therefore give
\begin{equation*}
  a_{n,x}(W)=a_{n,c}(W)\frac{c_{n,x}}{c_{n,c}}e^{-\rho_2(W)}\ge\frac{\delta_0}{L_{\max}}e^{-\rho_2(W)}.
\end{equation*}
Since all $s_m(W)$ are nonnegative and $s_n(W)\ge a_{n,x}(W)$, Lemma \ref{lem:vertex} yields
\begin{equation*}
  E_T(W)\ge\kappa_T\bar s(W)\ge\frac{\kappa_T}{N}s_n(W)\ge\frac{\kappa_T}{N}a_{n,x}(W)\ge\nu_Te^{-\rho_2(W)}.
\end{equation*}
\end{proof}

We now combine the descent recurrence with Lemma \ref{lem:margin-loss} to derive bounds on the excess loss and the normalized margin.

\begin{lemma}\label{lem:momentum-margin-rates}
Under the hypotheses of Lemma \ref{lem:descent}, $E_t>0$ for every integer $t\ge 0$.
There are constants $C_L,C_\rho>0$, depending only on $\Delta_T,\alpha,\beta,\mathsf N$ and the finite training data, and independent of $\eta_0,t$ and the maximizing directions, such that the following bounds hold.
\begin{enumerate}
  \item[(i)] For every integer $t\ge 0$,
  \begin{equation*}
    \log E_t\le -\gamma_{2,\mathsf N}S_t+C_LQ_t+C_L.
  \end{equation*}

  \item[(ii)] For every integer $t\ge 1$ satisfying $\gamma_{2,\mathsf N}S_t-C_LQ_t-C_\rho>0$, we have $W_t\ne 0$ and
  \begin{equation}\label{eq:common-normalized-margin}
    0\le \gamma_{2,\mathsf N}-\rho_2\left(\frac{W_t}{\mathsf N(W_t)}\right)\le \frac{C_LQ_t+C_\rho}{S_t}.
  \end{equation}
\end{enumerate}
If $0<\alpha\le 1$, then the iterates are eventually nonzero and $\rho_2(W_t/\mathsf N(W_t))\to\gamma_{2,\mathsf N}$ as $t\to\infty$.
\end{lemma}

\begin{proof}
Summing the updates from $W_0=0$ and using $\mathsf N(D_s)\le 1$ give
\begin{equation*}
  W_t=-\sum_{s=0}^{t-1}\eta_sD_s,\qquad \mathsf N(W_t)\le S_t.
\end{equation*}
Thus $\rho_2(W_t)$ is finite at every finite iteration.
Lemmas \ref{lem:descent} and \ref{lem:margin-loss} imply
\begin{equation*}
  E_t\ge \nu_Te^{-\rho_2(W_t)}>0.
\end{equation*}

The recurrence \eqref{eq:common-unified-descent} and the step size bound give, for every integer $m\ge 0$,
\begin{equation*}
  \frac{c_{\mathsf N}}{2}\sum_{s=0}^{m}\eta_sE_s\le \sum_{s=0}^{m}(E_s-E_{s+1})=E_0-E_{m+1}\le E_0.
\end{equation*}
Letting $m\to\infty$ yields
\begin{equation*}
  \sum_{s=0}^{\infty}\eta_sE_s\le \frac{2E_0}{c_{\mathsf N}}.
\end{equation*}

This summability bound controls the accumulated $C_T\eta_sE_s$ correction in the logarithmic loss estimate.
Since $E_s,E_{s+1}>0$, the factor multiplying $E_s$ in \eqref{eq:common-unified-descent} is positive.
Taking logarithms, using $\log(1+v)\le v$ for $v>-1$ and $\max\{c_{\mathsf N},\gamma_{2,\mathsf N}-C_TE_s\}\ge\gamma_{2,\mathsf N}-C_TE_s$, and summing give
\begin{equation*}
  \begin{aligned}
    \log E_t & \le \log E_0-\sum_{s=0}^{t-1}\eta_s\left[\max\{c_{\mathsf N},\gamma_{2,\mathsf N}-C_TE_s\}-C_{D,T,\alpha}\eta_s\right]\\
    & \le \log E_0-\gamma_{2,\mathsf N}S_t+C_T\sum_{s=0}^{t-1}\eta_sE_s+C_{D,T,\alpha}Q_t\\
    & \le -\gamma_{2,\mathsf N}S_t+C_{D,T,\alpha}Q_t+\log E_0+\frac{2C_TE_0}{c_{\mathsf N}}.
  \end{aligned}
\end{equation*}
Let
\begin{equation*}
  C_L:=\max\left\{1,C_{D,T,\alpha},|\log E_0|+\frac{2C_TE_0}{c_{\mathsf N}}\right\},\qquad C_\rho:=C_L+|\log\nu_T|.
\end{equation*}
Since $CW_{1,T}=\Delta_TB$ and $W_0=0$, the initial excess loss $E_0$ depends only on $\Delta_T$ and the training data.
Thus these constants have the stated dependence, and (i) follows from the definition of $C_L$.

Lemma \ref{lem:margin-loss} and (i) give
\begin{equation*}
  \begin{aligned}
    \rho_2(W_t) & \ge \log\nu_T-\log E_t\\
    & \ge \gamma_{2,\mathsf N}S_t-C_LQ_t-C_L+\log\nu_T\\
    & \ge \gamma_{2,\mathsf N}S_t-C_LQ_t-C_\rho.
  \end{aligned}
\end{equation*}
When this lower bound is positive, $\rho_2(W_t)>0$ and hence $W_t\ne 0$.
Positive homogeneity and $\mathsf N(W_t)\le S_t$ then imply
\begin{equation*}
  \rho_2\left(\frac{W_t}{\mathsf N(W_t)}\right)=\frac{\rho_2(W_t)}{\mathsf N(W_t)}\ge\frac{\rho_2(W_t)}{S_t}\ge\gamma_{2,\mathsf N}-\frac{C_LQ_t+C_\rho}{S_t}.
\end{equation*}
The maximum margin identity \eqref{eq:common-margin-witness} gives $\rho_2(W_t/\mathsf N(W_t))\le\gamma_{2,\mathsf N}$, proving (ii).

Finally, suppose that $0<\alpha\le 1$.
Then $S_t\to\infty$ and $\eta_s\to 0$.
For every $\varepsilon>0$, choose $m$ such that $\eta_s\le \varepsilon$ for all $s\ge m$.
For $t\ge m$,
\begin{equation*}
  Q_t=Q_m+\sum_{s=m}^{t-1}\eta_s^2\le Q_m+\varepsilon S_t.
\end{equation*}
Dividing by $S_t$, letting $t\to\infty$, and then letting $\varepsilon\downarrow 0$ give $Q_t/S_t\to 0$.
Consequently, $\gamma_{2,\mathsf N}S_t-C_LQ_t-C_\rho>0$ for all sufficiently large $t$, and (ii) proves the stated limit.
\end{proof}

\paragraph{Power law step sizes.}
We now express the bounds in Lemma \ref{lem:momentum-margin-rates} in terms of the iteration number.

\begin{corollary}\label{cor:common-power-law-rates}
Suppose that the hypotheses of Lemma \ref{lem:descent} hold and that $0<\alpha\le 1$.
Use the constants $C_L,C_\rho$ constructed in the proof of Lemma \ref{lem:momentum-margin-rates}.
Then the following bounds hold.
\begin{enumerate}
  \item[(i)] There are an integer $t_{\alpha,\mathsf N}\ge 1$ and a constant $C_{\alpha,\mathsf N}>0$ such that, for every integer $t\ge t_{\alpha,\mathsf N}$, we have $W_t\ne 0$ and
  \begin{equation}\label{eq:common-power-law-margin-rates}
    0\le \gamma_{2,\mathsf N}-\rho_2\left(\frac{W_t}{\mathsf N(W_t)}\right)\le C_{\alpha,\mathsf N}\begin{cases}
      t^{-\alpha}, & 0<\alpha<1/2,\\
      (1+\log t)/\sqrt{t}, & \alpha=1/2,\\
      t^{-(1-\alpha)}, & 1/2<\alpha<1,\\
      1/\log(t+1), & \alpha=1.
    \end{cases}
  \end{equation}

  \item[(ii)] If $\alpha=1/2$, there are constants $C_1,C_2,C_3,C_4>0$ such that, for every integer $t\ge 1$,
  \begin{equation}\label{eq:common-square-root-envelopes}
    \max\{E_t,e^{-\rho_2(W_t)}\}\le C_1t^{C_2}e^{-C_3\sqrt{t}}.
  \end{equation}
  Whenever $\gamma_{2,\mathsf N}S_t-C_LQ_t-C_\rho>0$, we also have $W_t\ne 0$ and
  \begin{equation}\label{eq:common-square-root-margin-rate}
    0\le \gamma_{2,\mathsf N}-\rho_2\left(\frac{W_t}{\mathsf N(W_t)}\right)\le C_4\frac{1+\log t}{\sqrt{t}}.
  \end{equation}
\end{enumerate}
The constants and the iteration threshold may depend on $\Delta_T,\alpha,\beta,\mathsf N$, the finite training data, and $\eta_0$, but not on $t$ or the maximizing directions.
\end{corollary}

\begin{proof}
For every integer $t\ge 1$, termwise and integral comparison give
\begin{equation*}
  \begin{aligned}
    & S_t\ge \eta_0\begin{cases}
      t^{1-\alpha}, & 0<\alpha<1,\\
      \log(t+1), & \alpha=1,
    \end{cases}\\
    & Q_t\le \eta_0^2\begin{cases}
      \displaystyle 1+\frac{t^{1-2\alpha}-1}{1-2\alpha}, & 0<\alpha<1/2,\\[2mm]
      1+\log t, & \alpha=1/2,\\[1mm]
      \displaystyle \frac{2\alpha}{2\alpha-1}, & 1/2<\alpha\le 1.
    \end{cases}
  \end{aligned}
\end{equation*}
For $0<\alpha<1$, the lower bound on $S_t$ follows from $k^{-\alpha}\ge t^{-\alpha}$ for $1\le k\le t$.
The positivity condition in \eqref{eq:common-normalized-margin} holds for all sufficiently large $t$ by Lemma \ref{lem:momentum-margin-rates}.

Substituting the bounds on $S_t,Q_t$ into \eqref{eq:common-normalized-margin} gives the four rates in (i).
These sum bounds and the constants $C_L,C_\rho$ are independent of the maximizing directions, so $C_{\alpha,\mathsf N}$ and $t_{\alpha,\mathsf N}$ can also be chosen independently of them.

For (ii), let $\alpha=1/2$.
For $t\ge 1$, integral comparison and $\sqrt{t+1}+1\le (1+\sqrt{2})\sqrt{t}$ give
\begin{equation*}
  S_t\ge 2\eta_0(\sqrt{t+1}-1)=\frac{2\eta_0t}{\sqrt{t+1}+1}\ge \frac{2\eta_0}{1+\sqrt{2}}\sqrt{t}.
\end{equation*}
For any $\mathcal K\ge C_L$, let
\begin{equation}\label{eq:common-square-root-rate-constants}
  \begin{aligned}
    & C_1:=\exp(\mathcal K+|\log\nu_T|+C_2+C_3),\\
    & C_2:=\mathcal K\eta_0^2,\qquad C_3:=2\gamma_{2,\mathsf N}\eta_0,\\
    & C_4:=\frac{1+\sqrt{2}}{2\eta_0}\left(C_2+\mathcal K+|\log\nu_T|\right).
  \end{aligned}
\end{equation}
Lemma \ref{lem:descent} verifies the attention condition needed for Lemma \ref{lem:margin-loss}.
Together with (i) of Lemma \ref{lem:momentum-margin-rates}, this gives
\begin{equation*}
  \max\{\log E_t,-\rho_2(W_t)\}\le -\gamma_{2,\mathsf N}S_t+C_LQ_t+C_L+|\log\nu_T|.
\end{equation*}
Using $\mathcal K\ge C_L$, $Q_t\le\eta_0^2(1+\log t)$, and $S_t\ge 2\eta_0(\sqrt{t}-1)$, we obtain
\begin{equation*}
  \begin{aligned}
    \max\{\log E_t,-\rho_2(W_t)\} & \le -C_3\sqrt{t}+C_2\log t+\mathcal K+|\log\nu_T|+C_2+C_3\\
    & = \log C_1+C_2\log t-C_3\sqrt{t}.
  \end{aligned}
\end{equation*}
Exponentiating proves \eqref{eq:common-square-root-envelopes}.

Finally, since $C_\rho=C_L+|\log\nu_T|$, the bounds on $S_t,Q_t$ give
\begin{equation*}
    \frac{C_LQ_t+C_\rho}{S_t} \le \frac{1+\sqrt{2}}{2\eta_0}\frac{C_2(1+\log t)+\mathcal K+|\log\nu_T|}{\sqrt{t}}
    \le C_4\frac{1+\log t}{\sqrt{t}}.
\end{equation*}
The last inequality uses $1+\log t\ge 1$.
Whenever the positivity condition holds, \eqref{eq:common-normalized-margin} therefore proves \eqref{eq:common-square-root-margin-rate}.
\end{proof}

\begin{proof}[Proof of Theorem \ref{thm:main}]
Let $\mathsf N=\Vert\cdot\Vert_{\op}$, $\alpha=1/2$, $\beta=\beta_2$, and $\eta_0=\eta_{2,0}$, so that $\gamma_{2,\mathsf N}=\gamma_2$.
Lemma \ref{lem:exact-U} gives $CW_{1,T}=\Delta_TB$.
For $T>\bar T$, Corollary \ref{cor:preprocessing-barrier} ensures that $\Delta_T\ge 2N(K-1)\log K$ and $\Loss_T(0)\le \log K$.
The choice $\bar\eta_T=\bar\eta_{T,1/2,\Vert\cdot\Vert_{\op}}$ ensures the step size condition in Lemma \ref{lem:descent}.

For nonzero momentum, Muon's compact polar factor maximizes the momentum inner product over the spectral norm unit ball; when the momentum is zero, the prescribed zero direction is also a maximizer.
Thus Lemma \ref{lem:descent} and Corollary \ref{cor:common-power-law-rates} apply.

Let $C_L,C_\rho$ be the constants constructed in the proof of Lemma \ref{lem:momentum-margin-rates}.
That lemma gives $E_T(W_{2,t})>0$ at every finite iteration.
Take $\mathcal K=C_L$ in \eqref{eq:common-square-root-rate-constants} to obtain $C_1,\ldots,C_4$.
Then \eqref{eq:common-square-root-envelopes} proves \eqref{eq:thm2-loss} and also gives
\begin{equation}\label{eq:phase2-margin-exponential-envelope}
  e^{-\rho_2(W_{2,t})}\le C_1t^{C_2}e^{-C_3\sqrt{t}},\qquad t\ge 1.
\end{equation}
We use the same $C_1,C_2,C_3$ in the proof of Theorem \ref{thm:generalization}.

By Lemma \ref{lem:momentum-margin-rates}, there is an integer $t_T\ge 1$ such that $\gamma_2S_t-C_LQ_t-C_\rho>0$ for every $t\ge t_T$.
Hence $W_{2,t}\ne 0$ for such $t$, and \eqref{eq:common-square-root-margin-rate} proves \eqref{eq:thm2-margin}.
The stated dependence of $C_1,\ldots,C_4$ follows from Lemma \ref{lem:momentum-margin-rates} and their definitions in \eqref{eq:common-square-root-rate-constants}.
The threshold $t_T$ has the stated dependence because its defining condition involves only $C_L,C_\rho,S_t,Q_t$.

Finally, $C_3>0$ and $\log t/\sqrt{t}\to 0$ imply $E_T(W_{2,t})\to 0$, and therefore $\Loss_T(W_{2,t})\to\ell_T^\star$.
The normalized margin bound also gives $\rho_2(W_{2,t}/\Vert W_{2,t}\Vert_{\op})\to\gamma_2$.
\end{proof}

\subsection{Prediction: proof of Theorem \ref{thm:generalization}}
We sum the training score gaps along the assumed chains to bound the attention assigned to other tokens, and then derive the prediction probability bound.

\begin{proof}
Let $W_t:=W_{2,t}$ and $\zeta_t:=C_1t^{C_2}e^{-C_3\sqrt{t}}$.
For all sufficiently large $t$, Theorem \ref{thm:main} gives $\rho_2(W_t)>0$, and \eqref{eq:phase2-margin-exponential-envelope} gives $e^{-\rho_2(W_t)}\le\zeta_t$.
Fix an unseen sample $(X,x_{L+1})$ satisfying the hypotheses of Theorem \ref{thm:generalization}, with $X=[x_1,\ldots,x_L]$ and $q=x_L$.

For each position $\ell\in[L]$ with $x_\ell\ne x_X^\star$, choose a chain $u_0,\ldots,u_r$ with $r\ge 1$, $u_0=x_X^\star$, $u_r=x_\ell$, and $(u_{j-1},u_j)\in\mathcal C_q$ for $j\in[r]$.
Each pair belongs to the training constraints with query $q$, so $(u_{j-1}-u_j)^\top W_tq\ge\rho_2(W_t)$.
Summing along the chain and using $\rho_2(W_t)>0$ give
\begin{equation*}
  (x_X^\star-x_\ell)^\top W_tq=\sum_{j=1}^r(u_{j-1}-u_j)^\top W_tq\ge r\rho_2(W_t)\ge\rho_2(W_t).
\end{equation*}
Consequently, $e^{-(x_X^\star-x_\ell)^\top W_tq}\le\zeta_t$ for every position containing another token.

Let $\alpha_t:=\softmax(X^\top W_tq)$, and let $a:=\sum_{\ell:x_\ell=x_X^\star}[\alpha_t]_\ell$ be the total attention weight assigned to the correct token.
The correct token occurs at least once, and at most $L-1$ positions contain other tokens.
Thus
\begin{equation*}
  \frac{1-a}{a}=\frac{\sum_{\ell:x_\ell\ne x_X^\star}e^{-(x_X^\star-x_\ell)^\top W_tq}}{|\{\ell\in[L]:x_\ell=x_X^\star\}|}\le (L-1)\zeta_t.
\end{equation*}
Since $a\le 1$, this implies $1-a\le (L-1)\zeta_t$.

Let $y:=I(x_{L+1})=h(x_X^\star)$, and let $\widetilde a\in\R^K$ collect the attention weights by output label:
\begin{equation*}
  \widetilde a_v:=\sum_{\ell\in[L]:h(x_\ell)=v}[\alpha_t]_\ell,\qquad v\in[K].
\end{equation*}
The coordinates of $\widetilde a$ are nonnegative and sum to one.
Since $h$ is bijective, $\widetilde a_y=a$.
Lemma \ref{lem:exact-U} gives $CW_{1,T}=\Delta_TB$.
Using $Bx=e_{h(x)}-K^{-1}\mathbf1$ for $x\in\V$, we obtain
\begin{equation*}
  CW_{1,T}X\alpha_t=\Delta_T\left(\widetilde a-K^{-1}\mathbf1\right).
\end{equation*}
Adding the same constant to all logits leaves the softmax probabilities unchanged.
Hence the prediction probability is $g(\widetilde a)$, where $g(u):=[\softmax(\Delta_Tu)]_y$.

For $u\in\R^K$, let $p:=\softmax(\Delta_Tu)$.
The softmax derivative and $p_v\le 1-p_y$ for $v\ne y$ give
\begin{equation*}
  \Vert\nabla g(u)\Vert_\infty=\Delta_Tp_y(1-p_y)\le \frac{\Delta_T}{4}.
\end{equation*}
Moreover, \eqref{eq:thm1-loss} gives
\begin{equation*}
  g(e_y)=\left[1+(K-1)e^{-\Delta_T}\right]^{-1}=e^{-\ell_T^\star}.
\end{equation*}
Since $\Vert\widetilde a-e_y\Vert_1=2(1-a)$, integrating the gradient bound along the segment from $e_y$ to $\widetilde a$ yields
\begin{equation*}
  \begin{aligned}
    \mathbb P_{(W_{1,T},W_{2,t})}\{x_{L+1}\mid X\} & = g(\widetilde a)\\
    & \ge g(e_y)-\frac{\Delta_T}{4}\Vert\widetilde a-e_y\Vert_1\\
    & = e^{-\ell_T^\star}-\frac{\Delta_T}{2}(1-a)\\
    & \ge e^{-\ell_T^\star}-\frac{\Delta_T(L-1)}{2}\zeta_t.
  \end{aligned}
\end{equation*}
Substituting $\zeta_t=C_1t^{C_2}e^{-C_3\sqrt{t}}$ gives \eqref{eq:generalization-probability-lower-bound}.
\end{proof}

\section{Convergence of Schatten Norm Updates}\label{app:schatten-p-nmd}

Fix $1\le p<\infty$, and let $p'$ be its H\"older conjugate, with $p'=\infty$ when $p=1$.
Both phases use the Schatten $S_p$ norm.
For a momentum matrix $H$, let
\begin{equation*}
  \mathcal D_p(H):=\operatorname*{argmax}_{\Vert D\Vert_{S_p}\le 1}\ip{H}{D}.
\end{equation*}
Schatten norm duality gives $\ip{H}{D}=\Vert H\Vert_{S_{p'}}$ for every $D\in\mathcal D_p(H)$.
Algorithm \ref{alg:schatten-two-phase} allows any exact maximizing direction when the momentum is nonzero and uses the zero direction otherwise.

\begin{algorithm}[!b]
\caption{Two Phase Schatten $S_p$ Training}
\label{alg:schatten-two-phase}
\begin{algorithmic}[1]
\STATE \textbf{\textsc{Phase I}: Preprocessing}
\STATE \textbf{Input} $p$; $\D_0$; $T$; $\{\eta_{1,s}\}_{s=0}^{T-1}$; $\beta_1\in[0,1)$.
\STATE \textbf{Initialization} $W_{1,0}=0$; $M_{1,-1}=0$.
\FOR{$s=0,\ldots,T-1$}
  \STATE $M_{1,s}=\beta_1M_{1,s-1}+(1-\beta_1)\nabla\Loss_0(W_{1,s})$.
  \IF{$M_{1,s}=0$}
    \STATE $D_{1,s}= 0$.
  \ELSE
    \STATE $D_{1,s}\in\mathcal D_p(M_{1,s})$.
  \ENDIF
  \STATE $W_{1,s+1}=W_{1,s}-\eta_{1,s}D_{1,s}$.
\ENDFOR
\par\vspace{0.4ex}\hrule\vspace{0.4ex}
\STATE \textbf{\textsc{Phase II}: Main Training}
\STATE \textbf{Input} $\D$; $W_{1,T}$; $t$; $\{\eta_{2,r}\}_{r=0}^{t-1}$; $\beta_2\in[0,1)$.
\STATE \textbf{Loss} $\Loss_T(W):=\Loss(W_{1,T},W)$.
\STATE \textbf{Initialization} $W_{2,0}=0$; $M_{2,-1}=0$.
\FOR{$r=0,\ldots,t-1$}
  \STATE $M_{2,r}=\beta_2M_{2,r-1}+(1-\beta_2)\nabla\Loss_T(W_{2,r})$.
  \IF{$M_{2,r}=0$}
    \STATE $D_{2,r}=0$.
  \ELSE
    \STATE $D_{2,r}\in\mathcal D_p(M_{2,r})$.
  \ENDIF
  \STATE $W_{2,r+1}=W_{2,r}-\eta_{2,r}D_{2,r}$.
\ENDFOR
\STATE \textbf{Output} $(W_{1,T},W_{2,t})$.
\end{algorithmic}
\end{algorithm}

\subsection*{\textbf{Preprocessing phase}}

We use the notation of Appendix \ref{app:preprocessing-proof}.
Recall that
\begin{equation*}
  BB^\top=C,\qquad CB=B,\qquad \sum_{i=1}^K\sum_{v\ne h(x_i)}C_{x_i,v}=KB.
\end{equation*}
The $K-1$ nonzero singular values of $B$ are all one.
Let $\gamma_{1,p}:=(K-1)^{-1/p}$.

The next lemma identifies the maximum preprocessing margin and bounds the distance to $\gamma_{1,p}B$ in terms of the margin gap.
We will use it to prove convergence of the normalized preprocessing iterates.

\begin{lemma}\label{lem:schatten-phase1-stability}
Suppose that $K\ge 2$, $h$ is bijective, and the vocabulary embeddings are orthonormal.
Then
\begin{equation}\label{eq:gen-phase1-gamma}
  \max_{\Vert V\Vert_{S_p}\le 1}\rho_1(V)=\gamma_{1,p}.
\end{equation}
There is a constant $C_{p,K}>0$, depending only on $p,K$, such that, for every $\varepsilon\ge 0$ and every $V\in\R^{K\times d}$ satisfying
\begin{equation*}
  CV=V,\qquad \Vert V\Vert_{S_p}\le 1,\qquad \rho_1(V)\ge\gamma_{1,p}-\varepsilon,
\end{equation*}
we have
\begin{equation*}
  \Vert V-\gamma_{1,p}B\Vert_{S_p}\le C_{p,K}\sqrt{\varepsilon}.
\end{equation*}
\end{lemma}

\begin{proof}
Let $r:=K-1$, $\gamma:=\gamma_{1,p}$, and $y_i:=h(x_i)$.
Since $B$ has $r$ unit singular values, $\Vert\gamma B\Vert_{S_p}=\gamma r^{1/p}=1$.
Also, $Bx_i=e_{y_i}-K^{-1}\mathbf1$, so every preprocessing margin of $\gamma B$ equals $\gamma$.

For any $V$ with $\Vert V\Vert_{S_p}\le 1$, let $g_{i,v}:=(e_{y_i}-e_v)^\top Vx_i$ for $i\in[K]$ and $v\in[K]\setminus\{y_i\}$.
Averaging these $Kr$ margins and using the identity $\sum_{i=1}^K\sum_{v\ne y_i}C_{x_i,v}=KB$ and Schatten norm duality give
\begin{equation*}
  \rho_1(V)\le \frac{1}{Kr}\sum_{i=1}^K\sum_{v\ne y_i}g_{i,v}=\frac{\ip{V}{B}}{r}\le \frac{\Vert V\Vert_{S_p}\Vert B\Vert_{S_{p'}}}{r}\le \gamma.
\end{equation*}
This proves \eqref{eq:gen-phase1-gamma}.

To prove the stability bound, let $V$ satisfy its assumptions.
We first bound $\Vert VX_{\V}-\gamma PC\Vert_F$ using the margin assumptions.
The margin lower bound and the averaged estimate give
\begin{equation*}
  g_{i,v}-\gamma\ge-\varepsilon,\qquad \sum_{i=1}^K\sum_{v\ne y_i}(g_{i,v}-\gamma)\le 0.
\end{equation*}
For any one term, the other $Kr-1$ terms have sum at least $-(Kr-1)\varepsilon$.
Consequently,
\begin{equation*}
  |g_{i,v}-\gamma|\le Kr\varepsilon,\qquad i\in[K],\quad v\ne y_i.
\end{equation*}

Let $A:=VX_{\V}$.
Since $CA=A$, each column of $A$ has coordinate sum zero.
Its entries therefore satisfy
\begin{equation*}
  A_{y_i,i}=\frac{1}{K}\sum_{v\ne y_i}g_{i,v},\qquad A_{v,i}=A_{y_i,i}-g_{i,v}\quad(v\ne y_i).
\end{equation*}
The corresponding entries of $\gamma PC$ are $\gamma r/K$ at row $y_i$ and $-\gamma/K$ elsewhere.
Thus every entry of $A-\gamma PC$ has absolute value at most $2Kr\varepsilon$, giving
\begin{equation*}
  \Vert A-\gamma PC\Vert_F\le L_K\varepsilon,\qquad L_K:=2K^2r.
\end{equation*}

We next use the Schatten norm constraint to control the component outside the vocabulary span.
Let $H:=V-AX_{\V}^\top$.
Orthonormality and $CV=V$ give
\begin{equation*}
  HX_{\V}=0,\qquad CH=H,\qquad VV^\top=AA^\top+HH^\top.
\end{equation*}
Let $\varepsilon_0:=\min\{1,\gamma/(2L_K)\}$.
First suppose that $0\le \varepsilon\le \varepsilon_0$, and let $a:=\gamma-L_K\varepsilon\ge\gamma/2$.
Since $P$ preserves $\mathbf1^\perp$ and $\Vert A-\gamma PC\Vert_{\op}\le L_K\varepsilon$, every $z\in\mathbf1^\perp$ satisfies
\begin{equation*}
  \Vert A^\top z\Vert_2\ge\gamma\Vert z\Vert_2-\Vert(A-\gamma PC)^\top z\Vert_2\ge a\Vert z\Vert_2.
\end{equation*}
Together with $A^\top\mathbf1=0$, this gives $AA^\top\succeq a^2C$.

Let $\lambda_1,\ldots,\lambda_r$ be the eigenvalues of $VV^\top$ on $\mathbf1^\perp$.
Since $VV^\top\succeq AA^\top\succeq a^2C$, we have $\lambda_i\ge a^2$.
The Schatten norm constraint also gives
\begin{equation*}
  \lambda_i\le 1,\qquad \sum_{i=1}^r\lambda_i^{p/2}\le 1.
\end{equation*}
Let $m_{p,K}>0$, depending only on $p,K$, be a lower bound for the derivative of $u^{p/2}$ on $[(\gamma/2)^2,1]$.
The mean value theorem gives
\begin{equation*}
  \lambda_i^{p/2}-a^p\ge m_{p,K}(\lambda_i-a^2),\qquad i\in[r].
\end{equation*}
Taking traces in $VV^\top=AA^\top+HH^\top$ and using $AA^\top\succeq a^2C$ yield
\begin{equation*}
  \sum_{i=1}^r(\lambda_i-a^2)=\Vert A\Vert_F^2+\Vert H\Vert_F^2-ra^2\ge\Vert H\Vert_F^2.
\end{equation*}
Summing the bounds on $\lambda_i^{p/2}-a^p$ and using $r\gamma^p=1$, we obtain
\begin{equation*}
  \begin{aligned}
    m_{p,K}\Vert H\Vert_F^2 & \le m_{p,K}\sum_{i=1}^r(\lambda_i-a^2)\\
    & \le \sum_{i=1}^r(\lambda_i^{p/2}-a^p)\\
    & \le 1-ra^p\\
    & = r(\gamma^p-a^p)\\
    & \le rp\gamma^{p-1}L_K\varepsilon.
  \end{aligned}
\end{equation*}
The last inequality follows from the mean value theorem, since $p\ge 1$, $a\le\gamma$, and $\gamma-a=L_K\varepsilon$.

Since $B=PCX_{\V}^\top$ and $HX_{\V}=0$, the matrices $(A-\gamma PC)X_{\V}^\top$ and $H$ are orthogonal in the Frobenius inner product.
Using $\varepsilon\le \varepsilon_0\le 1$, we therefore have
\begin{equation*}
  \begin{aligned}
    \Vert V-\gamma B\Vert_F^2 & = \Vert A-\gamma PC\Vert_F^2+\Vert H\Vert_F^2\\
    & \le L_K^2\varepsilon^2+\frac{rp\gamma^{p-1}L_K}{m_{p,K}}\varepsilon\\
    & \le \left(L_K^2+\frac{rp\gamma^{p-1}L_K}{m_{p,K}}\right)\varepsilon.
  \end{aligned}
\end{equation*}
Since $C(V-\gamma B)=V-\gamma B$, this matrix has rank at most $r$.
Comparison of its singular value norms yields
\begin{equation*}
  \Vert V-\gamma B\Vert_{S_p}\le r^{\max\{1/p-1/2,0\}}\Vert V-\gamma B\Vert_F\le C_{p,K}\sqrt{\varepsilon},
\end{equation*}
where $C_{p,K}$ depends only on $p,K$.

For $\varepsilon>\varepsilon_0$, the triangle inequality and the norm constraint give
\begin{equation*}
  \Vert V-\gamma B\Vert_{S_p}\le \Vert V\Vert_{S_p}+\Vert\gamma B\Vert_{S_p}\le 2\le \frac{2}{\sqrt{\varepsilon_0}}\sqrt{\varepsilon}.
\end{equation*}
Enlarge $C_{p,K}$, if necessary, so that $C_{p,K}\ge 2/\sqrt{\varepsilon_0}$.
This preserves the bound for $\varepsilon\le \varepsilon_0$ and gives the required estimate for all $\varepsilon\ge 0$.
The enlarged constant depends only on $p,K$, since $\varepsilon_0$ does.
\end{proof}

For $y_i:=h(x_i)$, let $p_i(W):=\softmax(Wx_i)$ and $\ell_i(W):=-\log[p_i(W)]_{y_i}$.
For the preprocessing iterates, write $L_s:=\Loss_0(W_{1,s})$ and $q_s:=q(W_{1,s})$, where
\begin{equation*}
  q(W):=\frac{1}{K}\sum_{i=1}^K\left(1-[p_i(W)]_{y_i}\right).
\end{equation*}
The next lemma bounds the loss decrease in terms of $q_s$.

\begin{lemma}\label{lem:schatten-phase1-descent}
Under the assumptions of Lemma \ref{lem:schatten-phase1-stability}, consider Phase I of Algorithm \ref{alg:schatten-two-phase} with
\begin{equation*}
  \eta_{1,s}=\eta_{1,0}(s+1)^{-\alpha_1},\qquad \eta_{1,0}>0,\qquad 0<\alpha_1\le 1.
\end{equation*}
There is a constant $A_{\mathrm D}>0$, depending only on $K,p,\alpha_1,\beta_1,\eta_{1,0}$, such that, whenever $\eta_{1,s}\le 1$,
\begin{equation}\label{eq:schatten-phase1-descent}
  L_{s+1}-L_s\le -\eta_{1,s}(\gamma_{1,p}-A_{\mathrm D}\eta_{1,s})q_s.
\end{equation}
\end{lemma}

\begin{proof}
Let $\gamma:=\gamma_{1,p}$, $\eta_s:=\eta_{1,s}$, and $G_s:=\nabla\Loss_0(W_{1,s})$.
Recall that $C_{x_i,v}=(e_{y_i}-e_v)x_i^\top$ for $v\ne y_i$.
The gradient expansion gives
\begin{equation*}
  -\nabla\Loss_0(W)=\frac{1}{K}\sum_{i=1}^K\sum_{v\ne y_i}[p_i(W)]_vC_{x_i,v}.
\end{equation*}
Since $\ip{B}{C_{x_i,v}}=1$ and $\Vert\gamma B\Vert_{S_p}=1$, norm duality yields
\begin{equation*}
  \Vert\nabla\Loss_0(W)\Vert_{S_{p'}}\ge-\ip{\nabla\Loss_0(W)}{\gamma B}=\gamma q(W).
\end{equation*}

For $d_W:=\Vert W'-W\Vert_{S_p}$, each output logit changes by at most $d_W$.
The corresponding softmax probability ratios lie in $[e^{-2d_W},e^{2d_W}]$, so
\begin{equation*}
  |[p_i(W')]_v-[p_i(W)]_v|\le (e^{2d_W}-1)[p_i(W)]_v,\qquad i,v\in[K].
\end{equation*}
Using the gradient expansion and $\Vert C_{x_i,v}\Vert_{S_{p'}}=\sqrt{2}$, we obtain
\begin{equation*}
    \Vert\nabla\Loss_0(W')-\nabla\Loss_0(W)\Vert_{S_{p'}} \le \frac{\sqrt{2}}{K}\sum_{i=1}^K\sum_{v\ne y_i}|[p_i(W')]_v-[p_i(W)]_v|
    \le \sqrt{2}(e^{2d_W}-1)q(W).
\end{equation*}

Let $\widehat M_{1,s}:=M_{1,s}/(1-\beta_1^{s+1})$.
For $0<\beta_1<1$, let
\begin{equation*}
  w_{s,r}:=\frac{(1-\beta_1)\beta_1^r}{1-\beta_1^{s+1}},\qquad 0\le r\le s.
\end{equation*}
Expanding the momentum recursion and using $\sum_{r=0}^sw_{s,r}=1$ give
\begin{equation*}
  \widehat M_{1,s}-G_s=\sum_{r=0}^sw_{s,r}(G_{s-r}-G_s).
\end{equation*}
Since $\Vert D_{1,j}\Vert_{S_p}\le 1$, summing the parameter updates gives
\begin{equation*}
  \Vert W_{1,s}-W_{1,s-r}\Vert_{S_p}\le \sum_{j=1}^r\eta_{s-j}.
\end{equation*}
The gradient variation bound and Lemma \ref{lem:ema-phase2} therefore imply
\begin{equation*}
  \begin{aligned}
    \Vert\widehat M_{1,s}-G_s\Vert_{S_{p'}} & \le \sum_{r=0}^sw_{s,r}\Vert G_{s-r}-G_s\Vert_{S_{p'}}\\
    & \le \sqrt{2}q_s\sum_{r=0}^sw_{s,r}\left[\exp\left(2\sum_{j=1}^r\eta_{s-j}\right)-1\right]\\
    & \le A_{\mathrm M}\eta_sq_s,
  \end{aligned}
\end{equation*}
where
\begin{equation*}
  A_{\mathrm M}:=\frac{\sqrt{2}C(\alpha_1,2\eta_{1,0},\beta_1)}{\eta_{1,0}}.
\end{equation*}
For the last inequality, we applied Lemma \ref{lem:ema-phase2} to the sequence $(j+1)^{-\alpha_1}$ with exponential parameter $2\eta_{1,0}$.
Its upper bound is $C(\alpha_1,2\eta_{1,0},\beta_1)(s+1)^{-\alpha_1}=C(\alpha_1,2\eta_{1,0},\beta_1)\eta_s/\eta_{1,0}$.
When $\beta_1=0$, we have $\widehat M_{1,s}=G_s$ and $A_{\mathrm M}=0$, so the same error bound holds.

The maximizing property of $D_{1,s}$ and the duality argument in the proof of Lemma \ref{lem:first-order} give
\begin{equation*}
  \ip{G_s}{D_{1,s}}\ge\Vert G_s\Vert_{S_{p'}}-2\Vert\widehat M_{1,s}-G_s\Vert_{S_{p'}}\ge(\gamma-2A_{\mathrm M}\eta_s)q_s.
\end{equation*}
Let $A_{\mathrm D}:=2A_{\mathrm M}+\sqrt{2}e^2$.
For $\eta_s\le 1$, integration along the update segment yields
\begin{equation*}
  \begin{aligned}
    L_{s+1}-L_s & \le -\eta_s\ip{G_s}{D_{1,s}}+\sqrt{2}q_s\int_0^{\eta_s}(e^{2\tau}-1)\,d\tau\\
    & \le -\eta_s(\gamma-2A_{\mathrm M}\eta_s)q_s+\sqrt{2}e^2\eta_s^2q_s\\
    & = -\eta_s(\gamma-A_{\mathrm D}\eta_s)q_s,
  \end{aligned}
\end{equation*}
where the second inequality uses $e^{2\tau}-1\le 2e^2\tau$ for $0\le\tau\le 1$.
\end{proof}

We next use the descent estimate to obtain loss and margin bounds, then apply Lemma \ref{lem:schatten-phase1-stability} to the normalized iterates.

\begin{lemma}\label{lem:schatten-phase1-asymptotics}
Under the hypotheses of Lemma \ref{lem:schatten-phase1-descent}, let $S_{1,s}:=\sum_{j=0}^{s-1}\eta_{1,j}$.
In the decomposition
\begin{equation*}
  CW_{1,s}=\delta_sB+R_s,\qquad \delta_s:=\frac{\ip{CW_{1,s}}{B}}{K-1},\qquad \ip{R_s}{B}=0,
\end{equation*}
we have $\delta_s>0$ eventually and
\begin{equation}\label{eq:schatten-phase1-output-limits}
  \frac{\delta_s}{S_{1,s}}\longrightarrow\gamma_{1,p},\qquad \frac{\Vert R_s\Vert_{S_p}}{\delta_s}\longrightarrow0.
\end{equation}
\end{lemma}

\begin{proof}
Let $\gamma:=\gamma_{1,p}$, $\eta_s:=\eta_{1,s}$, and $Q_{1,s}:=\sum_{j=0}^{s-1}\eta_j^2$.
Since $\eta_s\to0$, choose an integer $s_0\ge 0$, depending only on the fixed parameters, such that
\begin{equation*}
  \eta_s\le \min\{1,\gamma/(2A_{\mathrm D})\},\qquad s\ge s_0.
\end{equation*}
By \eqref{eq:schatten-phase1-descent},
\begin{equation*}
  L_{s+1}-L_s\le -\frac{\gamma}{2}\eta_sq_s\le 0,\qquad s\ge s_0.
\end{equation*}

Let $B_0:=\log K+2S_{1,s_0}$.
Since $W_{1,0}=0$ and $\Vert D_{1,j}\Vert_{S_p}\le 1$, we have $\Vert W_{1,s}\Vert_{S_p}\le S_{1,s}$ and hence $\ell_i(W_{1,s})\le B_0$ for $s\le s_0$.
For $s\ge s_0$, monotonicity gives $L_s\le L_{s_0}\le B_0$ and $\ell_i(W_{1,s})\le KL_s$.
Thus
\begin{equation*}
  L_s\le B_0,\qquad 0\le\ell_i(W_{1,s})\le KB_0,\qquad i\in[K],\quad s\ge 0.
\end{equation*}
On $[0,KB_0]$, there is a constant $c>0$, depending only on the fixed parameters, such that $1-e^{-u}\ge cu$.
Since $1-[p_i(W_{1,s})]_{y_i}=1-e^{-\ell_i(W_{1,s})}$, averaging gives $q_s\ge cL_s$.
Summing the descent estimate yields
\begin{equation*}
  \frac{c\gamma}{2}\sum_{j=s_0}^{s-1}\eta_jL_j\le L_{s_0}-L_s\le L_{s_0}.
\end{equation*}
For every integer $s\ge s_0$, splitting the sum at $s_0$ gives
\begin{equation*}
  \begin{aligned}
    \sum_{j=0}^{s-1}\eta_jL_j & = \sum_{j=0}^{s_0-1}\eta_jL_j+\sum_{j=s_0}^{s-1}\eta_jL_j\\
    & \le B_0\sum_{j=0}^{s_0-1}\eta_j+\frac{2L_{s_0}}{c\gamma}\\
    & \le B_0S_{1,s_0}+\frac{2B_0}{c\gamma}\\
    & = B_0\left(S_{1,s_0}+\frac{2}{c\gamma}\right).
  \end{aligned}
\end{equation*}
Here we used $L_j\le B_0$ for the first sum and the preceding descent estimate for the second.
Since all summands are nonnegative, letting $s\to\infty$ yields
\begin{equation*}
  \sum_{j=0}^{\infty}\eta_jL_j\le B_0\left(S_{1,s_0}+\frac{2}{c\gamma}\right)<\infty.
\end{equation*}

To obtain the logarithmic loss estimate with coefficient $\gamma$, we use $1-e^{-u}\ge u-u^2/2$ for $u\ge 0$:
\begin{equation*}
  q_s\ge L_s-\frac{1}{2K}\sum_{i=1}^K\ell_i(W_{1,s})^2\ge L_s-\frac{K}{2}L_s^2.
\end{equation*}
Every finite iterate has finite logits, so $L_s>0$.
Dividing \eqref{eq:schatten-phase1-descent} by $L_s$ therefore gives a positive upper bound on $L_{s+1}/L_s$.
For $s\ge s_0$, taking logarithms and using $\log(1+v)\le v$ give
\begin{equation*}
  \begin{aligned}
    \log L_s & \le \log L_{s_0}-\sum_{j=s_0}^{s-1}\eta_j(\gamma-A_{\mathrm D}\eta_j)\frac{q_j}{L_j}\\
    & \le \log L_{s_0}-\gamma(S_{1,s}-S_{1,s_0})+A_{\mathrm D}(Q_{1,s}-Q_{1,s_0})+\frac{\gamma K}{2}\sum_{j=s_0}^{s-1}\eta_jL_j.
  \end{aligned}
\end{equation*}
For the second inequality, we use $q_j/L_j\ge 1-KL_j/2$ and $\gamma-A_{\mathrm D}\eta_j>0$ to obtain
\begin{equation*}
  \begin{aligned}
    (\gamma-A_{\mathrm D}\eta_j)\frac{q_j}{L_j} & \ge (\gamma-A_{\mathrm D}\eta_j)\left(1-\frac{K}{2}L_j\right)\\
    & = \gamma-A_{\mathrm D}\eta_j-\frac{\gamma K}{2}L_j+\frac{A_{\mathrm D}K}{2}\eta_jL_j\\
    & \ge \gamma-A_{\mathrm D}\eta_j-\frac{\gamma K}{2}L_j.
  \end{aligned}
\end{equation*}
The weighted loss sum is uniformly bounded, and $\log L_{s_0}\le\log B_0$.
Hence all terms other than $-\gamma S_{1,s}+A_{\mathrm D}Q_{1,s}$ are bounded above by a constant depending only on the fixed parameters.

For each $i\in[K]$, the sample loss satisfies
\begin{equation*}
  e^{\ell_i(W)}-1=\sum_{v\ne y_i}e^{-(e_{y_i}-e_v)^\top Wx_i}.
\end{equation*}
By the definition of $\rho_1$ and the bound $\ell_i(W_{1,s})\le KL_s$, we obtain
\begin{equation*}
  \begin{aligned}
    e^{-\rho_1(W_{1,s})} & = \max_{i\in[K],\,v\ne y_i}e^{-(e_{y_i}-e_v)^\top W_{1,s}x_i}\\
    & \le \max_{i\in[K]}\left(e^{\ell_i(W_{1,s})}-1\right)\\
    & \le e^{KL_s}-1\\
    & \le Ke^{KB_0}L_s.
  \end{aligned}
\end{equation*}
The last inequality follows from the mean value theorem and $L_s\le B_0$.

Taking logarithms in the preceding margin bound gives
\begin{equation*}
  \rho_1(W_{1,s})\ge-\log L_s-\log K-KB_0.
\end{equation*}
Combining this with the logarithmic loss estimate, we may take $C_1:=A_{\mathrm D}$ and choose $C_0>0$ such that, for every $s\ge 0$,
\begin{equation*}
  \begin{aligned}
    & 0<L_s\le \exp(C_0-\gamma S_{1,s}+C_1Q_{1,s}),\\
    & \rho_1(W_{1,s})\ge\gamma S_{1,s}-C_1Q_{1,s}-C_0.
  \end{aligned}
\end{equation*}
For $0\le s<s_0$, we have $L_s\le B_0$ and $\rho_1(W_{1,s})\ge-2S_{1,s_0}$.
Increasing $C_0$, if necessary, makes both estimates valid on this interval as well.
The constants depend only on $K,p,\alpha_1,\beta_1,\eta_{1,0}$.

For $s\ge 1$, let
\begin{equation*}
  V_s:=\frac{CW_{1,s}}{S_{1,s}},\qquad e_s:=C_0+C_1Q_{1,s}.
\end{equation*}
Multiplication by $C$ does not increase the Schatten norm and leaves all preprocessing margins unchanged.
Together with $C^2=C$ and the preceding bounds, this gives
\begin{equation*}
  CV_s=V_s,\qquad \Vert V_s\Vert_{S_p}\le 1,\qquad \rho_1(V_s)\ge\gamma-\frac{e_s}{S_{1,s}}.
\end{equation*}
Since $0<\alpha_1\le 1$, we have $S_{1,s}\to\infty$.
The argument at the end of Lemma \ref{lem:momentum-margin-rates} also gives $Q_{1,s}/S_{1,s}\to0$.
Thus $e_s/S_{1,s}\to0$, and Lemma \ref{lem:schatten-phase1-stability} yields
\begin{equation*}
  \Vert V_s-\gamma B\Vert_{S_p}\le C_{p,K}\sqrt{\frac{e_s}{S_{1,s}}}\longrightarrow0.
\end{equation*}

Finally, the definition of $\delta_s$ and Schatten norm duality give
\begin{equation*}
  \left|\frac{\delta_s}{S_{1,s}}-\gamma\right|=\frac{|\ip{V_s-\gamma B}{B}|}{K-1}\le \gamma\Vert V_s-\gamma B\Vert_{S_p},
\end{equation*}
where $\Vert B\Vert_{S_{p'}}/(K-1)=\gamma$.
Since $R_s/S_{1,s}=V_s-(\delta_s/S_{1,s})B$ and $\gamma\Vert B\Vert_{S_p}=1$, we also have
\begin{equation*}
  \frac{\Vert R_s\Vert_{S_p}}{S_{1,s}}\le \Vert V_s-\gamma B\Vert_{S_p}+\left|\frac{\delta_s}{S_{1,s}}-\gamma\right|\Vert B\Vert_{S_p}\le 2\Vert V_s-\gamma B\Vert_{S_p}.
\end{equation*}
For all sufficiently large $s$, the first bound gives $\delta_s/S_{1,s}\ge\gamma/2>0$.
Therefore,
\begin{equation*}
  \frac{\Vert R_s\Vert_{S_p}}{\delta_s}\le \frac{4}{\gamma}\Vert V_s-\gamma B\Vert_{S_p}\longrightarrow0.
\end{equation*}
This proves \eqref{eq:schatten-phase1-output-limits}.
The bounds above are uniform over the maximizing directions and independent of $d$.
\end{proof}

\subsection*{\textbf{Main phase}}

Suppose that the data assumptions of Lemma \ref{lem:schatten-phase1-stability} hold.
For the full training data, suppose that Assumptions \ref{ass:token-label}, \ref{ass:weak-count}, and \ref{ass:separability} hold and $\I_2\ne\varnothing$.

\paragraph{Prediction matrix and loss.}
Let $U:=W_{1,T}$ be fixed during Phase II, and write
\begin{equation}\label{eq:fixed-output-decomposition}
  CU=\delta B+R,\qquad \delta:=\frac{\ip{CU}{B}}{K-1},\qquad \ip{R}{B}=0,\qquad r_U:=\Vert R\Vert_{\op}.
\end{equation}
By Lemma \ref{lem:schatten-phase1-asymptotics}, sufficiently long preprocessing ensures
\begin{equation}\label{eq:fixed-prediction-condition}
  \delta\ge 2(K-1)\left[N\log K+\sqrt{2}(N+1)r_U\right].
\end{equation}
In particular, $\delta>2r_U$.

For each sample $n\in[N]$, let $c:=x_n^\star$.
For $a\in\mathcal A_n$, let
\begin{equation*}
  m(a):=\sum_{x\in\V}a_xx,\qquad p_n(a):=\softmax(Um(a)).
\end{equation*}
The sample loss and its excess over the loss at $e_c$ are
\begin{equation*}
  f_{U,n}(a):=-\log[p_n(a)]_{y_n},\qquad F_{U,n}(a):=f_{U,n}(a)-f_{U,n}(e_c).
\end{equation*}
Let $\Loss_U(W):=\Loss(U,W)$ and
\begin{equation*}
  \ell_U^\star:=\frac{1}{N}\sum_{n=1}^Nf_{U,n}(e_{x_n^\star}),\qquad E_U(W):=\Loss_U(W)-\ell_U^\star.
\end{equation*}

For $x\in\X_n^-$, let
\begin{equation*}
  D_{n,x}^U(a):=\langle\nabla_af_{U,n}(a),e_x-e_c\rangle,\qquad \kappa_-:=\min_{(n,x)\in\I_2}D_{n,x}^U(e_{x_n^\star}).
\end{equation*}
Also let $Q_U:=\Vert CU\Vert_{\op}$.

The next two lemmas establish the loss and attention bounds needed for the common gradient estimates.

\begin{lemma}\label{lem:fixed-prediction-loss}
Suppose that $U$ satisfies \eqref{eq:fixed-prediction-condition}.
Then $D_{n,x}^U(a)>0$ for every $n\in[N]$, $a\in\mathcal A_n$, and $x\in\X_n^-$.
Moreover, $\kappa_->0$ and
\begin{equation}\label{eq:fixed-excess-attention}
  0<\kappa_-\bar s(W)\le E_U(W)\le 2Q_U\bar s(W),\qquad W\in\R^{d\times d}.
\end{equation}
The training loss satisfies $\inf_W\Loss_U(W)=\ell_U^\star$.
\end{lemma}

\begin{proof}
Fix $n\in[N]$, $a\in\mathcal A_n$, and $x\in\X_n^-$.
Let $y:=y_n$, $k:=h(x)$, and $q:=1-[p_n(a)]_y$.
The cross entropy derivative and $B(x-c)=e_k-e_y$ give
\begin{equation*}
  D_{n,x}^U(a)=(p_n(a)-e_y)^\top CU(x-c)=\delta\left([p_n(a)]_k+q\right)+(p_n(a)-e_y)^\top R(x-c).
\end{equation*}
Since $\Vert p_n(a)-e_y\Vert_2\le \sqrt{2}q$ and $\Vert x-c\Vert_2=\sqrt{2}$,
\begin{equation*}
  |(p_n(a)-e_y)^\top R(x-c)|\le \Vert p_n(a)-e_y\Vert_2r_U\Vert x-c\Vert_2\le 2r_Uq.
\end{equation*}
The logits are finite and $K\ge 2$, so $q>0$.
Using $\delta>2r_U$ and $[p_n(a)]_k\ge 0$, we obtain
\begin{equation*}
  D_{n,x}^U(a)\ge(\delta-2r_U)q>0.
\end{equation*}
Since $\I_2$ is finite and nonempty, the minimum defining $\kappa_-$ is positive.

The function $f_{U,n}$ is convex, so
\begin{equation*}
  F_{U,n}(a)\ge \langle\nabla_af_{U,n}(e_c),a-e_c\rangle=\sum_{x\in\X_n^-}a_xD_{n,x}^U(e_c)\ge \kappa_-s_n(a).
\end{equation*}
Since $\Vert p_n(a)-e_y\Vert_2\le \sqrt{2}q\le \sqrt{2}$ and $\Vert x-c\Vert_2=\sqrt{2}$, the derivative formula gives
\begin{equation*}
  D_{n,x}^U(a)\le \Vert p_n(a)-e_y\Vert_2Q_U\Vert x-c\Vert_2\le 2Q_U.
\end{equation*}
Integrating along the segment from $e_c$ to $a$, we obtain
\begin{equation*}
  F_{U,n}(a)=\int_0^1\sum_{x\in\X_n^-}a_xD_{n,x}^U(e_c+\tau(a-e_c))\,d\tau\le 2Q_Us_n(a).
\end{equation*}
Evaluating these bounds at $a_n(W)$ and averaging over the samples give the empirical bounds.
For finite $W$, every token present in a context has positive attention weight.
Since $\I_2\ne\varnothing$, we have $\bar s(W)>0$, proving the strict lower bound in \eqref{eq:fixed-excess-attention}.

By Assumption \ref{ass:separability}, choose $W^\star$ with $\ip{W^\star}{A_{n,x}}\ge 1$ for every $(n,x)\in\I_2$.
The attention ratio identity \eqref{eq:attn-odds-margin-bridge} gives $a_n(\lambda W^\star)\to e_c$ as $\lambda\to\infty$.
By continuity and finiteness of the training set, $\Loss_U(\lambda W^\star)\to\ell_U^\star$.
Together with $\Loss_U(W)\ge\ell_U^\star$, this proves $\inf_W\Loss_U(W)=\ell_U^\star$.
\end{proof}

\begin{lemma}\label{lem:fixed-prediction-noncollapse}
Suppose that $U$ satisfies \eqref{eq:fixed-prediction-condition}, and let $L_0:=\Loss_U(0)$.
Then
\begin{equation}\label{eq:fixed-initial-loss-bound}
  L_0\le \log K+\sqrt{2}r_U.
\end{equation}
For every $W\in\R^{d\times d}$,
\begin{equation*}
  \Loss_U(W)\le L_0\quad\Longrightarrow\quad a_{n,c}(W)\ge\delta_0,\qquad n\in[N].
\end{equation*}
On this sublevel set, with $\nu_U:=\kappa_-\delta_0/(NL_{\max})>0$,
\begin{equation}\label{eq:fixed-loss-margin}
  E_U(W)\ge\nu_Ue^{-\rho_2(W)}.
\end{equation}
\end{lemma}

\begin{proof}
Let $f_{\delta,n}$ be the sample loss with prediction matrix $\delta B$.
Adding the same constant to all logits leaves the cross entropy loss unchanged, so $f_{U,n}(a)$ can be computed from $CU m(a)=\delta Bm(a)+Rm(a)$.

For $z\in\R^K$, the cross entropy gradient $\softmax(z)-e_{y_n}$ has Euclidean norm at most $\sqrt{2}$.
Thus cross entropy is $\sqrt{2}$-Lipschitz in its logits.
Since $\Vert m(a)\Vert_2\le 1$, we obtain
\begin{equation*}
  |f_{U,n}(a)-f_{\delta,n}(a)|\le \sqrt{2}\Vert Rm(a)\Vert_2\le \sqrt{2}r_U.
\end{equation*}
At $W=0$, we have $a_{n,x}(0)=c_{n,x}/L_n$.
Assumption \ref{ass:weak-count} therefore makes the correct output logit under $\delta B$ maximal, giving $f_{\delta,n}(a_n(0))\le \log K$.
Averaging the loss perturbation bound proves \eqref{eq:fixed-initial-loss-bound}.

Now fix $W$ with $\Loss_U(W)\le L_0$ and $n\in[N]$, and let $u:=a_{n,c}(W)$.
The Jensen estimate in the proof of Lemma \ref{lem:noncollapse-entry}, applied with logit gap $\delta$, gives
\begin{equation*}
  f_{\delta,n}(a_n(W))\ge\frac{\delta(1-Ku)}{K-1}.
\end{equation*}
Combining this with the loss perturbation bound and nonnegativity of the sample losses yields
\begin{equation*}
  \frac{\delta(1-Ku)}{K-1}-\sqrt{2}r_U\le f_{U,n}(a_n(W))\le N\Loss_U(W)\le N(\log K+\sqrt{2}r_U).
\end{equation*}
Rearranging gives
\begin{equation*}
  u\ge\frac{1}{K}\left(1-\frac{(K-1)[N\log K+\sqrt{2}(N+1)r_U]}{\delta}\right)\ge\frac{1}{2K}=\delta_0,
\end{equation*}
where the last inequality follows from \eqref{eq:fixed-prediction-condition}.

Finally, choose $(n,x)\in\I_2$ attaining $\rho_2(W)$.
Since $c_{n,x}\ge 1$ and $c_{n,c}\le L_{\max}$, the attention ratio identity \eqref{eq:attn-odds-margin-bridge} gives
\begin{equation*}
  a_{n,x}(W)=a_{n,c}(W)\frac{c_{n,x}}{c_{n,c}}e^{-\rho_2(W)}\ge\frac{\delta_0}{L_{\max}}e^{-\rho_2(W)}.
\end{equation*}
Using \eqref{eq:fixed-excess-attention} and $\bar s(W)\ge a_{n,x}(W)/N$, we obtain
\begin{equation*}
  E_U(W)\ge\kappa_-\bar s(W)\ge\frac{\kappa_-}{N}a_{n,x}(W)\ge\nu_Ue^{-\rho_2(W)}.
\end{equation*}
This proves \eqref{eq:fixed-loss-margin}.
\end{proof}

\paragraph{Gradient and momentum estimates.}
We first apply Lemma \ref{lem:hierarchical-gradient} to $F_{U,n}$.
Let $C_U:=\frac{2\sqrt{2}NQ_U}{\kappa_-^2}$.

\begin{lemma}\label{lem:fixed-prediction-gradient}
Suppose that $U$ satisfies \eqref{eq:fixed-prediction-condition} and $\mathsf N$ satisfies \eqref{eq:common-norm-interface}.
For every $W\in\R^{d\times d}$,
\begin{align}
  & \mathsf N_*(\nabla_W\Loss_U(W))\ge\gamma_{2,\mathsf N}E_U(W)-C_UE_U(W)^2,\label{eq:fixed-global-margin-gradient}\\
  & \mathsf N_*(\nabla_W\Loss_U(W))\le 2\sqrt{2}Q_U\bar s(W).\label{eq:fixed-gradient-upper}
\end{align}
If $\Loss_U(W)\le L_0$, then
\begin{equation}\label{eq:fixed-gradient-excess}
  \mathsf N_*(\nabla_W\Loss_U(W))\ge E_U(W)\max\{c_{\mathsf N},\gamma_{2,\mathsf N}-C_UE_U(W)\},
\end{equation}
where $c_{\mathsf N}$ is the constant in Lemma \ref{lem:hierarchical-gradient}.
\end{lemma}

\begin{proof}
Fix $n\in[N]$ and $a\in\mathcal A_n$.
The term $f_{U,n}(e_c)$ is constant in $a$, and $p_n(a)-e_{y_n}$ is orthogonal to $\mathbf1$.
The cross entropy chain rule therefore gives
\begin{equation*}
  \partial_xF_{U,n}(a)=(p_n(a)-e_{y_n})^\top Ux=(p_n(a)-e_{y_n})^\top CUx.
\end{equation*}
Using the definition in Lemma \ref{lem:grouped-loss-chain-rule}, we obtain
\begin{equation*}
  d_{n,x}^U(a):=\partial_xF_{U,n}(a)-\sum_{z\in\V}a_z\partial_zF_{U,n}(a)=(p_n(a)-e_{y_n})^\top CU(x-m(a)).
\end{equation*}
Applying the matrix gradient identity in that lemma and averaging over the samples yield
\begin{equation}\label{eq:fixed-gradient-representation}
  -\nabla_W\Loss_U(W)=\frac{1}{N}\sum_{n=1}^N\sum_{x\in\X_n^-}a_{n,x}(W)d_{n,x}^U(a_n(W))A_{n,x}.
\end{equation}

Orthonormality of the vocabulary and $a\in\mathcal A_n$ give
\begin{equation*}
  \Vert x-m(a)\Vert_2^2=1-2a_x+\sum_{z\in\V}a_z^2\le 2.
\end{equation*}
Together with $\Vert p_n(a)-e_{y_n}\Vert_2\le \sqrt{2}$, this implies
\begin{equation*}
  |d_{n,x}^U(a)|\le \Vert p_n(a)-e_{y_n}\Vert_2Q_U\Vert x-m(a)\Vert_2\le 2Q_U.
\end{equation*}
Using $\mathsf N_*(A_{n,x})=\sqrt{2}$ and the dual norm triangle inequality, we obtain
\begin{equation*}
  \begin{aligned}
    \mathsf N_*(\nabla_W\Loss_U(W)) & \le \frac{\sqrt{2}}{N}\sum_{n=1}^N\sum_{x\in\X_n^-}a_{n,x}(W)|d_{n,x}^U(a_n(W))|\\
    & \le 2\sqrt{2}Q_U\bar s(W).
  \end{aligned}
\end{equation*}
This proves \eqref{eq:fixed-gradient-upper}.

For the lower bounds, each $F_{U,n}$ extends to a smooth convex function on $\R^{\V}$ and satisfies $F_{U,n}(e_c)=0$.
By Lemma \ref{lem:fixed-prediction-loss},
\begin{equation*}
  0<D_{n,x}^U(a)\le 2Q_U,\qquad F_{U,n}(a)\ge\kappa_-s_n(a).
\end{equation*}
Thus the hypotheses of Lemma \ref{lem:hierarchical-gradient}(ii) hold with $F_n=F_{U,n}$, $L=2Q_U$, and $\kappa=\kappa_-$.
Under this choice, $E_F=E_U$ and $\nabla_WE_F=\nabla_W\Loss_U$, so that lemma gives
\begin{equation*}
  \mathsf N_*(\nabla_W\Loss_U(W))\ge\gamma_{2,\mathsf N}E_U(W)-\frac{2\sqrt{2}NQ_U}{\kappa_-^2}E_U(W)^2=\gamma_{2,\mathsf N}E_U(W)-C_UE_U(W)^2.
\end{equation*}
This proves \eqref{eq:fixed-global-margin-gradient}.

If $\Loss_U(W)\le L_0$, Lemma \ref{lem:fixed-prediction-noncollapse} gives $a_{n,c}(W)\ge\delta_0$ for all $n\in[N]$.
Part (i) of Lemma \ref{lem:hierarchical-gradient} then supplies the lower bound $c_{\mathsf N}E_U(W)$.
Combining the two lower bounds and using $E_U(W)\ge 0$, we obtain
\begin{equation*}
  \begin{aligned}
    \mathsf N_*(\nabla_W\Loss_U(W)) & \ge \max\{c_{\mathsf N}E_U(W),\gamma_{2,\mathsf N}E_U(W)-C_UE_U(W)^2\}\\
    & = E_U(W)\max\{c_{\mathsf N},\gamma_{2,\mathsf N}-C_UE_U(W)\}.
  \end{aligned}
\end{equation*}
This proves \eqref{eq:fixed-gradient-excess}.
\end{proof}

We next bound the change in the loss gradient by a multiple of $\bar s(W)$.
Together with \eqref{eq:fixed-excess-attention}, this estimate will give bounds on the momentum error and the Taylor remainder in terms of the excess loss.

\begin{lemma}\label{lem:fixed-prediction-stability}
Let $U$ be fixed, and let $\mathsf N$ satisfy \eqref{eq:common-norm-interface}.
There is a constant $L_{1,U}>0$, depending only on $Q_U$, such that, for all $W,W'\in\R^{d\times d}$,
\begin{equation*}
  \mathsf N_*(\nabla_W\Loss_U(W')-\nabla_W\Loss_U(W))\le L_{1,U}(e^{2\mathsf N(W'-W)}-1)\bar s(W).
\end{equation*}
\end{lemma}

\begin{proof}
Fix $n\in[N]$, and let $a:=a_n(W)$, $a':=a_n(W')$, and $u:=e^{2\mathsf N(W'-W)}-1$.
Since $\Vert W'-W\Vert_{\op}\le \mathsf N(W'-W)$, Lemma \ref{lem:attn-stab} gives
\begin{equation*}
  |a'_x-a_x|\le ua_x,\qquad \Vert a'-a\Vert_1\le 2us_n(W).
\end{equation*}
Also,
\begin{equation*}
  \Vert m(a')-m(a)\Vert_2\le \sum_{x\in\V}|a'_x-a_x|=\Vert a'-a\Vert_1.
\end{equation*}

To bound the probability change, let $a_\tau:=(1-\tau)a+\tau a'$ for $\tau\in[0,1]$.
The softmax Jacobian along this segment is
\begin{equation*}
  J_\tau:=\operatorname{diag}(p_n(a_\tau))-p_n(a_\tau)p_n(a_\tau)^\top.
\end{equation*}
It satisfies $J_\tau\mathbf1=0$ and $0\preceq J_\tau\preceq I_K$, so $\Vert J_\tau\Vert_{\op}\le 1$.
Integration along the segment therefore gives
\begin{equation*}
  \begin{aligned}
    \Vert p_n(a')-p_n(a)\Vert_2 & = \left\Vert\int_0^1J_\tau CU(m(a')-m(a))\,d\tau\right\Vert_2\\
    & \le Q_U\Vert m(a')-m(a)\Vert_2\\
    & \le Q_U\Vert a'-a\Vert_1.
  \end{aligned}
\end{equation*}

For this sample, let $d_x:=d_{n,x}^U(a)$ and $d'_x:=d_{n,x}^U(a')$.
The formula for $d_{n,x}^U$ gives
\begin{equation*}
    d'_x-d_x = (p_n(a')-p_n(a))^\top CU(x-m(a'))-(p_n(a)-e_{y_n})^\top CU(m(a')-m(a)).
\end{equation*}
Using $\Vert x-m(a')\Vert_2\le 2$ and $\Vert p_n(a)-e_{y_n}\Vert_2\le \sqrt{2}$, we obtain
\begin{equation*}
  \begin{aligned}
    |d'_x-d_x| & \le 2Q_U\Vert p_n(a')-p_n(a)\Vert_2+\sqrt{2}Q_U\Vert m(a')-m(a)\Vert_2\\
    & \le (2Q_U^2+\sqrt{2}Q_U)\Vert a'-a\Vert_1.
  \end{aligned}
\end{equation*}

We now use the identity
\begin{equation*}
  a'_xd'_x-a_xd_x=(a'_x-a_x)d'_x+a_x(d'_x-d_x).
\end{equation*}
The bound $|d'_x|\le 2Q_U$ and the attention estimates above yield
\begin{equation*}
  \begin{aligned}
    \sum_{x\in\X_n^-}|a'_xd'_x-a_xd_x| & \le 2Q_Uus_n(W)+(2Q_U^2+\sqrt{2}Q_U)s_n(W)\Vert a'-a\Vert_1\\
    & \le 2Q_Uus_n(W)+2(2Q_U^2+\sqrt{2}Q_U)us_n(W)^2\\
    & \le 2Q_U(1+2Q_U+\sqrt{2})us_n(W),
  \end{aligned}
\end{equation*}
where the last inequality uses $s_n(W)\le 1$.

Let
\begin{equation*}
  L_{1,U}:=\max\{1,2\sqrt{2}Q_U(1+2Q_U+\sqrt{2})\}.
\end{equation*}
Applying the dual norm triangle inequality to \eqref{eq:fixed-gradient-representation} and using $\mathsf N_*(A_{n,x})=\sqrt{2}$, we obtain
\begin{equation*}
  \mathsf N_*(\nabla_W\Loss_U(W')-\nabla_W\Loss_U(W))\le \frac{L_{1,U}u}{N}\sum_{n=1}^Ns_n(W)=L_{1,U}u\bar s(W).
\end{equation*}
Substituting the definition of $u$ proves the result.
\end{proof}

\paragraph{Loss decrease and margin convergence.}
For $\mathsf N=\Vert\cdot\Vert_{S_p}$, let $\gamma_{2,p}:=\gamma_{2,\mathsf N}$ and $c_p:=c_{\mathsf N}$.
Equation \eqref{eq:common-margin-witness} gives $\gamma_{2,p}>0$.

For $\alpha>0$, let
\begin{equation*}
  C_{D,U,\alpha}:=\frac{L_{1,U}[2C(\alpha,2,\beta_2)+e^2]}{\kappa_-},\qquad \bar\eta_{U,p,\alpha}:=\min\left\{1,\frac{c_p}{2C_{D,U,\alpha}}\right\}.
\end{equation*}

The next lemma proves loss decrease under this step size bound and gives a recurrence for the excess loss.
We will use this recurrence to derive the loss and margin estimates below.

\begin{lemma}\label{lem:schatten-fixed-prediction-descent}
Suppose that $U$ satisfies \eqref{eq:fixed-prediction-condition}.
Consider Phase II of Algorithm \ref{alg:schatten-two-phase} with any exact maximizing directions and
\begin{equation*}
  \eta_{2,t}=\eta_{2,0}(t+1)^{-\alpha},\qquad \alpha>0,\qquad 0<\eta_{2,0}\le \bar\eta_{U,p,\alpha}.
\end{equation*}
Let $E_t:=E_U(W_{2,t})$, $S_t:=\sum_{s=0}^{t-1}\eta_{2,s}$, and $Q_t:=\sum_{s=0}^{t-1}\eta_{2,s}^2$.
Then, for every $t\ge 0$,
\begin{align}
  & \Loss_U(W_{2,t})\le L_0,\qquad a_{n,c}(W_{2,t})\ge\delta_0\quad(n\in[N]),\notag\\
  & 0<E_{t+1}\le \left[1-\eta_{2,t}\max\{c_p,\gamma_{2,p}-C_UE_t\}+C_{D,U,\alpha}\eta_{2,t}^2\right]E_t.\label{eq:schatten-fixed-unified-descent}
\end{align}
\end{lemma}

\begin{proof}
For this proof, let $W_t:=W_{2,t}$, $\eta_t:=\eta_{2,t}$, and $G_t:=\nabla_W\Loss_U(W_t)$.
Let
\begin{equation*}
  \widehat M_t:=\frac{M_{2,t}}{1-\beta_2^{t+1}},\qquad \Omega_t:=\widehat M_t-G_t.
\end{equation*}
Since $\Vert D_{2,s}\Vert_{S_p}\le 1$, the updates give
\begin{equation*}
  \Vert W_t-W_{t-r}\Vert_{S_p}\le \sum_{j=1}^r\eta_{t-j},\qquad 0\le r\le t.
\end{equation*}
For $0<\beta_2<1$, let
\begin{equation*}
  w_{t,r}:=\frac{(1-\beta_2)\beta_2^r}{1-\beta_2^{t+1}},\qquad 0\le r\le t.
\end{equation*}
Expanding the momentum recursion gives $\Omega_t=\sum_{r=0}^tw_{t,r}(G_{t-r}-G_t)$.
Lemmas \ref{lem:fixed-prediction-stability} and \ref{lem:ema-phase2} then yield
\begin{equation*}
  \begin{aligned}
    \Vert\Omega_t\Vert_{S_{p'}} & \le L_{1,U}\bar s(W_t)\sum_{r=0}^tw_{t,r}\left[\exp\left(2\sum_{j=1}^r\eta_{t-j}\right)-1\right]\\
    & \le L_{1,U}C(\alpha,2,\beta_2)\eta_t\bar s(W_t).
  \end{aligned}
\end{equation*}
When $\beta_2=0$, we have $\Omega_t=0$, so the same bound holds.

The direction $D_{2,t}$ maximizes the inner product with $\widehat M_t$ over the Schatten $S_p$ unit ball.
As in the proof of Lemma \ref{lem:first-order}, norm duality gives
\begin{equation*}
  \ip{G_t}{D_{2,t}}=\Vert\widehat M_t\Vert_{S_{p'}}-\ip{\Omega_t}{D_{2,t}}\ge\Vert G_t\Vert_{S_{p'}}-2\Vert\Omega_t\Vert_{S_{p'}}.
\end{equation*}
Since $\eta_t\le 1$, integration along the update segment and the momentum error bound imply
\begin{equation*}
  \begin{aligned}
    E_{t+1}-E_t & \le -\eta_t\ip{G_t}{D_{2,t}}+L_{1,U}\bar s(W_t)\int_0^{\eta_t}(e^{2\tau}-1)\,d\tau\\
    & \le -\eta_t\ip{G_t}{D_{2,t}}+e^2L_{1,U}\eta_t^2\bar s(W_t)\\
    & \le -\eta_t\Vert G_t\Vert_{S_{p'}}+L_{1,U}[2C(\alpha,2,\beta_2)+e^2]\eta_t^2\bar s(W_t)\\
    & = -\eta_t\Vert G_t\Vert_{S_{p'}}+\kappa_-C_{D,U,\alpha}\eta_t^2\bar s(W_t).
  \end{aligned}
\end{equation*}

We now prove $\Loss_U(W_t)\le L_0$ by induction.
The claim holds at $t=0$ because $W_0=0$.
Suppose that $\Loss_U(W_t)\le L_0$.
Lemma \ref{lem:fixed-prediction-noncollapse} gives $a_{n,c}(W_t)\ge\delta_0$ for all $n\in[N]$, so \eqref{eq:fixed-gradient-excess} applies.
Using this gradient lower bound and $\kappa_-\bar s(W_t)\le E_t$, we obtain
\begin{equation*}
  E_{t+1}-E_t\le -\eta_tE_t\max\{c_p,\gamma_{2,p}-C_UE_t\}+C_{D,U,\alpha}\eta_t^2E_t.
\end{equation*}
This is the upper bound in \eqref{eq:schatten-fixed-unified-descent}.
The step size condition gives
\begin{equation*}
  \max\{c_p,\gamma_{2,p}-C_UE_t\}-C_{D,U,\alpha}\eta_t\ge c_p-C_{D,U,\alpha}\bar\eta_{U,p,\alpha}\ge\frac{c_p}{2}.
\end{equation*}
Consequently,
\begin{equation*}
  E_t-E_{t+1}\ge\frac{c_p}{2}\eta_tE_t\ge 0.
\end{equation*}
Since $E_t=\Loss_U(W_t)-\ell_U^\star$, this implies $\Loss_U(W_{t+1})\le\Loss_U(W_t)\le L_0$ and closes the induction.
The attention lower bound therefore holds at every iteration.
Every finite iterate is a finite matrix, so \eqref{eq:fixed-excess-attention} also gives $E_t>0$.

Finally, summing the loss decrease yields
\begin{equation*}
  \frac{c_p}{2}\sum_{s=0}^{t-1}\eta_{2,s}E_s\le \sum_{s=0}^{t-1}(E_s-E_{s+1})=E_0-E_t\le E_0.
\end{equation*}
\end{proof}

The next lemma uses the bound on $\sum_{s=0}^{t-1}\eta_{2,s}E_s$ to obtain a loss estimate with leading term $-\gamma_{2,p}S_t$.
Equation \eqref{eq:fixed-loss-margin} and an upper bound on the parameter norm then give the margin estimates.

\begin{lemma}\label{lem:schatten-fixed-rates}
Under the hypotheses of Lemma \ref{lem:schatten-fixed-prediction-descent}, there are $C_{L,U},C_{\rho,U}>0$, independent of $\eta_{2,0}$ and the Phase II maximizing directions, such that, for every integer $t\ge 0$,
\begin{equation*}
  \begin{aligned}
    & \log E_t\le -\gamma_{2,p}S_t+C_{L,U}Q_t+C_{L,U},\\
    & \rho_2(W_{2,t})\ge\gamma_{2,p}S_t-C_{L,U}Q_t-C_{\rho,U}.
  \end{aligned}
\end{equation*}
For every integer $t\ge 1$ satisfying $\gamma_{2,p}S_t-C_{L,U}Q_t-C_{\rho,U}>0$, we have $W_{2,t}\ne0$ and
\begin{equation}\label{eq:schatten-fixed-normalized-margin}
  0\le \gamma_{2,p}-\rho_2\left(\frac{W_{2,t}}{\Vert W_{2,t}\Vert_{S_p}}\right)\le \frac{C_{L,U}Q_t+C_{\rho,U}}{S_t}.
\end{equation}
\end{lemma}

\begin{proof}
Since $E_t,E_{t+1}>0$, the factor multiplying $E_t$ in \eqref{eq:schatten-fixed-unified-descent} is positive.
Taking logarithms and using $\log(1+z)\le z$ give
\begin{equation*}
  \begin{aligned}
    \log E_{t+1}-\log E_t & \le -\eta_{2,t}\max\{c_p,\gamma_{2,p}-C_UE_t\}+C_{D,U,\alpha}\eta_{2,t}^2\\
    & \le -\gamma_{2,p}\eta_{2,t}+C_U\eta_{2,t}E_t+C_{D,U,\alpha}\eta_{2,t}^2.
  \end{aligned}
\end{equation*}
Summing over $s=0,\ldots,t-1$, we obtain
\begin{equation*}
  \begin{aligned}
    \log E_t & \le \log E_0-\gamma_{2,p}S_t+C_U\sum_{s=0}^{t-1}\eta_{2,s}E_s+C_{D,U,\alpha}Q_t\\
    & \le -\gamma_{2,p}S_t+C_{D,U,\alpha}Q_t+\log E_0+\frac{2C_UE_0}{c_p}.
  \end{aligned}
\end{equation*}
The last inequality uses $\sum_{s=0}^{t-1}\eta_{2,s}E_s\le 2E_0/c_p$ from Lemma \ref{lem:schatten-fixed-prediction-descent}.
Let
\begin{equation*}
  C_{L,U}:=\max\left\{1,C_{D,U,\alpha},|\log E_0|+\frac{2C_UE_0}{c_p}\right\}.
\end{equation*}
This choice proves the loss estimate.

Let $C_{\rho,U}:=C_{L,U}+|\log\nu_U|$.
By \eqref{eq:fixed-loss-margin},
\begin{equation*}
  \begin{aligned}
    \rho_2(W_{2,t}) & \ge \log\nu_U-\log E_t\\
    & \ge \gamma_{2,p}S_t-C_{L,U}Q_t-C_{L,U}+\log\nu_U\\
    & \ge \gamma_{2,p}S_t-C_{L,U}Q_t-C_{\rho,U}.
  \end{aligned}
\end{equation*}
Since $E_0=E_U(0)$, the quantities defining $C_{L,U}$ and $C_{\rho,U}$ are independent of $\eta_{2,0}$ and the Phase II maximizing directions.

Summing the updates from $W_{2,0}=0$ gives
\begin{equation*}
  \Vert W_{2,t}\Vert_{S_p}\le \sum_{s=0}^{t-1}\eta_{2,s}\Vert D_{2,s}\Vert_{S_p}\le S_t.
\end{equation*}
If $\gamma_{2,p}S_t-C_{L,U}Q_t-C_{\rho,U}>0$, then $\rho_2(W_{2,t})>0$ and hence $W_{2,t}\ne0$.
Positive homogeneity and the norm upper bound therefore yield
\begin{equation*}
    \rho_2\left(\frac{W_{2,t}}{\Vert W_{2,t}\Vert_{S_p}}\right) = \frac{\rho_2(W_{2,t})}{\Vert W_{2,t}\Vert_{S_p}}
    \ge \frac{\rho_2(W_{2,t})}{S_t}
    \ge \gamma_{2,p}-\frac{C_{L,U}Q_t+C_{\rho,U}}{S_t}.
\end{equation*}
On the other hand, the maximum margin identity \eqref{eq:common-margin-witness} gives
\begin{equation*}
  \rho_2\left(\frac{W_{2,t}}{\Vert W_{2,t}\Vert_{S_p}}\right)\le \max_{\Vert V\Vert_{S_p}\le 1}\rho_2(V)=\gamma_{2,p}.
\end{equation*}
Combining these bounds proves \eqref{eq:schatten-fixed-normalized-margin}.
\end{proof}

We now combine the preprocessing and main phase estimates.

\begin{theorem}\label{thm:schatten-unregularized-phase2}
Fix $1\le p<\infty$ and allow any exact maximizing directions in both phases of Algorithm \ref{alg:schatten-two-phase}.
Suppose that $K\ge 2$, $h$ is bijective, the vocabulary embeddings are orthonormal, Assumptions \ref{ass:token-label}, \ref{ass:weak-count}, and \ref{ass:separability} hold, and $\I_2\ne\varnothing$.
Let $\beta_1,\beta_2\in[0,1)$ and use
\begin{equation*}
  \eta_{1,s}=\eta_{1,0}(s+1)^{-1/2},\qquad \eta_{2,t}=\eta_{2,0}(t+1)^{-1/2},
\end{equation*}
with $\eta_{1,0}>0$.
There is an integer $T_p\ge 1$, independent of the Phase I maximizing directions and the ambient dimension, such that every $T\ge T_p$ produces $U:=W_{1,T}$ satisfying \eqref{eq:fixed-prediction-condition}.
If $0<\eta_{2,0}\le \bar\eta_{U,p,1/2}$, then there are constants $C_{1,U},C_{2,U},C_{3,U},C_{4,U}>0$, independent of the Phase II maximizing directions, such that the following bounds hold.

\textbf{\textup{(i) Training loss.}}
For every integer $t\ge 1$,
\begin{equation}\label{eq:schatten-phase2-loss-rate}
  0<E_U(W_{2,t})\le C_{1,U}t^{C_{2,U}}e^{-C_{3,U}\sqrt{t}}.
\end{equation}
In particular, $\Loss_U(W_{2,t})\to\ell_U^\star$.

\textbf{\textup{(ii) Normalized Schatten margin.}}
There is an integer $t_U\ge 1$ such that, for every $t\ge t_U$, we have $W_{2,t}\ne0$ and
\begin{equation}\label{eq:schatten-phase2-margin-rate}
  0\le \gamma_{2,p}-\rho_2\left(\frac{W_{2,t}}{\Vert W_{2,t}\Vert_{S_p}}\right)\le C_{4,U}\frac{1+\log t}{\sqrt{t}}.
\end{equation}
\end{theorem}

\begin{proof}
Lemma \ref{lem:schatten-phase1-asymptotics} gives $\delta_T/S_{1,T}\to\gamma_{1,p}>0$.
Since $S_{1,T}\ge\eta_{1,0}\sqrt{T}$, for all sufficiently large $T$ we have
\begin{equation*}
  \delta_T\ge\frac{\gamma_{1,p}}{2}S_{1,T}\ge\frac{\gamma_{1,p}\eta_{1,0}}{2}\sqrt{T}\longrightarrow\infty.
\end{equation*}
The same lemma also gives
\begin{equation*}
  \frac{\Vert R_T\Vert_{\op}}{\delta_T}\le \frac{\Vert R_T\Vert_{S_p}}{\delta_T}\longrightarrow0.
\end{equation*}
Consequently,
\begin{equation*}
  \frac{2(K-1)[N\log K+\sqrt{2}(N+1)\Vert R_T\Vert_{\op}]}{\delta_T}\longrightarrow0.
\end{equation*}
We can therefore choose an integer $T_p\ge 1$ such that $\delta_T>0$ and this ratio is at most one for every $T\ge T_p$.
This is precisely \eqref{eq:fixed-prediction-condition}.
The preprocessing bounds are uniform over the Phase I maximizing directions and independent of the ambient dimension, so $T_p$ has the stated independence.

Fix such a $T$ and let $U:=W_{1,T}$.
Lemmas \ref{lem:schatten-fixed-prediction-descent} and \ref{lem:schatten-fixed-rates} apply with $\alpha=1/2$.
For $t\ge 1$, integral comparison gives
\begin{equation*}
  S_t\ge 2\eta_{2,0}(\sqrt{t+1}-1),\qquad Q_t\le \eta_{2,0}^2(1+\log t).
\end{equation*}
Substituting into the logarithmic loss estimate yields
\begin{equation*}
  \begin{aligned}
    \log E_t & \le -\gamma_{2,p}S_t+C_{L,U}Q_t+C_{L,U}\\
    & \le -2\gamma_{2,p}\eta_{2,0}(\sqrt{t+1}-1)+C_{L,U}\eta_{2,0}^2(1+\log t)+C_{L,U}\\
    & \le -2\gamma_{2,p}\eta_{2,0}\sqrt{t}+C_{L,U}\eta_{2,0}^2\log t+C_{L,U}(1+\eta_{2,0}^2)+2\gamma_{2,p}\eta_{2,0}.
  \end{aligned}
\end{equation*}
Let
\begin{equation*}
    C_{1,U}:=\exp\left(C_{L,U}(1+\eta_{2,0}^2)+2\gamma_{2,p}\eta_{2,0}\right),\qquad
    C_{2,U}:=C_{L,U}\eta_{2,0}^2,\qquad C_{3,U}:=2\gamma_{2,p}\eta_{2,0}.
\end{equation*}
Exponentiating the preceding inequality gives the upper bound in \eqref{eq:schatten-phase2-loss-rate}.
Strict positivity follows from Lemma \ref{lem:schatten-fixed-prediction-descent}.

Since $S_t\to\infty$ and $Q_t/S_t\to0$, choose an integer $t_U\ge 1$ such that
\begin{equation*}
  \gamma_{2,p}S_t-C_{L,U}Q_t-C_{\rho,U}>0,\qquad t\ge t_U.
\end{equation*}
Lemma \ref{lem:schatten-fixed-rates} then gives $W_{2,t}\ne0$ and \eqref{eq:schatten-fixed-normalized-margin} for every $t\ge t_U$.
We also have $S_t\ge\eta_{2,0}\sqrt{t}$, since $k^{-1/2}\ge t^{-1/2}$ for $1\le k\le t$.
Let
\begin{equation*}
  C_{4,U}:=\frac{C_{L,U}\eta_{2,0}^2+C_{\rho,U}}{\eta_{2,0}}.
\end{equation*}
For $t\ge t_U$, we obtain
\begin{equation*}
  \begin{aligned}
    \gamma_{2,p}-\rho_2\left(\frac{W_{2,t}}{\Vert W_{2,t}\Vert_{S_p}}\right) & \le \frac{C_{L,U}Q_t+C_{\rho,U}}{S_t}\\
    & \le \frac{C_{L,U}\eta_{2,0}^2(1+\log t)+C_{\rho,U}}{\eta_{2,0}\sqrt{t}}\\
    & \le C_{4,U}\frac{1+\log t}{\sqrt{t}}.
  \end{aligned}
\end{equation*}
This proves \eqref{eq:schatten-phase2-margin-rate}.
The constructions give the stated dependence of the constants.

Finally, $C_{3,U}>0$ implies $E_U(W_{2,t})\to0$, and hence $\Loss_U(W_{2,t})\to\ell_U^\star$.
Since $(1+\log t)/\sqrt{t}\to0$, the normalized margin also converges to $\gamma_{2,p}$.
\end{proof}

We next give the normalized margin rates for general power law step sizes in Phase II.

\begin{corollary}\label{cor:schatten-general-power-law}
Under the data and preprocessing conditions of Theorem \ref{thm:schatten-unregularized-phase2}, fix $T\ge T_p$ and let $U:=W_{1,T}$.
For Phase II, use
\begin{equation*}
  \eta_{2,t}=\eta_{2,0}(t+1)^{-\alpha},\qquad 0<\alpha\le 1,\qquad 0<\eta_{2,0}\le \bar\eta_{U,p,\alpha}.
\end{equation*}
For all sufficiently large $t$, we have $W_{2,t}\ne0$, and the normalized margin satisfies the four rates in \eqref{eq:common-power-law-margin-rates}, with $W_t=W_{2,t}$, $\mathsf N=\Vert\cdot\Vert_{S_p}$, and $\gamma_{2,\mathsf N}=\gamma_{2,p}$.
\end{corollary}

\begin{proof}
Lemmas \ref{lem:schatten-fixed-prediction-descent} and \ref{lem:schatten-fixed-rates} apply with the chosen exponent $\alpha$.
Since $S_t\to\infty$ and $Q_t/S_t\to0$, the margin lower bound is eventually positive.
Applying the scalar sum estimates used in the proof of Corollary \ref{cor:common-power-law-rates} to \eqref{eq:schatten-fixed-normalized-margin} gives the four stated rates.
\end{proof}

\section{Convergence of Adam}\label{subsec:adam}

We analyze Adam with a fixed constant $\varepsilon>0$.
Both phases use this constant outside the square root in the denominator.
For phase $j\in\{1,2\}$, the parameters $\beta_{j,1},\beta_{j,2}\in[0,1)$ control the first and second moments, respectively.
All squares, divisions, and square roots in the updates are computed separately for each matrix entry.
Phase II holds the prediction matrix $W_{1,T}$ fixed.

\begin{algorithm}[htbp]
\caption{Two Phase Adam Training}
\label{alg:adam-two-phase}
\begin{algorithmic}[1]
\STATE \textbf{\textsc{Phase I}: Preprocessing}
\STATE \textbf{Input} $\D_0$; $T$; $\{\eta_{1,s}\}_{s=0}^{T-1}$; $\beta_{1,1},\beta_{1,2}\in[0,1)$; $\varepsilon>0$.
\STATE \textbf{Initialization} $W_{1,0}=m_{1,-1}=v_{1,-1}=0$.
\FOR{$s=0,\ldots,T-1$}
  \STATE $g_{1,s}=\nabla\Loss_0(W_{1,s})$.
  \STATE $m_{1,s}=\beta_{1,1}m_{1,s-1}+(1-\beta_{1,1})g_{1,s}$.
  \STATE $v_{1,s}=\beta_{1,2}v_{1,s-1}+(1-\beta_{1,2})g_{1,s}^{\odot 2}$.
  \STATE $\widehat m_{1,s}=m_{1,s}/(1-\beta_{1,1}^{s+1}),\qquad \widehat v_{1,s}=v_{1,s}/(1-\beta_{1,2}^{s+1})$.
  \STATE $W_{1,s+1}=W_{1,s}-\eta_{1,s}\widehat m_{1,s}/(\sqrt{\widehat v_{1,s}}+\varepsilon)$.
\ENDFOR
\par\vspace{0.4ex}\hrule\vspace{0.4ex}
\STATE \textbf{\textsc{Phase II}: Main Training}
\STATE \textbf{Input} $\D$; $W_{1,T}$; $t$; $\{\eta_{2,r}\}_{r=0}^{t-1}$; $\beta_{2,1},\beta_{2,2}\in[0,1)$.
\STATE \textbf{Loss} $\Loss_T(W):=\Loss(W_{1,T},W)$.
\STATE \textbf{Initialization} $W_{2,0}=m_{2,-1}=v_{2,-1}=0$.
\FOR{$r=0,\ldots,t-1$}
  \STATE $g_{2,r}=\nabla\Loss_T(W_{2,r})$.
  \STATE $m_{2,r}=\beta_{2,1}m_{2,r-1}+(1-\beta_{2,1})g_{2,r}$.
  \STATE $v_{2,r}=\beta_{2,2}v_{2,r-1}+(1-\beta_{2,2})g_{2,r}^{\odot 2}$.
  \STATE $\widehat m_{2,r}=m_{2,r}/(1-\beta_{2,1}^{r+1}),\qquad \widehat v_{2,r}=v_{2,r}/(1-\beta_{2,2}^{r+1})$.
  \STATE $W_{2,r+1}=W_{2,r}-\eta_{2,r}\widehat m_{2,r}/(\sqrt{\widehat v_{2,r}}+\varepsilon)$.
\ENDFOR
\STATE \textbf{Output} $(W_{1,T},W_{2,t})$.
\end{algorithmic}
\end{algorithm}

\subsection*{\textbf{Preprocessing phase}}

Let $R_{1,F}^\star$ be the optimal value of
\begin{equation}\label{eq:adam-phase1-frobenius-hard-margin}
  \begin{aligned}
    \min_{W\in\R^{K\times d}}\quad & \Vert W\Vert_F\\
    \text{subject to}\quad & (e_{h(x)}-e_v)^\top Wx\ge 1,\\
    & x\in\V,\quad v\in[K]\setminus\{h(x)\}.
  \end{aligned}
\end{equation}
Use $\eta_{1,s}=\eta_{1,0}(s+1)^{-\alpha_1}$ with $0<\alpha_1\le 1$, and let $L_s:=\Loss_0(W_{1,s})$ and $S_{1,s}:=\sum_{r=0}^{s-1}\eta_{1,r}$.
For a matrix $H$, let $\Vert H\Vert_{\max}:=\max_{u,v}|H_{uv}|$.
The next lemma compares the two moments with the current gradient.
These estimates give loss decrease for a sufficiently small initial step size.

\begin{lemma}\label{lem:adam-phase1-descent}
Suppose that $K\ge 2$, $h$ is bijective, the vocabulary embeddings are orthonormal, and the $K$ single token preprocessing samples are equally weighted.
There are $C_{1,m},C_{1,v}\ge 0$ and $\bar\eta_{1,\mathrm A}\in(0,1]$, depending only on $K,\alpha_1,\beta_{1,1},\beta_{1,2},\varepsilon$, such that, whenever $0<\eta_{1,0}\le \bar\eta_{1,\mathrm A}$, the following bounds hold for every integer $s\ge 0$:
\begin{equation}\label{eq:adam-phase1-moment-tracking}
  \begin{aligned}
    & \Vert\widehat m_{1,s}-\nabla\Loss_0(W_{1,s})\Vert_F\le C_{1,m}\eta_{1,s}\Vert\nabla\Loss_0(W_{1,s})\Vert_F,\\
    & \Vert\sqrt{\widehat v_{1,s}}\Vert_{\max}\le C_{1,v}\Vert\nabla\Loss_0(W_{1,s})\Vert_F.
  \end{aligned}
\end{equation}
Moreover, there is $c_D>0$ such that
\begin{align}
  & L_s\le L_0=\log K,\label{eq:adam-phase1-sublevel}\\
  & L_{s+1}\le L_s-c_D\eta_{1,s}\Vert\nabla\Loss_0(W_{1,s})\Vert_F^2,\label{eq:adam-phase1-descent}\\
  & \frac{c_K}{\sqrt{K}}L_s\le \Vert\nabla\Loss_0(W_{1,s})\Vert_F\le \sqrt{2}L_s,\label{eq:adam-phase1-gradient-loss-bilateral}
\end{align}
where $c_K:=(1-e^{-K\log K})/(K\log K)$.
Let $c_L:=c_Dc_K^2/K$.
\end{lemma}

\begin{proof}
For this proof, let $G_s:=\nabla\Loss_0(W_{1,s})$ and $y_i:=h(x_i)$, and use $p_i(W)=\softmax(Wx_i)$.
Let
\begin{equation*}
  r_i(W):=p_i(W)-e_{y_i},\qquad q_i(W):=1-[p_i(W)]_{y_i}.
\end{equation*}
The probabilities of the other labels sum to $q_i(W)$, so
\begin{equation*}
  \Vert r_i(W)\Vert_2^2=q_i(W)^2+\sum_{v\ne y_i}[p_i(W)]_v^2\le 2q_i(W)^2\le 2.
\end{equation*}
Orthonormality of the vocabulary gives
\begin{equation*}
  \begin{aligned}
    & \nabla\Loss_0(W)=\frac{1}{K}\sum_{i=1}^Kr_i(W)x_i^\top,\\
    & \Vert\nabla\Loss_0(W)\Vert_F^2=\frac{1}{K^2}\sum_{i=1}^K\Vert r_i(W)\Vert_2^2\le \frac{2}{K}.
  \end{aligned}
\end{equation*}

Let $d_W:=\Vert W'-W\Vert_{\op}$.
Each output logit changes by at most $d_W$, so the softmax probability ratios lie in $[e^{-2d_W},e^{2d_W}]$.
For $v\ne y_i$, this gives
\begin{equation*}
  |[r_i(W')]_v-[r_i(W)]_v|\le (e^{2d_W}-1)[p_i(W)]_v.
\end{equation*}
For the correct label, we use $[r_i(W)]_{y_i}=-\sum_{v\ne y_i}[p_i(W)]_v$ to obtain
\begin{equation*}
  |[r_i(W')]_{y_i}-[r_i(W)]_{y_i}|\le \sum_{v\ne y_i}|[p_i(W')]_v-[p_i(W)]_v|\le (e^{2d_W}-1)q_i(W).
\end{equation*}
Squaring and summing these component bounds, followed by the gradient decomposition, yield
\begin{equation*}
  \Vert\nabla\Loss_0(W')-\nabla\Loss_0(W)\Vert_F^2=\frac{1}{K^2}\sum_{i=1}^K\Vert r_i(W')-r_i(W)\Vert_2^2\le (e^{2d_W}-1)^2\Vert\nabla\Loss_0(W)\Vert_F^2.
\end{equation*}
Taking square roots gives the relative gradient difference bound.
The triangle inequality also gives
\begin{equation*}
  \Vert\nabla\Loss_0(W')\Vert_F\le \Vert\nabla\Loss_0(W)\Vert_F+\Vert\nabla\Loss_0(W')-\nabla\Loss_0(W)\Vert_F\le e^{2d_W}\Vert\nabla\Loss_0(W)\Vert_F.
\end{equation*}

We first establish the moment estimates for $0<\eta_{1,0}\le 1$.
For $q\in\{1,2\}$, let
\begin{equation*}
  w_{s,r}^{(q)}:=\frac{(1-\beta_{1,q})\beta_{1,q}^r}{1-\beta_{1,q}^{s+1}},\qquad 0\le r\le s.
\end{equation*}
These weights are nonnegative and sum to one.
Expanding the moment recursions gives
\begin{equation*}
  \widehat m_{1,s}=\sum_{r=0}^sw_{s,r}^{(1)}G_{s-r},\qquad \widehat v_{1,s}=\sum_{r=0}^sw_{s,r}^{(2)}G_{s-r}^{\odot 2}.
\end{equation*}
When a moment parameter is zero, the corresponding sum contains only the term with $r=0$.
Let
\begin{equation*}
  D_1:=\frac{\sqrt{2/K}}{\varepsilon},\qquad \Pi_{s,r}:=\sum_{j=1}^r\eta_{1,s-j}.
\end{equation*}
The denominator lower bound and the global gradient bound imply
\begin{equation*}
  \Vert W_{1,s+1}-W_{1,s}\Vert_F\le \frac{\eta_{1,s}}{\varepsilon}\sum_{r=0}^sw_{s,r}^{(1)}\Vert G_{s-r}\Vert_F\le D_1\eta_{1,s}.
\end{equation*}
Summing the updates therefore gives
\begin{equation*}
  \Vert W_{1,s}-W_{1,s-r}\Vert_{\op}\le D_1\Pi_{s,r}.
\end{equation*}
The relative gradient bounds now yield
\begin{equation*}
  \begin{aligned}
    & \Vert\widehat m_{1,s}-G_s\Vert_F\le \Vert G_s\Vert_F\sum_{r=0}^sw_{s,r}^{(1)}(e^{2D_1\Pi_{s,r}}-1),\\
    & \Vert\sqrt{\widehat v_{1,s}}\Vert_{\max}^2\le \sum_{r=0}^sw_{s,r}^{(2)}\Vert G_{s-r}\Vert_F^2\le \Vert G_s\Vert_F^2\sum_{r=0}^sw_{s,r}^{(2)}e^{4D_1\Pi_{s,r}}.
  \end{aligned}
\end{equation*}
Let
\begin{equation*}
  C_{1,m}:=C(\alpha_1,2D_1,\beta_{1,1}),\qquad C_{1,v}:=\sqrt{1+C(\alpha_1,4D_1,\beta_{1,2})}.
\end{equation*}
Lemma \ref{lem:ema-phase2} gives the first moment bound in \eqref{eq:adam-phase1-moment-tracking}.
For the second moment, the same lemma and $\eta_{1,s}\le 1$ give
\begin{equation*}
    \sum_{r=0}^sw_{s,r}^{(2)}e^{4D_1\Pi_{s,r}} = 1+\sum_{r=0}^sw_{s,r}^{(2)}(e^{4D_1\Pi_{s,r}}-1)
    \le 1+C(\alpha_1,4D_1,\beta_{1,2})\eta_{1,s}
    \le C_{1,v}^2.
\end{equation*}
This proves \eqref{eq:adam-phase1-moment-tracking} with constants independent of $\eta_{1,0}\in(0,1]$.

We next prove loss decrease.
Let $P_s$ be the linear operator
\begin{equation*}
  P_sZ:=(\sqrt{\widehat v_{1,s}}+\varepsilon)^{-1}\odot Z,
\end{equation*}
and let $p_-:=(\varepsilon+\sqrt{2/K})^{-1}$.
Each entry of $\widehat v_{1,s}$ is an average of squared gradient entries and is therefore at most $2/K$.
Hence
\begin{equation*}
  p_-\mathrm{Id}\preceq P_s\preceq\varepsilon^{-1}\mathrm{Id}.
\end{equation*}

For any $H\in\R^{K\times d}$, the softmax Jacobian formula gives
\begin{equation*}
  \begin{aligned}
    0\le \nabla^2\Loss_0(W)[H,H] & = \frac{1}{K}\sum_{i=1}^K(Hx_i)^\top\left(\operatorname{diag}(p_i(W))-p_i(W)p_i(W)^\top\right)(Hx_i)\\
    & \le \frac{1}{K}\sum_{i=1}^K\Vert Hx_i\Vert_2^2\\
    & \le \frac{1}{K}\Vert H\Vert_F^2.
  \end{aligned}
\end{equation*}
Thus the Hessian is bounded above by $K^{-1}\mathrm{Id}$ in the Frobenius inner product.

Choose $\bar\eta_{1,\mathrm A}\in(0,1]$ sufficiently small that
\begin{equation*}
  \frac{C_{1,m}\bar\eta_{1,\mathrm A}}{\varepsilon}\le \frac{p_-}{2},\qquad \frac{(1+C_{1,m})^2\bar\eta_{1,\mathrm A}}{2K\varepsilon^2}\le \frac{p_-}{4}.
\end{equation*}
We also require
\begin{equation}\label{eq:adam-phase1-explicit-step-threshold}
  \bar\eta_{1,\mathrm A}\le K\varepsilon,
\end{equation}
which will be used in the comparison with gradient descent below.
All quantities in these conditions are fixed independently of $\eta_{1,0}$, so such a positive threshold exists.

For $0<\eta_{1,0}\le\bar\eta_{1,\mathrm A}$, the operator bounds and the first moment estimate imply
\begin{equation*}
    \ip{G_s}{P_s\widehat m_{1,s}} = \ip{G_s}{P_sG_s}+\ip{G_s}{P_s(\widehat m_{1,s}-G_s)}
    \ge \left(p_--\frac{C_{1,m}\eta_{1,s}}{\varepsilon}\right)\Vert G_s\Vert_F^2
    \ge \frac{p_-}{2}\Vert G_s\Vert_F^2.
\end{equation*}
Also,
\begin{equation*}
  \Vert P_s\widehat m_{1,s}\Vert_F\le \frac{\Vert G_s\Vert_F+\Vert\widehat m_{1,s}-G_s\Vert_F}{\varepsilon}\le \frac{1+C_{1,m}}{\varepsilon}\Vert G_s\Vert_F.
\end{equation*}
Taylor's theorem therefore gives
\begin{equation*}
  \begin{aligned}
    L_{s+1} & \le L_s-\eta_{1,s}\ip{G_s}{P_s\widehat m_{1,s}}+\frac{\eta_{1,s}^2}{2K}\Vert P_s\widehat m_{1,s}\Vert_F^2\\
    & \le L_s-\eta_{1,s}\left(\frac{p_-}{2}-\frac{(1+C_{1,m})^2}{2K\varepsilon^2}\eta_{1,s}\right)\Vert G_s\Vert_F^2\\
    & \le L_s-\frac{p_-}{4}\eta_{1,s}\Vert G_s\Vert_F^2.
  \end{aligned}
\end{equation*}
Taking $c_D:=p_-/4$ proves \eqref{eq:adam-phase1-descent}.
Induction from $W_{1,0}=0$ then gives $L_s\le L_0=\log K$, proving \eqref{eq:adam-phase1-sublevel}.

It remains to compare the gradient and the loss.
At $W_{1,s}$, let $\ell_i:=-\log[p_i(W_{1,s})]_{y_i}$ and $q_i:=1-e^{-\ell_i}$.
The loss bound gives $0<\ell_i\le K\log K$.
Concavity of $1-e^{-u}$ on $[0,K\log K]$ and $1-e^{-u}\le u$ imply
\begin{equation*}
  c_K\ell_i\le q_i\le \ell_i.
\end{equation*}
The residual formula above also gives $q_i\le\Vert r_i(W_{1,s})\Vert_2\le\sqrt{2}q_i$.
Consequently,
\begin{equation*}
  \begin{aligned}
    & \Vert G_s\Vert_F\ge \frac{c_K}{K}\left(\sum_{i=1}^K\ell_i^2\right)^{1/2}\ge \frac{c_K}{\sqrt{K}}L_s,\\
    & \Vert G_s\Vert_F\le \frac{\sqrt{2}}{K}\left(\sum_{i=1}^K\ell_i^2\right)^{1/2}\le \sqrt{2}L_s.
  \end{aligned}
\end{equation*}
This proves \eqref{eq:adam-phase1-gradient-loss-bilateral}.
The construction gives the stated dependence of all constants.
\end{proof}

The next lemma gives the growth and limiting direction of the preprocessing iterates.
We prove it by comparing Adam with gradient descent and showing that the sum of the update errors is finite.

\begin{lemma}\label{lem:adam-phase1-frobenius}
Suppose that the hypotheses of Lemma \ref{lem:adam-phase1-descent} hold.
Then $B$ is the unique minimizer of \eqref{eq:adam-phase1-frobenius-hard-margin}, and $R_{1,F}^\star=\sqrt{K-1}$.
For every $s\ge 0$,
\begin{equation}\label{eq:adam-phase1-explicit-loss-reciprocal}
  0<L_s\le (L_0^{-1}+c_LS_{1,s})^{-1}.
\end{equation}
As $s\to\infty$,
\begin{equation}\label{eq:adam-phase1-matrix-asymptotic}
  W_{1,s}=(\log S_{1,s})B+O_F(1),\qquad CW_{1,s}=(\log S_{1,s})B+O_F(1).
\end{equation}
Here $O_F(1)$ denotes a matrix term whose Frobenius norm remains bounded as $s\to\infty$.

In particular, $W_{1,s}$ and $CW_{1,s}$ are nonzero for all sufficiently large $s$.
Their Frobenius normalizations converge to $B/\sqrt{K-1}$ with error $O((\log S_{1,s})^{-1})$.
\end{lemma}

\begin{proof}
For every feasible $W$, summing the margin constraints gives
\begin{equation*}
  K\ip{W}{B}=\sum_{i=1}^K\sum_{v\ne y_i}(e_{y_i}-e_v)^\top Wx_i\ge K(K-1).
\end{equation*}
Since $\Vert B\Vert_F=\sqrt{K-1}$, the Cauchy-Schwarz inequality yields
\begin{equation*}
  K-1\le \ip{W}{B}\le \Vert W\Vert_F\sqrt{K-1}.
\end{equation*}
Thus $\Vert W\Vert_F\ge\sqrt{K-1}$.
The matrix $B$ is feasible and attains this bound.
For any minimizer $W$, we therefore have
\begin{equation*}
  \Vert W-B\Vert_F^2=\Vert W\Vert_F^2+(K-1)-2\ip{W}{B}\le 0.
\end{equation*}
Hence $W=B$, proving uniqueness.

Equations \eqref{eq:adam-phase1-descent} and \eqref{eq:adam-phase1-gradient-loss-bilateral} imply
\begin{equation*}
  L_{s+1}\le L_s-c_L\eta_{1,s}L_s^2.
\end{equation*}
Since $0<L_{s+1}\le L_s$,
\begin{equation*}
  L_{s+1}^{-1}-L_s^{-1}=\frac{L_s-L_{s+1}}{L_sL_{s+1}}\ge c_L\eta_{1,s}\frac{L_s}{L_{s+1}}\ge c_L\eta_{1,s}.
\end{equation*}
Summing gives
\begin{equation*}
  L_s^{-1}\ge L_0^{-1}+c_L\sum_{j=0}^{s-1}\eta_{1,j}=L_0^{-1}+c_LS_{1,s},
\end{equation*}
which proves \eqref{eq:adam-phase1-explicit-loss-reciprocal}.
Summing the loss decrease also yields
\begin{equation*}
  c_L\sum_{j=0}^{s-1}\eta_{1,j}L_j^2\le \sum_{j=0}^{s-1}(L_j-L_{j+1})=L_0-L_s\le L_0.
\end{equation*}
Letting $s\to\infty$, we obtain
\begin{equation}\label{eq:adam-phase1-explicit-loss-summability}
  \sum_{s=0}^{\infty}\eta_{1,s}L_s^2\le \frac{L_0}{c_L}<\infty.
\end{equation}

To compare Adam with gradient descent, we use the operator $P_s$ from the preceding proof.
Let $G_s:=\nabla\Loss_0(W_{1,s})$ and
\begin{equation*}
  \xi_{1,s}:=P_s\widehat m_{1,s}-\varepsilon^{-1}G_s.
\end{equation*}
We separate the first moment error from the change in the denominator:
\begin{equation*}
  \xi_{1,s}=P_s(\widehat m_{1,s}-G_s)+(P_s-\varepsilon^{-1}\mathrm{Id})G_s.
\end{equation*}
For $z\ge 0$,
\begin{equation*}
  0\le \varepsilon^{-1}-(\varepsilon+z)^{-1}=\frac{z}{\varepsilon(\varepsilon+z)}\le \frac{z}{\varepsilon^2}.
\end{equation*}
Using \eqref{eq:adam-phase1-moment-tracking} and \eqref{eq:adam-phase1-gradient-loss-bilateral}, we obtain
\begin{equation*}
  \begin{aligned}
    \Vert\xi_{1,s}\Vert_F & \le \varepsilon^{-1}\Vert\widehat m_{1,s}-G_s\Vert_F+\varepsilon^{-2}\Vert\sqrt{\widehat v_{1,s}}\Vert_{\max}\Vert G_s\Vert_F\\
    & \le \frac{C_{1,m}}{\varepsilon}\eta_{1,s}\Vert G_s\Vert_F+\frac{C_{1,v}}{\varepsilon^2}\Vert G_s\Vert_F^2\\
    & \le \frac{\sqrt{2}C_{1,m}}{\varepsilon}\eta_{1,s}L_s+\frac{2C_{1,v}}{\varepsilon^2}L_s^2.
  \end{aligned}
\end{equation*}
For $s\ge 1$, we have $S_{1,s}\ge\eta_{1,0}s^{1-\alpha_1}$, so
\begin{equation*}
  \eta_{1,s}^2L_s\le \frac{\eta_{1,0}^2(s+1)^{-2\alpha_1}}{c_LS_{1,s}}\le \frac{\eta_{1,0}}{c_L}s^{-1-\alpha_1}.
\end{equation*}
Since $\alpha_1>0$, it follows that $\sum_{s=0}^{\infty}\eta_{1,s}^2L_s<\infty$.
Together with \eqref{eq:adam-phase1-explicit-loss-summability}, this gives
\begin{equation}\label{eq:adam-phase1-scalarization}
  \begin{aligned}
    & W_{1,s+1}=W_{1,s}-(\eta_{1,s}/\varepsilon)G_s-\eta_{1,s}\xi_{1,s},\\
    & \sum_{s=0}^{\infty}\eta_{1,s}\Vert\xi_{1,s}\Vert_F\le \frac{\sqrt{2}C_{1,m}}{\varepsilon}\sum_{s=0}^{\infty}\eta_{1,s}^2L_s+\frac{2C_{1,v}}{\varepsilon^2}\sum_{s=0}^{\infty}\eta_{1,s}L_s^2<\infty.
  \end{aligned}
\end{equation}

Let $\Phi_s(W):=W-(\eta_{1,s}/\varepsilon)\nabla\Loss_0(W)$, and let $Z_0:=0$ and $Z_{s+1}:=\Phi_s(Z_s)$.
The Hessian bound from the previous proof and \eqref{eq:adam-phase1-explicit-step-threshold} give
\begin{equation*}
  0\preceq D\Phi_s(W)=\mathrm{Id}-\frac{\eta_{1,s}}{\varepsilon}\nabla^2\Loss_0(W)\preceq\mathrm{Id}.
\end{equation*}
For any $W,Z\in\R^{K\times d}$, integration along the segment from $Z$ to $W$ therefore yields
\begin{equation*}
  \Vert\Phi_s(W)-\Phi_s(Z)\Vert_F\le \int_0^1\Vert D\Phi_s(Z+\tau(W-Z))[W-Z]\Vert_F\,d\tau\le \Vert W-Z\Vert_F.
\end{equation*}
Using $W_{1,s+1}=\Phi_s(W_{1,s})-\eta_{1,s}\xi_{1,s}$, we obtain
\begin{equation*}
  \Vert W_{1,s+1}-Z_{s+1}\Vert_F\le \Vert W_{1,s}-Z_s\Vert_F+\eta_{1,s}\Vert\xi_{1,s}\Vert_F.
\end{equation*}
Since $W_{1,0}=Z_0=0$, repeated application of this inequality gives
\begin{equation}\label{eq:adam-phase1-gradient-comparison}
  \Vert W_{1,s}-Z_s\Vert_F\le \sum_{j=0}^{s-1}\eta_{1,j}\Vert\xi_{1,j}\Vert_F\le \sum_{j=0}^{\infty}\eta_{1,j}\Vert\xi_{1,j}\Vert_F<\infty.
\end{equation}

The gradient computation in the proof of Lemma \ref{lem:exact-U} gives $\nabla\Loss_0(bB)=-B/(e^b+K-1)$.
Induction therefore shows that $Z_s=b_sB$, where $b_0=0$ and
\begin{equation*}
  b_{s+1}-b_s=\frac{\eta_{1,s}}{\varepsilon(e^{b_s}+K-1)}.
\end{equation*}
The increments are nonnegative, so $b_s\ge 0$.
Let $u_s:=b_{s+1}-b_s$.
Since $b_s\ge 0$ and $\eta_{1,s}\le\eta_{1,0}\le\bar\eta_{1,\mathrm A}\le K\varepsilon$, we have
\begin{equation*}
  0\le u_s=\frac{\eta_{1,s}}{\varepsilon(e^{b_s}+K-1)}\le \frac{\eta_{1,s}}{K\varepsilon}\le 1.
\end{equation*}
Using $u\le e^u-1\le eu$ on $[0,1]$ and $1/K\le e^{b_s}/(e^{b_s}+K-1)\le 1$, we obtain
\begin{equation*}
  \frac{\eta_{1,s}}{K\varepsilon}\le e^{b_s}u_s\le e^{b_{s+1}}-e^{b_s}\le e\,e^{b_s}u_s\le \frac{e\eta_{1,s}}{\varepsilon}.
\end{equation*}
Summing and using $b_0=0$ give
\begin{equation*}
  1+\frac{S_{1,s}}{K\varepsilon}\le e^{b_s}\le 1+\frac{eS_{1,s}}{\varepsilon}.
\end{equation*}

For $s\ge 1$, dividing the previous bounds by $S_{1,s}$ and taking logarithms give
\begin{equation*}
  \log\left(\frac{1}{S_{1,s}}+\frac{1}{K\varepsilon}\right)\le b_s-\log S_{1,s}\le \log\left(\frac{1}{S_{1,s}}+\frac{e}{\varepsilon}\right).
\end{equation*}
Since $S_{1,s}\to\infty$, both bounds converge to finite constants.
Thus $b_s-\log S_{1,s}=O(1)$.

Using $Z_s=b_sB$, \eqref{eq:adam-phase1-gradient-comparison}, and the triangle inequality, we obtain
\begin{equation*}
    \Vert W_{1,s}-(\log S_{1,s})B\Vert_F \le \Vert W_{1,s}-Z_s\Vert_F+|b_s-\log S_{1,s}|\Vert B\Vert_F= O(1).
\end{equation*}
This proves the first identity in \eqref{eq:adam-phase1-matrix-asymptotic}.

Since $CB=B$ and multiplication by $C$ does not increase the Frobenius norm,
\begin{equation*}
  \Vert CW_{1,s}-(\log S_{1,s})B\Vert_F=\Vert C(W_{1,s}-(\log S_{1,s})B)\Vert_F\le \Vert W_{1,s}-(\log S_{1,s})B\Vert_F=O(1).
\end{equation*}
This proves the second identity in \eqref{eq:adam-phase1-matrix-asymptotic}.

For either matrix in \eqref{eq:adam-phase1-matrix-asymptotic}, write $A_s=a_sB+H_s$, where $a_s:=\log S_{1,s}\to\infty$ and $\Vert H_s\Vert_F=O(1)$.
For all sufficiently large $s$, we have $a_s>0$ and
\begin{equation*}
  \Vert A_s\Vert_F\ge a_s\sqrt{K-1}-\Vert H_s\Vert_F>0.
\end{equation*}
The triangle inequality and $\bigl|\Vert A_s\Vert_F-a_s\sqrt{K-1}\bigr|\le\Vert H_s\Vert_F$ then give
\begin{equation*}
  \begin{aligned}
    \left\Vert\frac{A_s}{\Vert A_s\Vert_F}-\frac{B}{\sqrt{K-1}}\right\Vert_F & \le \frac{\Vert H_s\Vert_F+\bigl|\Vert A_s\Vert_F-a_s\sqrt{K-1}\bigr|}{a_s\sqrt{K-1}}\\
    & \le \frac{2\Vert H_s\Vert_F}{a_s\sqrt{K-1}}.
  \end{aligned}
\end{equation*}
This proves the stated convergence and rate.
\end{proof}

Consequently, for $0<\alpha_1<1$, we have $L_s=O(s^{-(1-\alpha_1)})$, and the error after Frobenius normalization is $O((\log(s+1))^{-1})$.
For $\alpha_1=1$, we have $L_s=O((\log(s+1))^{-1})$, and the error after Frobenius normalization is $O((\log\log s)^{-1})$.

\subsection*{\textbf{Main phase}}

Under the data assumptions of Theorem \ref{thm:schatten-unregularized-phase2}, fix a prediction matrix $U$ satisfying \eqref{eq:fixed-prediction-condition}.
We use the notation $\Loss_U(W)=\Loss(U,W)$ and $E_U(W)=\Loss_U(W)-\ell_U^\star$.
Let $R_{2,F}^\star$ be the optimal value of \eqref{eq:common-hard-margin} for $\mathsf N=\Vert\cdot\Vert_F$, and let $W_F^\star$ be a minimizer.
The minimizer is unique because the feasible set is convex and the squared Frobenius norm is strictly convex.
Let $\gamma_{2,F}:=1/R_{2,F}^\star$.

Use $\eta_{2,t}=\eta_{2,0}(t+1)^{-\alpha_2}$ with $0<\alpha_2\le 1$, and let
\begin{equation*}
  E_t:=E_U(W_{2,t}),\qquad L_0:=\Loss_U(0),\qquad S_t:=\sum_{s=0}^{t-1}\eta_{2,s}.
\end{equation*}
The next lemma bounds the two moments and proves that the training loss never exceeds its initial value.
It also gives an excess loss bound and shows that the total attention weight on the correct token tends to one for every sample.

\begin{lemma}\label{lem:adam-phase2-descent}
There are constants $C_m,C_v\ge 0$, depending only on $U$, the finite training data, $\alpha_2,\beta_{2,1},\beta_{2,2},\varepsilon$, such that, for every $0<\eta_{2,0}\le 1$ and every integer $t\ge 0$,
\begin{equation}\label{eq:adam-phase2-moment-tracking}
    \Vert\widehat m_{2,t}-\nabla\Loss_U(W_{2,t})\Vert_F\le C_m\eta_{2,t}\bar s(W_{2,t}),\qquad
    \Vert\sqrt{\widehat v_{2,t}}\Vert_{\max}\le C_v\bar s(W_{2,t}).
\end{equation}
Here $C_m=0$ if $\beta_{2,1}=0$.
Moreover, there are $\bar\eta_U\in(0,1]$ and $c_D,c_R>0$ such that, whenever $0<\eta_{2,0}\le\bar\eta_U$, we have $\Loss_U(W_{2,t})\le L_0$, $a_{n,c}(W_{2,t})\ge\delta_0$ for all $n\in[N]$, and
\begin{align}
  & E_{t+1}\le E_t-c_D\eta_{2,t}\Vert\nabla\Loss_U(W_{2,t})\Vert_F^2,\notag\\
  & 0<E_t\le (E_0^{-1}+c_RS_t)^{-1}\label{eq:adam-phase2-excess-rate}
\end{align}
for every integer $t\ge 0$.
Furthermore, $a_{n,c}(W_{2,t})\to1$ for every $n\in[N]$.
\end{lemma}

\begin{proof}
For this proof, let $W_t:=W_{2,t}$, $\eta_t:=\eta_{2,t}$, and $G_t:=\nabla\Loss_U(W_t)$.
Let $C_0:=2\sqrt{2}Q_U$ and $D_0:=C_0/\varepsilon$.
Equation \eqref{eq:fixed-gradient-upper} gives $\Vert G_t\Vert_F\le C_0\bar s(W_t)\le C_0$.

For $q\in\{1,2\}$, let
\begin{equation*}
  w_{t,r}^{(q)}:=\frac{(1-\beta_{2,q})\beta_{2,q}^r}{1-\beta_{2,q}^{t+1}},\qquad 0\le r\le t.
\end{equation*}
These weights are nonnegative and sum to one, and the moment recursions give
\begin{equation*}
  \widehat m_{2,t}=\sum_{r=0}^tw_{t,r}^{(1)}G_{t-r},\qquad \widehat v_{2,t}=\sum_{r=0}^tw_{t,r}^{(2)}G_{t-r}^{\odot 2}.
\end{equation*}
If $\beta_{2,q}=0$, the corresponding sum contains only the term with $r=0$.
Since every denominator is at least $\varepsilon$,
\begin{equation*}
  \Vert W_{t+1}-W_t\Vert_F\le \frac{\eta_t}{\varepsilon}\Vert\widehat m_{2,t}\Vert_F\le D_0\eta_t.
\end{equation*}
Thus, with $\Pi_{t,r}:=\sum_{j=1}^r\eta_{t-j}$,
\begin{equation*}
  \Vert W_t-W_{t-r}\Vert_F\le D_0\Pi_{t,r}.
\end{equation*}
Lemma \ref{lem:fixed-prediction-stability} gives
\begin{equation*}
  \Vert G_{t-r}-G_t\Vert_F\le L_{1,U}(e^{2D_0\Pi_{t,r}}-1)\bar s(W_t).
\end{equation*}
The gradient upper bound and Lemma \ref{lem:attn-stab} also give
\begin{equation*}
  \Vert G_{t-r}\Vert_F\le C_0\bar s(W_{t-r})\le C_0e^{2D_0\Pi_{t,r}}\bar s(W_t).
\end{equation*}

Let
\begin{equation*}
  C_m:=L_{1,U}C(\alpha_2,2D_0,\beta_{2,1}),\qquad C_v:=C_0\sqrt{1+C(\alpha_2,4D_0,\beta_{2,2})}.
\end{equation*}
Expanding the first moment error and applying Lemma \ref{lem:ema-phase2}, we obtain
\begin{equation*}
    \Vert\widehat m_{2,t}-G_t\Vert_F \le L_{1,U}\bar s(W_t)\sum_{r=0}^tw_{t,r}^{(1)}(e^{2D_0\Pi_{t,r}}-1)
    \le C_m\eta_t\bar s(W_t).
\end{equation*}
For the second moment, the same lemma gives
\begin{equation*}
  \begin{aligned}
    \Vert\sqrt{\widehat v_{2,t}}\Vert_{\max}^2 & \le \sum_{r=0}^tw_{t,r}^{(2)}\Vert G_{t-r}\Vert_F^2\\
    & \le C_0^2\bar s(W_t)^2\left[1+\sum_{r=0}^tw_{t,r}^{(2)}(e^{4D_0\Pi_{t,r}}-1)\right]\\
    & \le C_0^2\left[1+C(\alpha_2,4D_0,\beta_{2,2})\eta_t\right]\bar s(W_t)^2\\
    & \le C_v^2\bar s(W_t)^2.
  \end{aligned}
\end{equation*}
The last inequality uses $\eta_t\le 1$.
This proves \eqref{eq:adam-phase2-moment-tracking} without assuming a loss upper bound.

We next obtain a global Hessian bound from Lemma \ref{lem:fixed-prediction-stability} with $\mathsf N=\Vert\cdot\Vert_F$.
For any $W,Z\in\R^{d\times d}$, smoothness of the loss gives
\begin{equation*}
  \begin{aligned}
    \Vert\nabla^2\Loss_U(W)[Z]\Vert_F & = \lim_{h\to0}\frac{\Vert\nabla\Loss_U(W+hZ)-\nabla\Loss_U(W)\Vert_F}{|h|}\\
    & \le L_{1,U}\bar s(W)\lim_{h\to0}\frac{e^{2|h|\Vert Z\Vert_F}-1}{|h|}\\
    & \le 2L_{1,U}\Vert Z\Vert_F.
  \end{aligned}
\end{equation*}
Here we used $\bar s(W)\le 1$ in the last inequality.
Consequently,
\begin{equation}\label{eq:adam-phase2-global-smoothness}
  \sup_W\Vert\nabla^2\Loss_U(W)\Vert_{F\to F}\le H_U,\qquad H_U:=2L_{1,U}.
\end{equation}

Let $c_F:=c_{\Vert\cdot\Vert_F}$ be the constant from Lemma \ref{lem:hierarchical-gradient}, and let
\begin{equation*}
  P_tZ:=(\sqrt{\widehat v_{2,t}}+\varepsilon)^{-1}\odot Z,\qquad p_-:=(\varepsilon+C_v)^{-1},\qquad A:=\frac{C_m}{c_F\kappa_-}.
\end{equation*}
The second moment bound and $\bar s(W_t)\le 1$ imply
\begin{equation*}
  p_-\mathrm{Id}\preceq P_t\preceq\varepsilon^{-1}\mathrm{Id}.
\end{equation*}
Choose $\bar\eta_U\in(0,1]$ sufficiently small that
\begin{equation*}
  \frac{A\bar\eta_U}{\varepsilon}\le \frac{p_-}{2},\qquad \frac{H_U(1+A)^2\bar\eta_U}{2\varepsilon^2}\le \frac{p_-}{4}.
\end{equation*}
All quantities in these conditions are independent of the choice of $\eta_{2,0}$.

We prove $\Loss_U(W_t)\le L_0$ by induction.
The claim holds at $t=0$ because $W_0=0$.
Suppose that $\Loss_U(W_t)\le L_0$.
Lemma \ref{lem:fixed-prediction-noncollapse} gives $a_{n,c}(W_t)\ge\delta_0$ for all $n\in[N]$.
Equations \eqref{eq:fixed-gradient-excess} and \eqref{eq:fixed-excess-attention} therefore give
\begin{equation*}
  \Vert G_t\Vert_F\ge c_FE_t\ge c_F\kappa_-\bar s(W_t).
\end{equation*}
The first moment estimate now implies
\begin{equation*}
  \Vert\widehat m_{2,t}-G_t\Vert_F\le C_m\eta_t\bar s(W_t)\le A\eta_t\Vert G_t\Vert_F.
\end{equation*}
Using the operator bounds on $P_t$, we obtain
\begin{equation*}
  \begin{aligned}
    \ip{G_t}{P_t\widehat m_{2,t}} & = \ip{G_t}{P_tG_t}+\ip{G_t}{P_t(\widehat m_{2,t}-G_t)}\\
    & \ge \left(p_--\frac{A\eta_t}{\varepsilon}\right)\Vert G_t\Vert_F^2\\
    & \ge \frac{p_-}{2}\Vert G_t\Vert_F^2.
  \end{aligned}
\end{equation*}
Also,
\begin{equation*}
  \Vert P_t\widehat m_{2,t}\Vert_F\le \frac{\Vert G_t\Vert_F+\Vert\widehat m_{2,t}-G_t\Vert_F}{\varepsilon}\le \frac{1+A}{\varepsilon}\Vert G_t\Vert_F.
\end{equation*}
Taylor's theorem and \eqref{eq:adam-phase2-global-smoothness} yield
\begin{equation*}
  \begin{aligned}
    E_{t+1} & \le E_t-\eta_t\ip{G_t}{P_t\widehat m_{2,t}}+\frac{H_U\eta_t^2}{2}\Vert P_t\widehat m_{2,t}\Vert_F^2\\
    & \le E_t-\eta_t\left(\frac{p_-}{2}-\frac{H_U(1+A)^2}{2\varepsilon^2}\eta_t\right)\Vert G_t\Vert_F^2\\
    & \le E_t-\frac{p_-}{4}\eta_t\Vert G_t\Vert_F^2.
  \end{aligned}
\end{equation*}
Let $c_D:=p_-/4$.
Since $E_t=\Loss_U(W_t)-\ell_U^\star$, this gives $\Loss_U(W_{t+1})\le\Loss_U(W_t)\le L_0$ and closes the induction.
The loss and attention bounds therefore hold at every iteration.

Let $c_R:=c_Dc_F^2$.
Using $\Vert G_t\Vert_F\ge c_FE_t$ in the descent estimate gives
\begin{equation*}
  E_{t+1}\le E_t-c_R\eta_tE_t^2.
\end{equation*}
Equation \eqref{eq:fixed-excess-attention} gives $E_t>0$ at every finite iteration.
Thus
\begin{equation*}
  E_{t+1}^{-1}-E_t^{-1}=\frac{E_t-E_{t+1}}{E_tE_{t+1}}\ge c_R\eta_t\frac{E_t}{E_{t+1}}\ge c_R\eta_t.
\end{equation*}
Summing yields $E_t^{-1}\ge E_0^{-1}+c_RS_t$, proving \eqref{eq:adam-phase2-excess-rate}.
The same descent estimate gives
\begin{equation*}
  c_R\sum_{s=0}^{t-1}\eta_{2,s}E_s^2\le \sum_{s=0}^{t-1}(E_s-E_{s+1})=E_0-E_t\le E_0.
\end{equation*}
Letting $t\to\infty$, we obtain
\begin{equation}\label{eq:adam-phase2-summability}
  \sum_{t=0}^{\infty}\eta_{2,t}E_t^2\le \frac{E_0}{c_R}<\infty.
\end{equation}

Since $0<\alpha_2\le 1$, we have $S_t\to\infty$ and hence $E_t\to0$.
For every $n\in[N]$, \eqref{eq:fixed-excess-attention} implies
\begin{equation*}
  0\le 1-a_{n,c}(W_t)=s_n(W_t)\le N\bar s(W_t)\le \frac{N}{\kappa_-}E_t\longrightarrow0.
\end{equation*}
This proves the attention convergence.
\end{proof}

We next bound the Frobenius norm of the iterates in terms of the logarithm of the inverse excess loss.
Combining this estimate with the loss lower bound in \eqref{eq:fixed-loss-margin} gives convergence of the normalized margin and the parameter direction.

\begin{lemma}\label{lem:adam-phase2-sharp-path}
Consider the iterates in Lemma \ref{lem:adam-phase2-descent} with $0<\eta_{2,0}\le \bar\eta_U$, and let $J_t:=\log(E_t^{-1})$.
There is $C_W>0$ such that, for every integer $t\ge 0$,
\begin{equation}\label{eq:adam-phase2-sharp-path-norm}
  \Vert W_{2,t}\Vert_F\le \gamma_{2,F}^{-1}J_t+C_W.
\end{equation}
Moreover, there are constants $C_{\mathrm{mar}},C_{\mathrm{dir}}>0$ and an integer $t_{\mathrm{mar}}\ge 1$ such that $W_{2,t}\ne0$ and
\begin{align}
  & 0\le \gamma_{2,F}-\frac{\rho_2(W_{2,t})}{\Vert W_{2,t}\Vert_F}\le \frac{C_{\mathrm{mar}}}{1+\log(1+S_t)},\label{eq:adam-main-normalized-margin-rate}\\
  & \left\Vert\frac{W_{2,t}}{\Vert W_{2,t}\Vert_F}-\frac{W_F^\star}{R_{2,F}^\star}\right\Vert_F\le \frac{C_{\mathrm{dir}}}{\sqrt{1+\log(1+S_t)}}\label{eq:adam-main-normalized-direction-rate}
\end{align}
for every integer $t\ge t_{\mathrm{mar}}$.
\end{lemma}

\begin{proof}
For this proof, let $W_t:=W_{2,t}$, $\eta_t:=\eta_{2,t}$, and $G_t:=\nabla\Loss_U(W_t)$.
Equations \eqref{eq:fixed-gradient-upper} and \eqref{eq:fixed-excess-attention} give
\begin{equation*}
  \Vert G_t\Vert_F\le 2\sqrt{2}Q_U\bar s(W_t)\le \frac{2\sqrt{2}Q_U}{\kappa_-}E_t.
\end{equation*}
As in the previous proof, let $P_tZ:=(\sqrt{\widehat v_{2,t}}+\varepsilon)^{-1}\odot Z$.
We write the Adam update as
\begin{equation*}
  W_{t+1}-W_t=-\frac{\eta_t}{\varepsilon}G_t+r_t,
\end{equation*}
where
\begin{equation*}
  r_t=-\eta_tP_t(\widehat m_{2,t}-G_t)-\eta_t(P_t-\varepsilon^{-1}\mathrm{Id})G_t.
\end{equation*}
For $z\ge 0$, we have
\begin{equation*}
  0\le \varepsilon^{-1}-(\varepsilon+z)^{-1}=\frac{z}{\varepsilon(\varepsilon+z)}\le \frac{z}{\varepsilon^2}.
\end{equation*}
Using this inequality, \eqref{eq:adam-phase2-moment-tracking}, and $\bar s(W_t)\le E_t/\kappa_-$, we obtain
\begin{equation*}
    \Vert r_t\Vert_F \le \frac{\eta_t}{\varepsilon}\Vert\widehat m_{2,t}-G_t\Vert_F+\frac{\eta_t}{\varepsilon^2}\Vert\sqrt{\widehat v_{2,t}}\Vert_{\max}\Vert G_t\Vert_F
    \le C(\eta_t^2E_t+\eta_tE_t^2).
\end{equation*}

For $t\ge 1$, the step sizes satisfy $S_t\ge \eta_{2,0}t^{1-\alpha_2}$.
Hence \eqref{eq:adam-phase2-excess-rate} gives
\begin{equation*}
  \eta_t^2E_t\le \frac{\eta_{2,0}^2(t+1)^{-2\alpha_2}}{c_RS_t}\le \frac{\eta_{2,0}}{c_R}t^{-1-\alpha_2}.
\end{equation*}
Since $\alpha_2>0$, we have $\sum_{t=0}^{\infty}\eta_t^2E_t<\infty$.
Together with \eqref{eq:adam-phase2-summability}, this proves
\begin{equation*}
  \sum_{t=0}^{\infty}\Vert r_t\Vert_F\le C\sum_{t=0}^{\infty}(\eta_t^2E_t+\eta_tE_t^2)<\infty.
\end{equation*}

We now relate the accumulated gradient terms to $J_t$.
Let $\Delta_t:=W_{t+1}-W_t$.
The moment estimates and the gradient upper bound imply
\begin{equation*}
  \Vert\Delta_t\Vert_F\le \frac{\eta_t}{\varepsilon}\left(\Vert G_t\Vert_F+\Vert\widehat m_{2,t}-G_t\Vert_F\right)\le C\eta_tE_t.
\end{equation*}
Taylor's theorem and \eqref{eq:adam-phase2-global-smoothness} give
\begin{equation*}
  E_{t+1}=E_t-\frac{\eta_t}{\varepsilon}\Vert G_t\Vert_F^2+q_t,
\end{equation*}
where
\begin{equation*}
    |q_t| \le \Vert G_t\Vert_F\Vert r_t\Vert_F+\frac{H_U}{2}\Vert\Delta_t\Vert_F^2
    \le CE_t\Vert r_t\Vert_F+C\eta_t^2E_t^2.
\end{equation*}
It follows that
\begin{equation*}
  \sum_{t=0}^{\infty}\frac{|q_t|}{E_t}\le C\sum_{t=0}^{\infty}\Vert r_t\Vert_F+C\sum_{t=0}^{\infty}\eta_t^2E_t<\infty.
\end{equation*}
Since $E_t,E_{t+1}>0$, the inequality $-\log z\ge 1-z$ gives
\begin{equation*}
    J_{t+1}-J_t = -\log(E_{t+1}/E_t)
    \ge \frac{E_t-E_{t+1}}{E_t}
    = \frac{\eta_t}{\varepsilon}\frac{\Vert G_t\Vert_F^2}{E_t}-\frac{q_t}{E_t}.
\end{equation*}
Summing over $s=0,\ldots,t-1$, we obtain
\begin{equation*}
  \sum_{s=0}^{t-1}\frac{\eta_s}{\varepsilon}\frac{\Vert G_s\Vert_F^2}{E_s}\le J_t-J_0+\sum_{s=0}^{t-1}\frac{|q_s|}{E_s}\le J_t+C.
\end{equation*}

For the Frobenius norm, \eqref{eq:fixed-global-margin-gradient} gives
\begin{equation*}
  \Vert G_t\Vert_F\ge \gamma_{2,F}E_t-C_UE_t^2.
\end{equation*}
Dividing by $E_t>0$ and rearranging yield $\gamma_{2,F}\le \Vert G_t\Vert_F/E_t+C_UE_t$.
Multiplying by $\eta_t\Vert G_t\Vert_F/\varepsilon$, and then using $\Vert G_t\Vert_F\le CE_t$, we obtain
\begin{equation*}
    \gamma_{2,F}\frac{\eta_t}{\varepsilon}\Vert G_t\Vert_F \le \frac{\eta_t}{\varepsilon}\frac{\Vert G_t\Vert_F^2}{E_t}+\frac{C_U\eta_t}{\varepsilon}E_t\Vert G_t\Vert_F
    \le \frac{\eta_t}{\varepsilon}\frac{\Vert G_t\Vert_F^2}{E_t}+C\eta_tE_t^2.
\end{equation*}
Since $W_0=0$, summing the updates gives
\begin{equation*}
  \begin{aligned}
    \gamma_{2,F}\Vert W_t\Vert_F & \le \gamma_{2,F}\sum_{s=0}^{t-1}\frac{\eta_s}{\varepsilon}\Vert G_s\Vert_F+\gamma_{2,F}\sum_{s=0}^{t-1}\Vert r_s\Vert_F\\
    & \le \sum_{s=0}^{t-1}\frac{\eta_s}{\varepsilon}\frac{\Vert G_s\Vert_F^2}{E_s}+C\sum_{s=0}^{t-1}\eta_sE_s^2+\gamma_{2,F}\sum_{s=0}^{t-1}\Vert r_s\Vert_F\\
    & \le J_t+C.
  \end{aligned}
\end{equation*}
This proves \eqref{eq:adam-phase2-sharp-path-norm} for a sufficiently large $C_W>0$.

We next combine the norm bound with the margin lower bound.
Equation \eqref{eq:fixed-loss-margin} gives
\begin{equation*}
  \rho_2(W_t)\ge J_t+\log\nu_U\ge J_t-|\log\nu_U|.
\end{equation*}
Since $E_t\to0$, we have $J_t\to\infty$.
Thus, for all sufficiently large $t$, the margin is positive and $W_t\ne0$.
For such $t$, let $V_t:=W_t/\Vert W_t\Vert_F$ and $m_t:=\rho_2(V_t)$.
The maximum margin identity \eqref{eq:common-margin-witness} gives $m_t\le \gamma_{2,F}$.
Using positive homogeneity and \eqref{eq:adam-phase2-sharp-path-norm}, we obtain
\begin{equation*}
  m_t=\frac{\rho_2(W_t)}{\Vert W_t\Vert_F}\ge \frac{J_t-|\log\nu_U|}{\gamma_{2,F}^{-1}J_t+C_W}.
\end{equation*}
Consequently,
\begin{equation*}
  0\le \gamma_{2,F}-m_t\le \frac{\gamma_{2,F}C_W+|\log\nu_U|}{\gamma_{2,F}^{-1}J_t+C_W}.
\end{equation*}
Moreover, \eqref{eq:adam-phase2-excess-rate} gives
\begin{equation*}
  J_t\ge \log(E_0^{-1}+c_RS_t)\ge \log(1+S_t)+\log\min\{E_0^{-1},c_R\}.
\end{equation*}
As $S_t\to\infty$, this implies $J_t\ge \frac{1}{2}[1+\log(1+S_t)]$ for all sufficiently large $t$.
Substituting into the previous margin estimate proves \eqref{eq:adam-main-normalized-margin-rate}.

Finally, we use the Frobenius hard margin problem to control the direction error.
Since $m_t>0$, the matrix $V_t/m_t$ is feasible.
The feasible set is convex, and $W_F^\star$ also minimizes $\frac{1}{2}\Vert W\Vert_F^2$ over this set.

For $0\le \lambda\le 1$, the matrix $W_F^\star+\lambda(V_t/m_t-W_F^\star)$ is feasible because it is a convex combination of two feasible matrices.
Since $W_F^\star$ minimizes the squared Frobenius norm over the feasible set, for $0<\lambda\le 1$ we have
\begin{equation*}
    0 \le \frac{1}{2}\left\Vert W_F^\star+\lambda(V_t/m_t-W_F^\star)\right\Vert_F^2-\frac{1}{2}\Vert W_F^\star\Vert_F^2
    = \lambda\ip{W_F^\star}{V_t/m_t-W_F^\star}+\frac{\lambda^2}{2}\Vert V_t/m_t-W_F^\star\Vert_F^2.
\end{equation*}
Dividing by $\lambda>0$ and letting $\lambda\downarrow0$ give
\begin{equation*}
  0\le \ip{W_F^\star}{V_t/m_t-W_F^\star}=\frac{\ip{W_F^\star}{V_t}}{m_t}-\Vert W_F^\star\Vert_F^2.
\end{equation*}
Since $m_t>0$ and $\Vert W_F^\star\Vert_F=R_{2,F}^\star=\gamma_{2,F}^{-1}$, it follows that
\begin{equation*}
  \ip{W_F^\star}{V_t}\ge m_t\Vert W_F^\star\Vert_F^2=\frac{m_t}{\gamma_{2,F}^2}.
\end{equation*}

Both $V_t$ and $\gamma_{2,F}W_F^\star$ have unit Frobenius norm.
Expanding the squared distance and using the inner product bound, we obtain
\begin{equation*}
  \begin{aligned}
    \left\Vert V_t-\gamma_{2,F}W_F^\star\right\Vert_F^2 & = \Vert V_t\Vert_F^2+\gamma_{2,F}^2\Vert W_F^\star\Vert_F^2-2\gamma_{2,F}\ip{W_F^\star}{V_t}\\
    & = 2-2\gamma_{2,F}\ip{W_F^\star}{V_t}\\
    & \le \frac{2(\gamma_{2,F}-m_t)}{\gamma_{2,F}}.
  \end{aligned}
\end{equation*}
Taking square roots and applying \eqref{eq:adam-main-normalized-margin-rate} prove \eqref{eq:adam-main-normalized-direction-rate}, since $\gamma_{2,F}W_F^\star=W_F^\star/R_{2,F}^\star$.
Both rates hold for every $t\ge t_{\mathrm{mar}}$ after choosing $t_{\mathrm{mar}}$ sufficiently large.
\end{proof}

We now combine the preprocessing and main phase estimates for Adam.

\begin{theorem}\label{thm:adam-two-phase-convergence}
Suppose that $K\ge 2$, $h$ is bijective, the vocabulary embeddings are orthonormal, and the $K$ single token preprocessing samples are equally weighted.
Assume that Assumptions \ref{ass:token-label}, \ref{ass:weak-count}, and \ref{ass:separability} hold and $\I_2\ne\varnothing$.
Use Algorithm \ref{alg:adam-two-phase} with fixed $\varepsilon>0$, $\beta_{j,1},\beta_{j,2}\in[0,1)$, and
\begin{equation*}
  \eta_{j,s}=\eta_{j,0}(s+1)^{-\alpha_j},\qquad 0<\alpha_j\le 1,\qquad j\in\{1,2\}.
\end{equation*}
There is $\bar\eta_{1,\mathrm A}\in(0,1]$ such that, for every $0<\eta_{1,0}\le \bar\eta_{1,\mathrm A}$, there is an integer $T_{\mathrm A}\ge 1$ for which every $T\ge T_{\mathrm A}$ produces $U:=W_{1,T}$ satisfying \eqref{eq:fixed-prediction-condition}.
For each such $U$, there are $\bar\eta_U\in(0,1]$ and $c_R>0$, independent of $\eta_{2,0}$, such that every $0<\eta_{2,0}\le \bar\eta_U$ gives the following conclusions for some constants $C_{1,\mathrm A},C_{2,\mathrm A}>0$.

\textbf{\textup{(i) Training loss.}}
For every integer $t\ge 0$,
\begin{equation}\label{eq:adam-main-excess-rate}
  0<E_U(W_{2,t})\le \left(E_U(0)^{-1}+c_RS_t\right)^{-1}.
\end{equation}
In particular, $\Loss(U,W_{2,t})\to\ell_U^\star$.

\textbf{\textup{(ii) Normalized Frobenius margin.}}
There is an integer $t_U\ge 1$ such that, for every $t\ge t_U$, we have $W_{2,t}\ne0$ and
\begin{equation}\label{eq:adam-thm-margin}
  0\le \gamma_{2,F}-\rho_2\left(\frac{W_{2,t}}{\Vert W_{2,t}\Vert_F}\right)\le \frac{C_{1,\mathrm A}}{1+\log(1+S_t)}.
\end{equation}
The normalized margin converges to $\gamma_{2,F}$, with gap $O((\log(t+1))^{-1})$ for $0<\alpha_2<1$ and $O((\log\log t)^{-1})$ for $\alpha_2=1$.

\textbf{\textup{(iii) Normalized direction.}}
For every $t\ge t_U$,
\begin{equation}\label{eq:adam-thm-direction}
  \left\Vert\frac{W_{2,t}}{\Vert W_{2,t}\Vert_F}-\frac{W_F^\star}{R_{2,F}^\star}\right\Vert_F\le \frac{C_{2,\mathrm A}}{\sqrt{1+\log(1+S_t)}}.
\end{equation}
Consequently, $W_{2,t}/\Vert W_{2,t}\Vert_F\to W_F^\star/R_{2,F}^\star$.
\end{theorem}

\begin{proof}
Choose $\bar\eta_{1,\mathrm A}$ as in Lemma \ref{lem:adam-phase1-descent}, and fix $0<\eta_{1,0}\le \bar\eta_{1,\mathrm A}$.
Since $0<\alpha_1\le 1$, we have $S_{1,T}\to\infty$.
Lemma \ref{lem:adam-phase1-frobenius} gives
\begin{equation*}
  CW_{1,T}=a_TB+H_T,\qquad a_T:=\log S_{1,T}\to\infty,\qquad \Vert H_T\Vert_F=O(1).
\end{equation*}
For $U=W_{1,T}$, the definition of $\delta$ in \eqref{eq:fixed-output-decomposition} and $\Vert B\Vert_F^2=K-1$ imply
\begin{equation*}
  \delta=a_T+\frac{\ip{H_T}{B}}{K-1},\qquad |\delta-a_T|\le \frac{\Vert H_T\Vert_F}{\sqrt{K-1}}=O(1).
\end{equation*}
Also, $R=H_T-(\delta-a_T)B$, so
\begin{equation*}
  r_U\le \Vert R\Vert_F\le \Vert H_T\Vert_F+|\delta-a_T|\Vert B\Vert_F\le 2\Vert H_T\Vert_F=O(1).
\end{equation*}
Thus $\delta\to\infty$ while $r_U$ remains bounded, and hence
\begin{equation*}
  \frac{2(K-1)[N\log K+\sqrt{2}(N+1)r_U]}{\delta}\longrightarrow0.
\end{equation*}
We can therefore choose an integer $T_{\mathrm A}\ge 1$ such that $\delta>0$ and this ratio is at most one for every $T\ge T_{\mathrm A}$.
This proves \eqref{eq:fixed-prediction-condition}.

Fix such a $T$ and let $U:=W_{1,T}$.
Lemma \ref{lem:adam-phase2-descent} supplies $\bar\eta_U$ and $c_R$, independently of $\eta_{2,0}$.
For $0<\eta_{2,0}\le \bar\eta_U$, its excess loss estimate gives \eqref{eq:adam-main-excess-rate}.
Since $S_t\to\infty$, we have $E_U(W_{2,t})\to0$ and therefore $\Loss(U,W_{2,t})\to\ell_U^\star$.

Lemma \ref{lem:adam-phase2-sharp-path} gives an integer $t_U\ge 1$ such that $W_{2,t}\ne0$ and both \eqref{eq:adam-thm-margin} and \eqref{eq:adam-thm-direction} hold for every $t\ge t_U$.
The right sides tend to zero because $S_t\to\infty$, proving the margin and direction limits.
Finally, $S_t=\Theta(t^{1-\alpha_2})$ for $0<\alpha_2<1$ and $S_t=\Theta(\log(t+1))$ for $\alpha_2=1$.
Substitution into \eqref{eq:adam-thm-margin} gives the stated rates in terms of $t$.
\end{proof}

\end{document}